\documentclass{article}

\usepackage{microtype}
\usepackage{graphicx}
\graphicspath{{./img/}}
\usepackage{subfig}			% subfloat
\usepackage{booktabs} % for professional tables
\usepackage[usenames,dvipsnames]{xcolor}  % to remove warnings due to xcolor

\usepackage{hyperref}
\hypersetup{
    pdfsubject={Symmetric Spaces for Graph Embeddings: A Finsler-Riemannian Approach},
    pdftitle={Symmetric Spaces for Graph Embeddings: A Finsler-Riemannian Approach},
    pdfauthor={Federico Lopez},
    pdfkeywords={hermitian symmetric spaces, graph embeddings, siegel space, hyperbolic geometry, upper half space, bounded domain model, matrix models, deep learning, riemannian manifold learning, non-euclidean geometry, finsler metrics, vector valued distance, symmetric positive definite matrices, spd, hyperbolic space, manifold learning}
}

\newcounter{toolkit}
\makeatletter
\newenvironment{toolkit}[1][htb]{%
  \let\c@algorithm\c@toolkit
  \renewcommand{\ALG@name}{Toolkit}% Update algorithm name
  \begin{algorithm}[#1]%
  }{\end{algorithm}
}
\makeatother

\usepackage[accepted]{icml2021}

\usepackage{todonotes}
\usepackage{amsmath}
\usepackage{amssymb}
\usepackage{amsfonts}       % blackboard math symbols
\usepackage{adjustbox}
\usepackage{wrapfig}
\usepackage{nicefrac}       % compact symbols for 1/2, etc.
\usepackage{amsthm}

\newtheorem*{remark}{Remark}

\newcommand{\Id}{{\rm Id}}
\newcommand{\R}{\mathbb R}
\newcommand{\Sym}{{\rm Sym}}
\newcommand{\C}{\mathbb C}
\newcommand{\Sp}{{\rm Sp}}
\newcommand{\calB}{\mathcal B}
\newcommand{\SL}{{\rm SL}}
\newcommand{\SO}{{\rm SO}}
\newcommand{\SU}{{\rm SU}}
\newcommand{\UU}{{\rm U}}
\newcommand{\rk}{{\rm rk}}
\newcommand{\tr}{{\rm tr}}

\newcommand{\calS}{\mathcal S}
\newcommand{\bpm}{\begin{pmatrix}}
\newcommand{\epm}{\end{pmatrix}}
\newcommand{\bsm}{\left(\begin{smallmatrix}}
\newcommand{\esm}{\end{smallmatrix}\right)}
\newcommand{\ov}{\overline}
\newcommand{\bigO}{\mathcal{O}}
\newcommand{\HH}{\mathbb H}
\newcommand{\finf}{F_{\infty}}
\newcommand{\fone}{F_{1}}
\newcommand{\spd}{\operatorname{SPD}}
\newcommand{\calD}{\mathcal D}

\icmltitlerunning{Symmetric Spaces for Graph Embeddings}

\begin{document}

\twocolumn[
\icmltitle{Symmetric Spaces for Graph Embeddings: A Finsler-Riemannian Approach}



\icmlsetsymbol{equal}{*}

\begin{icmlauthorlist}
\icmlauthor{Federico L\'opez}{hits}
\icmlauthor{Beatrice Pozzetti}{hduni}
\icmlauthor{Steve Trettel}{sta}
\icmlauthor{Michael Strube}{hits}
\icmlauthor{Anna Wienhard}{hduni}
\end{icmlauthorlist}

\icmlaffiliation{hduni}{Mathematical Institute, Heidelberg University, Heidelberg, Germany}
\icmlaffiliation{hits}{Heidelberg Institute for Theoretical Studies, Heidelberg, Germany}
\icmlaffiliation{sta}{Department of Mathematics, Stanford University, California, USA}

\icmlcorrespondingauthor{Federico López}{federico.lopez@h-its.org}

\icmlkeywords{symmetric space, symmetric matrices, graph embeddings, riemannian geometry, Finsler metrics, representation learning}

\vskip 0.3in
]



\printAffiliationsAndNotice{} % otherwise use the standard text.

\begin{abstract}
%\todo[inline]{to be rewritten to take up some of the points made in the introduction} 
%The quality of the embedding is usually determined by how well the geometry of the target space matches the structure of the data. 
Learning faithful graph representations as sets of vertex embeddings has become a fundamental intermediary step in a wide range of machine learning applications.
We propose the systematic use of symmetric spaces in representation learning, a class encompassing many of the previously used embedding targets. 
This enables us to introduce a new method, the use of Finsler metrics integrated in a Riemannian optimization scheme, that better adapts to dissimilar structures in the graph. We develop a tool to analyze the embeddings and infer structural properties of the data sets. 
For implementation, we choose Siegel spaces, a versatile family of symmetric spaces.
Our approach outperforms competitive baselines for graph reconstruction tasks on various synthetic and real-world datasets.
We further demonstrate its applicability on two downstream tasks, recommender systems and node classification.
\end{abstract}

% {\bf Possible title}
% "Symmetric spaces - a unified approach for graph embeddings" 
% "Symmetric spaces for Graph Embeddings - a Riemannian-Finslerian approach", "Leveraging Finsler Metrics in Hermitian Symmetric Spaces for Graph Embeddings", "The Finsler metric on Hermitian Symmetric spaces for Graph Embeddings", "Finsler Metrics in Hermitian Symmetric Spaces for Graph Embeddings", "Generalizing Something ..."

%%%%%%%%%%%%%%%%%%%%%%%% INTRO %%%%%%%%%%%%%%%%%%%%%%%%%%%%%%%%%%%%%
\section{Introduction}
%
% Introduction of the problem we want to solve
%
The goal of representation learning  is to embed real-world data, frequently modeled on a graph, into an ambient space. This embedding space can then be used to analyze and perform tasks on the discrete graph. 
The predominant approach has been to embed discrete structures in an Euclidean space. Nonetheless, data in many domains 
%including social \cite{lazer09lifeinnet, verbeek14social}, sensor \cite{gao2012sensor}, gene \cite{Davidson1669}, protein  molecular \cite{gainza2020molecules} and complex \cite{krioukov2010hypernetworks} networks or the Internet \cite{boguna2010internet} 
exhibit non-Euclidean features \cite{krioukov2010hypernetworks, bronstein2018geomdeeplearning}, making embeddings into Riemannian manifolds with a richer structure necessary. 
% 
% What previous work has made to solve the problem
%
For this reason, embeddings into hyperbolic \cite{krioukov2009tempNets, nickel2017poincare, deSa18tradeoffs, lopez2020fullyhyper} and spherical spaces \cite{wilson2014sphere, liu2017sphereface, xu2018sphericalVAE} have been developed.
Recent work proposes to combine different curvatures through several layers \cite{chami2019hgcnn,bachmann2020ccgcn, grattarola2020constantCurvatureGraphEmbeds},  to enrich the geometry by considering Cartesian products of spaces \cite{gu2019lmixedCurvature, tifrea2018poincareGlove, skopek2020mixedva}, or to use Grassmannian manifolds or the space of  symmetric positive definite matrices (SPD) as a trade-off between the representation capability and the computational tractability of the space \cite{huang2017riemannianNetForSPDMatrix, huang2018DeepNetsOnGrassmannManifolds, cruceru20matrixGraph}.
%
% Issues with previous work
%
A unified framework in which to encompass these various examples is still missing.

%However, their ability to model complex patterns is inherently bounded by the geometric properties of the target embedding space, which has to be picked beforehand. 
%The choice of a metric space where to embed the data can be understood as selecting an inductive bias \cite{tifrea2018poincareGlove}, which does not always adapt to all parts of the graphs. 
%Moreover, there is a trade-off between the representation capability and the computational tractability of the space. In this view, Grassmannian manifolds or the space of positive definite symmetric matrices have been proposed as suitable alternatives \cite{huang2017riemannianNetForSPDMatrix, huang2018DeepNetsOnGrassmannManifolds, cruceru20matrixGraph}.

\begin{figure}[!t]
    \centering
    \includegraphics[width=0.4\textwidth,keepaspectratio]{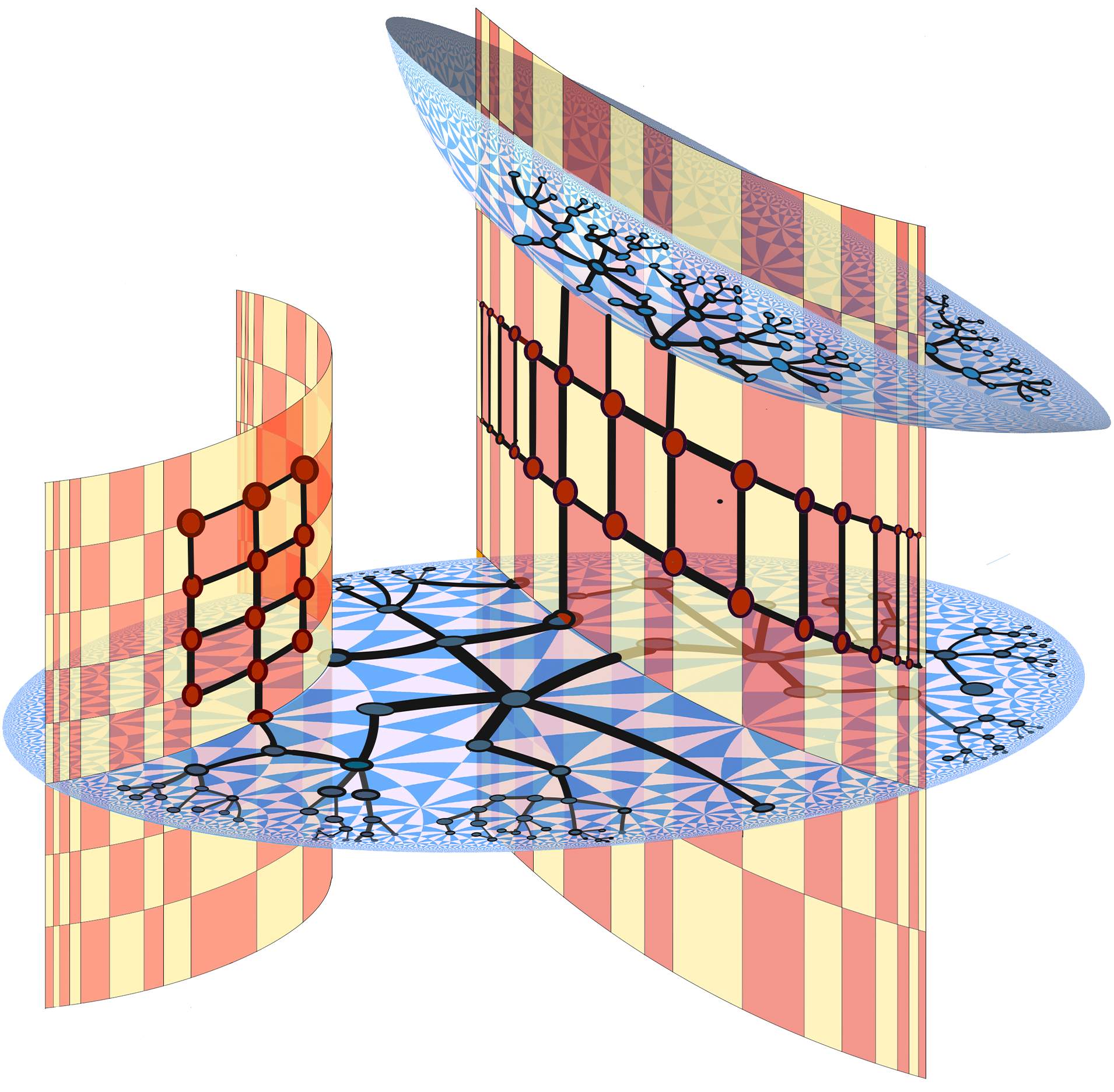}
    \caption{{Symmetric spaces have a rich structure of totally geodesic subspaces, including flat subspaces (orange) and hyperbolic planes (blue). This compound, yet computationally tractable geometry allows isometric embeddings of many graphs, including those with subgraphs of dissimilar geometry. For example the graph embedded in the picture has both trees and grids as subgraphs}.}
    %\caption{The rich geometry of symmetric spaces allows isometric embedding of many graphs, including those with subgraphs of dissimilar geometries.}
    \label{fig:varying-curvature}
    \vspace{-4mm}
\end{figure}

In this work, we propose the systematic use of symmetric spaces in representation learning: this is a class comprising all the aforementioned spaces.
Symmetric spaces are Riemannian manifolds with rich symmetry groups which makes them algorithmically tractable. They have a compound geometry that simultaneously contains Euclidean as well as hyperbolic or spherical subspaces, allowing them to automatically adapt to dissimilar features in the graph. 
We develop a general framework to choose a Riemannian symmetric space and implement the mathematical tools required to learn graph embeddings (\S\ref{sec:symmetric-spaces-for-embeddings}). 
Our systematic view enables us to introduce the use of Finsler metrics integrated with a Riemannian optimization scheme as a new method to achieve graph representations. 
Moreover, we use a vector-valued distance function on symmetric spaces to develop a new tool for the analysis of the structural properties of the embedded graphs.

To demonstrate a concrete implementation of our general framework, we choose Siegel spaces \cite{siegel1943symplectic}; a family of symmetric spaces that has not been explored in geometric deep learning, despite them being among the most versatile symmetric spaces of non-positive curvature. Key features of Siegel spaces are that they are matrix versions of the hyperbolic plane, they contain many products of hyperbolic planes as well as copies of SPD as submanifolds, and they support Finsler metrics that induce the $\ell^1$ or the $\ell^\infty$ metric on the Euclidean subspaces.
As we verify in experiments, these metrics are well suited to embed graphs of mixed geometric features. This makes Siegel spaces with Finsler metrics an excellent %\todo{MS: device, means}
device for embedding complex networks without a priori knowledge of their internal structure. 
% \todo{F: so what? what are the consequences of this? what does that imply in concrete terms? Suggestion: add the phrase "This makes them an excellent target for learning embeddings of complex networks without a priori knowledge of their internal structure." here as well.}
%; version of these metrics on the Euclidean spaces have been very successfully applied in compressed sensing \todo{Add citation}{\cite{??}}.

Siegel spaces are realized as spaces of symmetric matrices with coefficients in the complex numbers $\C$. 
By combining their explicit models and the general structure theory of symmetric spaces with the Takagi factorization \cite{takagiFactorization1924} and the Cayley transform \cite{cayley1846transform}, we achieve a tractable and automatic-differentiable algorithm to compute distances in Siegel spaces (\S\ref{sec:implementation-of-matrix-models}). This allows us to learn embeddings through Riemannian optimization \cite{bonnabel2011rsgd}, which is easily parallelizable and scales to large datasets.
Moreover, we highlight the properties of the Finsler metrics on these spaces (\S\ref{sec:Siegel}) and integrate them with the Riemannian optimization tools.
We evaluate the representation capacities of the Siegel spaces for the task of graph reconstruction on real and synthetic datasets.
%For synthetic graphs we introduce a new class of benchmarking graphs that mix tree- and grid-like features at different levels and scales. This  makes them more complex than the graphs used in \cite{gu2019lmixedCurvature} and gives a better approximation of real-world datasets.
We find that Siegel spaces endowed with Finsler metrics outperform Euclidean, hyperbolic, Cartesian products of these spaces and SPD in all analyzed datasets. These results manifest the effectiveness and versatility of the proposed approach, particularly for graphs with varying and intricate structures.

To showcase potential applications of our approach in different graph embedding pipelines, we also test its capabilities for recommender systems and node classification. We find that our models surpass competitive baselines (constant-curvature, products thereof and SPD) for several real-world datasets.

{\bf Related Work:} Riemannian manifold learning has regained attention due to appealing geometric properties that allow methods to represent non-Euclidean data arising in several domains \cite{rubindelanchy2020manifold}. 
Our systematic approach to symmetric spaces  comprises embeddings in hyperbolic spaces \cite{Chamberlain2017, ganea2018hyperEntailmentCones, nickel2018lorentz, lopez2019figetinHS}, spherical spaces \cite{meng2019sphericalTextEmbed, defferrard2020DeepSphere}, combinations thereof \cite{bachmann2020ccgcn, grattarola2020constantCurvatureGraphEmbeds, law2020ultrahyperbolic}, Cartesian products of spaces \cite{gu2019lmixedCurvature, tifrea2018poincareGlove}, Grassmannian manifolds \cite{huang2018DeepNetsOnGrassmannManifolds} and the space of  symmetric positive definite matrices (SPD) \cite{huang2017riemannianNetForSPDMatrix, cruceru20matrixGraph}, among others.
%, but can also be seen as a generalization that includes previous work on learning graph embeddings in %Hilbert spaces \cite{Sriperumbudur2010hilberSpaceEmbeds, herath2017LearningHilbertSpace}, 
%Lie groups \cite{falorsi2018homeovae} (such as the torus \cite{ebisu2018toruse}), non-abelian groups \cite{yang2020nonAbelianGroupsKG}, Dihedral groups \cite{xu2019dihedralKG}, or pseudo-Riemannian manifold of constant nonzero curvature \cite{law2020ultrahyperbolic}.
We implement our method on Siegel spaces. To the best of our knowledge, we are the first work to apply them in Geometric Deep Learning.

%Graph convolutional networks adapt conventional CNNs to graph data. They can be thought of as implicitly learning an embedding, by taking the output of the last layer before the supervised component. Works exploring different geometries include spherical space \cite{lei2019sphericalKernelforGCN, defferrard2020DeepSphere, yang2020sphericalgcn}, hyperbolic space \cite{chami2019hgcnn, liu2019hypergraphsnn}, and combinations of different curvatures \cite{bachmann2020ccgcn, ye2020CurvatureGCN, pei2020GeomGCN}).
%     \item \citet{pei2020GeomGCN} performs the aggregation of node embeddings over a meaningful space (hyperbolic or Euclidean).
%     \item Adapts to the curvature of the graph in different regions \cite{ye2020CurvatureGCN}.
% \end{itemize}

%\todo{B: Add Finsler at the beginning}{Geometries} induced by different metrics have been studied in the context of kernel methods \cite{kloft2009lpNormKernelLearning, scetbon2019elloneGeometry}. 
Our general view allows us to to endow Riemannian symmetric spaces with Finsler metrics, which have been applied 
%Versions of these metrics on the Euclidean spaces have been very successfully applied in compressed sensing \cite{Donoho:2008aa}.
in compressed sensing \cite{Donoho:2008aa}, for clustering categorical distributions \cite{Nielsen2019clusteringWithFinsler}, and in robotics \cite{ratliff2020finslerRobotics}.
We provide strong experimental evidence that supports the intuition on how they offer a less distorted representation than Euclidean metrics for graphs with different structure. 
With regard to optimization, we derive the explicit formulations to employ a generalization of stochastic gradient descent \cite{bonnabel2011rsgd} as a Riemannian adaptive optimization method \cite{becigneul2019riemannianMethods}.

\begin{figure}[!b]
    \vspace{-4mm}
     \centering
     \includegraphics[width=0.85\linewidth]{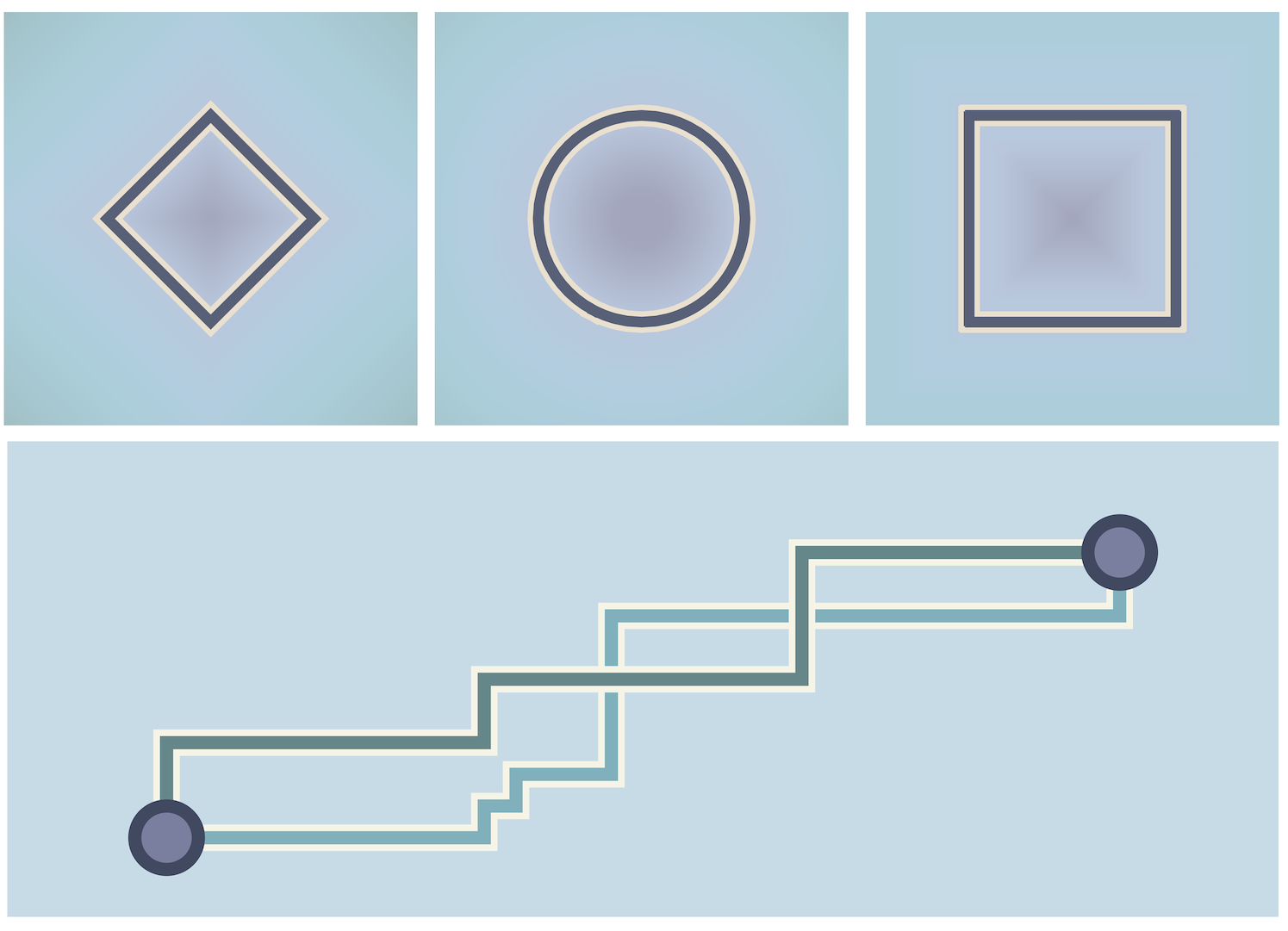}
     \caption{Above, from left to right: the unit spheres for the $\ell^1$, $\ell^2$ (Euclidean), and $\ell^\infty$ metrics on the plane.
   Below: Distance minimizing geodesics are not necessarily unique in Finsler geometry. The two paths shown have the same (minimal) $\ell^1$ length.}
     \label{fig:Finsler}
   %  \vspace{-5mm}
\end{figure}
 
\section{Symmetric Spaces for Embedding Problems}
\label{sec:symmetric-spaces-for-embeddings}
%%%%%%%%%%%%%
%We propose the systematic use of Symmetric spaces in representation learning. Symmetric spaces are Riemannian manifolds with rich symmetry groups. This makes them amenable to analysis and explicit computations. 

% Riemannian symmetric spaces (RSS) are Riemannian mani\-folds with rich symmetry groups, which makes them amenable to analytical tools as well as to explicit computations. This rich class, that includes many of the spaces previously applied in representation learning, has been extensively studied by mathematicians, who established many general tools that can be used for explicit algorithmic implementation. We refer to the appendix for a general introduction to the theory, the duality between compact and non-compact RSS, general algorithms to compute the distance function in these spaces, and formulas for computing the exponential map, curvature and parallel transport. These unify the discussions for Euclidean space, hyperbolic spaces, spherical spaces, Grassmannian manifolds, and SPD already available in the representation learning literature.

% Key features of (non-compact) RSS are that they offer a rich combination of Euclidean and hyperbolic geometry, without necessarily separating it in different factors. Furthermore, they have many subspaces isometric to Euclidean, hyperbolic spaces and products thereof. This makes them an excellent target for learning embeddings of complex networks without a priori knowledge of their internal structure.

Riemannian symmetric spaces (RSS) are Riemannian mani\-folds with large symmetry groups, which makes them amenable to analytical tools as well as to explicit computations.  A key feature of (non-compact) RSS is that they offer a rich combination of geometric features, including many subspaces isometric to Euclidean, hyperbolic spaces and products thereof. This makes them an excellent target tool for learning embeddings of complex networks without a priori knowledge of their internal structure.

First, we introduce two aspects of the general theory of RSS to representation learning: Finsler distances and vector-valued distances. These give us, respectively, a concrete method to obtain better graph representations, and a new tool to analyze graph embeddings. Then, we describe our general implementation framework for RSS.

{\bf Finsler Distances:} 
Riemannian metrics are not well adapted to represent graphs. For example, though a two dimensional grid intuitively looks like a plane, any embedding of it in the Euclidean plane $\R^2$ necessarily distorts some distances by a factor of at least $\sqrt 2$. This is due to the fact that while in the Euclidean plane length minimizing paths (geodesics) are unique, in graphs there are generally several shortest paths (see Figure \ref{fig:Finsler}).  Instead, it is possible to find an abstract isometric embedding of the grid in $\R^2$ if the latter is endowed with the $\ell^1$ (or taxicab) metric. This is a first example of a Finsler distance. Another Finsler distance on $\R^n$ that plays a role in our work is the $\ell^\infty$ metric. See Appendix \ref{a.finsler} for a brief introduction.

RSS do not only support a Riemannian metric, but a whole family of Finsler distances with the same symmetry group (group of isometries). 
For the reasons explained above, these Finsler metrics are more suitable to embed complex networks. We verify these assumptions through concrete experiments in Section \ref{s.graphrec}.
Since Finsler metrics are in general not convex, they are less suitable for optimization problems. Due to this, we propose to combine the Riemannian and Finsler structure, by using a Riemannian optimization scheme, with loss functions based on the Finsler metric.

%One reason for this might be their different geometric properties. 
%\todo{this intuition is GOLD: stress it}{
%For example, between two points in a symmetric space, there is always a unique shortest path (geodesic) with respect to the Riemannian metric, but with respect to a Finsler metric, there might be several shortest paths. 
%This is a feature which we can also often observe in graphs, for example in a grid, or in the product of two trees.}

% \begin{minipage}{\linewidth}
% \vspace{-2mm}
% 	\begin{toolkit}[!b]
% 		\caption{Choosing an RSS}
% 		\label{alg:choosing-rss}
% 		\begin{algorithmic}[1]
% 			\small
% 			\STATE RSS are completely classified: all are built from a list of elemental buiding blocks, called \emph{irreducible RSS}.
% 			\STATE There are 11 infinite families of irreducible RSS, see Table \ref{tab.1}.  Mathematical properties of each are well-documented, see for example \citet{helgason1078diffGeom}.
% 			\STATE A {\bf model} of an RSS is a choice of manifold $M$ representing its points and an action of its symmetry group $G$. %Many models are well explored, see for example \citet{helgason1078diffGeom}. 
% 		    \STATE {\bf To implement an irreducible RSS}, select a family based on desired mathematical properties, and construct a model based on computational tractability.  
% 		   \STATE {\bf To implement a product RSS}, select a collection of irreducible spaces $\{M_i\}$, and take the product $M=\prod M_i$.
% 		\end{algorithmic}
% 	\end{toolkit}
% \end{minipage}%

{\bf Vector-valued Distance:}
In Euclidean space, in the sphere or in hyperbolic space, the only invariant of two points is their distance. A pair of points can be mapped to any other pair of points iff their \textit{distance} is the same. Instead, in a general RSS the invariant between two points is a {\em distance vector} in $\R^n$, where $n$ is the rank of the RSS. %This distance vector can always be represented by a diagonal matrix with real non-zero entries. 
% The distance between two points with respect to the Riemannian metric or any Finsler metric can be directly read off from the distance vector. 
This is, two pairs of points can be separated by the same distance, but have different \textit{distance vectors}.
This vector-valued distance gives us a new tool to analyze graph embeddings, as we illustrate in Section~\ref{sec:vectorialdistance}.

{
The dimension of the space in which the vector-valued distance takes values in defines the \emph{rank} of the RSS.
Geometrically, this represents the largest Euclidean subspace which can be isometrically embedded (hence, hyperbolic and spherical spaces are of $\operatorname{rank} - 1$).
The symmetries of an RSS fixing such a \emph{maximal flat} form a finite group --- the Weyl group of the RSS. 
In the example of  Siegel spaces discussed below, the Weyl group acts by permutations and reflections of the coordinates, allowing us to canonically represent each vector-valued distance as an $n$-tuple of non-increasing positive numbers.
Such a uniform choice of standard representative for all vector-valued distances is a fundamental domain for this group action, known as a \emph{Weyl chamber} for the RSS.
}
%{\bf Combining Riemannian and Finsler structure}
%\todo{there are 3 paragraphs starting with "Finsler metrics". Everything should be in one.}{Finsler metrics} are in general not convex, and less suitable for optimization problems. For this reason we combine the Riemannian and Finsler structure, by using a Riemannian optimization scheme, but with a loss function involving the Finsler metric. 

%Whereas the Riemannian metric is uniquely geodesic, i.e. between any two points there exists a shortest path, the Finsler metrics are in general not uniquely geodesic, and between two points there are many shortest paths with respect to a chosen Finsler metric (see Figure~\ref{fig:Finsler}). 
%This is one reason why their representation capabilities are better.
%Already in rather simple graphs such as a grid, or the product of two trees there are in general several shortest paths between two nodes. 

{\bf Implementation Schema:}
The general theory of RSS not only unifies many spaces previously applied in representation learning, but also systematises their implementation. 
{
Using standard tools of this theory, we provide a general framework to  implement the mathematical methods required to learn graph embeddings in a given RSS.
}

{
\textbf{Step 1, choosing an RSS:} We may utilize the classical theory of symmetric spaces to inform our choice of RSS.
Every symmetric space $M$ can be decomposed into an (almost) product $M = M_1 \times \cdots \times M_k$ of irreducible symmetric spaces. 
Apart from twelve exceptional examples, there are eleven infinite families irreducible symmetric spaces --- 
see \citet{helgason1078diffGeom} for more details, or Appendix~\ref{app:appendix-a}, Table \ref{tab.1}.
Each family of irreducible symmetric space has a distinct family of symmetry groups, which in turn determines many mathematical properties of interest (for instance, the symmetry group determines the shape of the Weyl chambers, which determines the admissible Finsler metrics).
Given a geometric property of interest,  the theory of RSS allows one to  determine which (if any) symmetric spaces enjoy it.
For example, we choose Siegel spaces also because they admit Finsler metrics induced by the $\ell^1$ metric on flats, which agrees with the intrinsic metric on grid-like graphs.
}

% \begin{minipage}{\linewidth}
% \vspace{-2mm}
	\begin{toolkit}[!t]
		\caption{Computing Distances}
		\label{alg:distances}
		\begin{algorithmic}[1]
			\small
			\STATE {\bf Input from Model:} Choice of basepoint $m$, maximal flat $F$, identification $\phi\colon F\to\R^n$, choice of Weyl Chamber $C\subset\R^n$, and Finsler norm $\|\cdot\|_F$ on $\R^n$.  
			\STATE Given $p,q\in M$:
			\STATE Compute $g\in G$ such that $g(p)=m$ and $g(q)\in F$.
			\STATE Compute $v'=\phi(g(q))\in\R^n$, and $h\in G$ the Weyl group element such that $h(v')=v\in C$.
			\STATE The {\bf Vector-valued Distance (VVD)} is $\mathrm{vDist}(p,q)=v$.
			\STATE The {\bf Riemannian Distance (RD)} is  $d^R(p,q)=\sqrt{\sum_i v_i^2}.$
			\STATE The {\bf Finsler Distance (FD)} is $d^F(p,q)=\|v\|_F.$
			\STATE {\bf For a product $\prod M_i$}, the VVD is the vector $\left(\mathrm{vDist}(p_i,q_i)\right)$ of VVDs for each $M_i$. The RD, FD satisfy the pythagorean theorem: $d^X(p,q)^2=\sum_i d^{X_i}(p_i,q_i)^2$, for $X\in\{R,F\}$.
		\end{algorithmic}
	\end{toolkit}
% \end{minipage}%

{
\textbf{Step 2, choosing a model of the RSS:} Having selected an RSS, we must also select a model: a space $M$ representing its points equipped with an action of its symmetry group $G$.
Such a choice is of practical, rather than theoretical concern: the points of $M$ should be easy to work with, and the symmetries of $G$ straightforward to compute and apply.
Each RSS may have many already-understood models in the literature to select from.
In our example of Siegel spaces, we implement two distinct models, selected because both their points and symmetries may be encoded by $n\times n$ matrices. See Section \ref{sec:Siegel}.
}

{
Implementing a product of symmetric spaces requires implementing each factor simultaneously.  Given models $M_1,\ldots, M_k$ with symmetry groups $G_1,\ldots G_k$, the product $M=M_1\times \cdots \times M_k$ has as its points $m=(m_1,\ldots, m_k)$ the $k-$tuples with $m_i\in M_i$, with the group $G=G_1\times\cdots\times G_k$ acting componentwise.
This general implementation of products directly generalizes products of constant curvature spaces.
}

%(see Toolkit~\ref{alg:choosing-rss}).
{
\textbf{Step 3, computing distances:} Given a choice of RSS, the fundamental quantity to compute is a distance function on $M$, typically used in the loss function.
In contrast to general Riemannian manifolds, the rich symmetry of RSS allows this computation to be factored into a sequence of geometric steps.
See Toolkit~\ref{alg:distances} for a schematic implementation using data from the standard theory of RSS (choice of maximal flat, Weyl chamber, and Finsler norm) and Algorithm~\ref{alg:distances} for a concrete implementation in the Siegel spaces.
}
%{\color{red}
%All formulas in this implementation schema are given in terms of data for %this chosen model readily computable from the general theory.}

{\textbf{Step 4, computing gradients:} To perform gradient-based optimization, the Riemannian gradient of these distance functions is required.
Depending on the Riemannian optimization methods used, additional local geometry including parallel transport and the exponential map may be useful \cite{bonnabel2011rsgd, becigneul2019riemannianMethods}. See Toolkit~\ref{alg:backprop} for the relationships of these components to elements of the classical theory of RSS.
}

% \begin{minipage}{\linewidth}
% \vspace{-2mm}
	\begin{toolkit}[!t]
		\caption{Computing Local Geometry}
		\label{alg:backprop}
		\begin{algorithmic}[1]
			\small
	\STATE {\bf Input From Model:} Geodesic reflections $\sigma_p\in G$, the metric tensor $\langle\cdot,\cdot\rangle$, basepoint $m\in M$, orthogonal decomposition $\mathfrak{stab}(m)\oplus\mathfrak{p}=\mathfrak{g}$, and identification $\phi\colon T_mM\to\mathfrak{p}.$
	\STATE Given $f\colon M\to\R$, a geodesic $\gamma$, or $v\in T_mM$ respectively:
	\STATE The {\bf Riemannian Gradient} of $f$ is computed from the metric tensor by solving $\langle {\rm grad}_R(f),-\rangle=df(-)$
	\STATE {\bf Parallel Transport} along $\gamma$ is achieved by the differentials $(d\tau_t)_{\gamma(t_0)}$ of transvections $\tau_t=\sigma_{\gamma(t/2)}\sigma_{\gamma(t_0)}$ along $\gamma$.
	\STATE The {\bf Riemannian Exponential} $\exp_m^R(v)=g(m)$ is the matrix exponential $g=\exp(\phi(v))\in G$ applied to $m$.
	\STATE {\bf For a product $\prod M_i$} the Riemannian gradient, Parallel Transport, and Exponential map are computed component-wise.
		\end{algorithmic}
	\end{toolkit}
% \end{minipage}%

See Appendix~\ref{app:appendix-a} and \ref{a.Siegel} for a review of the general theory relevant to this schema, and for an explicit implementation in the Siegel spaces.

\begin{figure*}[!t]
\vspace{-4mm}
    \centering 
    \subfloat[Bounded Domain Model $\calB_n$\label{fig:BddDom}]{\includegraphics[width=0.4\textwidth]{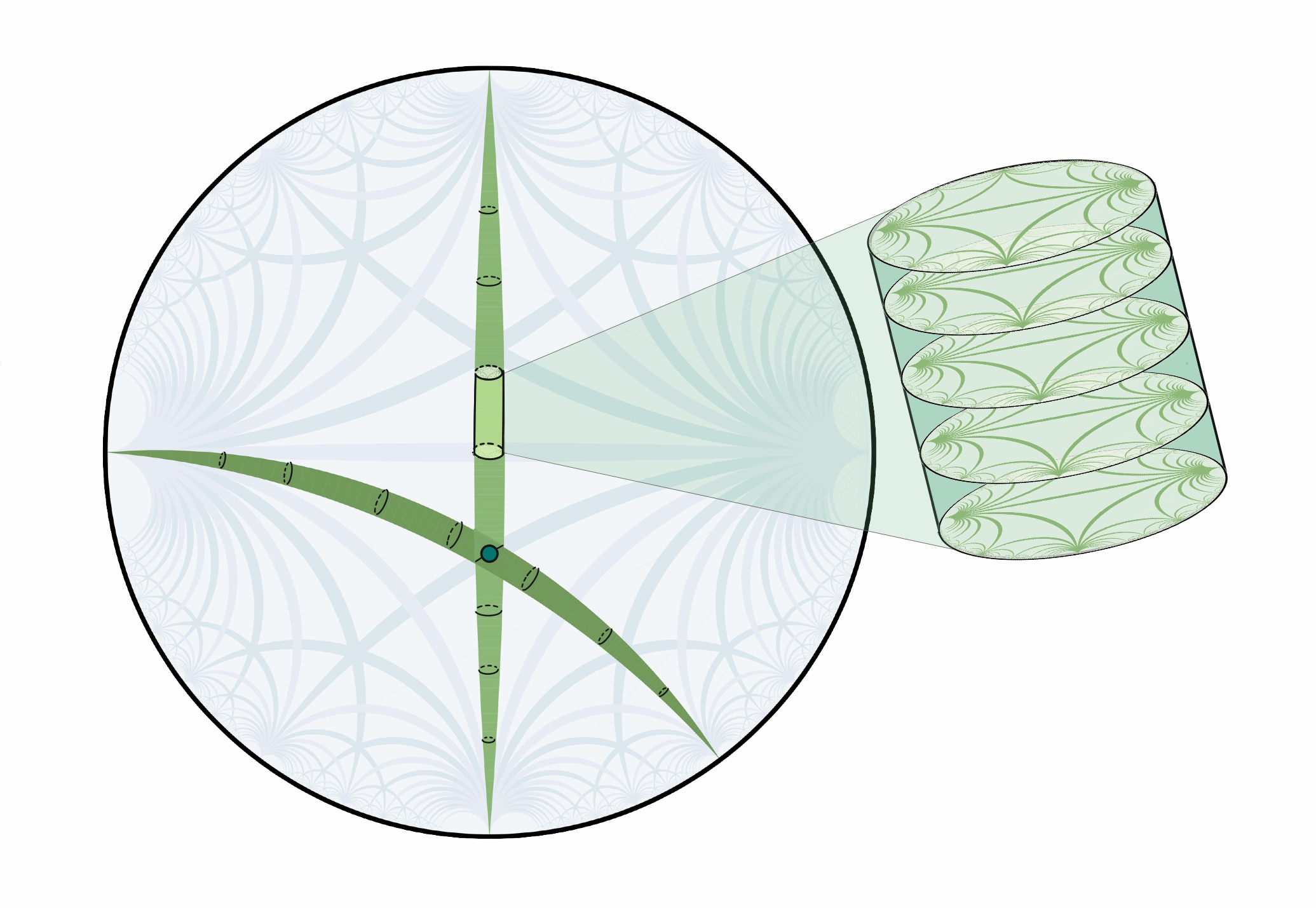}}
    \hspace{20mm}
    \subfloat[Siegel Upper Half Space $\mathcal S_n$\label{fig:UHS}]{\includegraphics[width=0.35\textwidth]{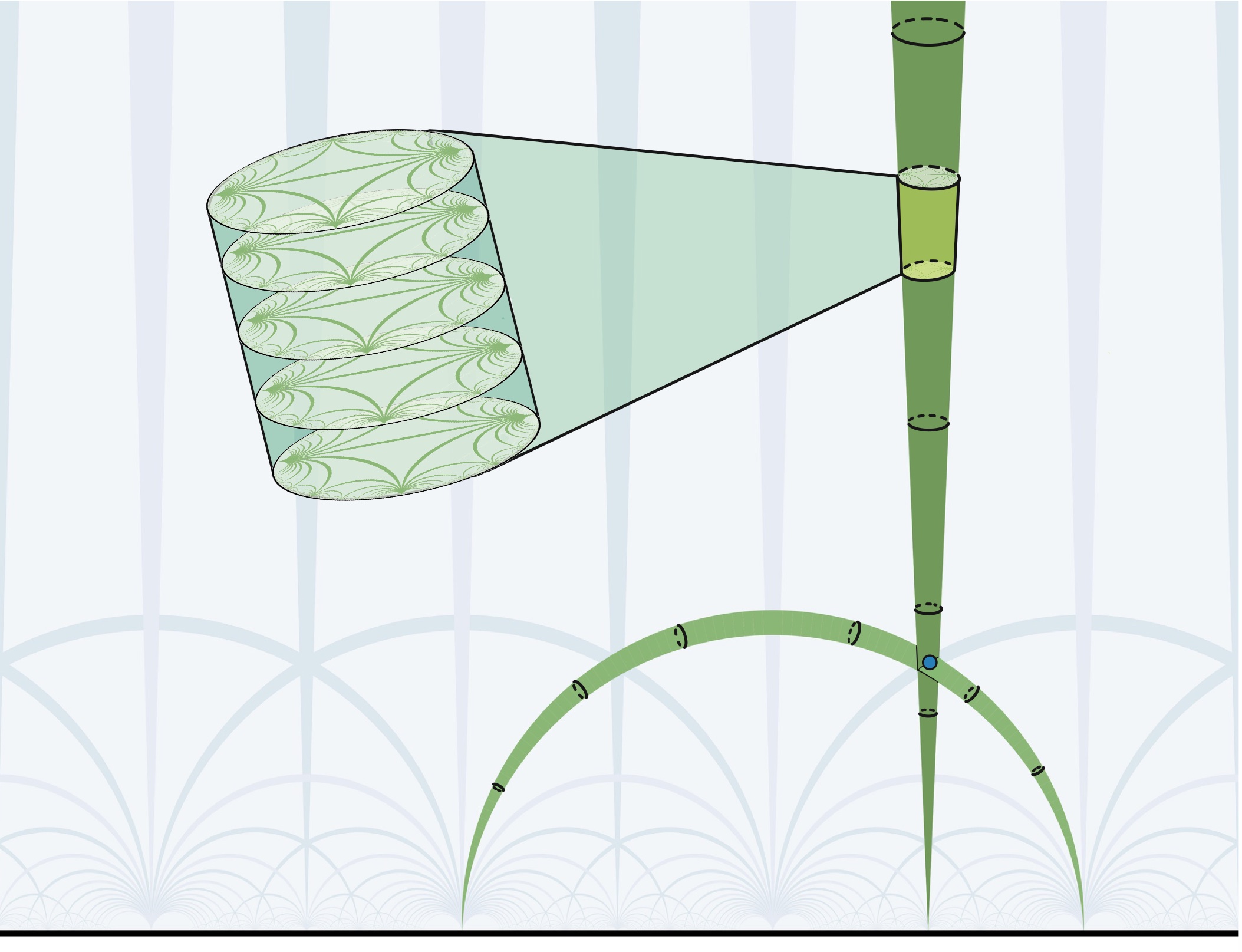}}
    \vspace{-2mm}
    \caption{a) Every point of the disk is a complex symmetric $n$-dimensional matrix. b) A hyperbolic plane over SPD. $\mathcal S_2$  is a 6 dimensional manifold, the green lines represent totally geodesic submanifolds isometric to SPD that intersect in exactly one point. In dimension 2, SPD is isometric to the product of a hyperbolic plane and the line}
    \label{fig:model-diagrams}
    \vspace{-3mm}
\end{figure*}

\section{Siegel Space}
\label{sec:Siegel}
We implement the general aspects of the theory of RSS outlined above in the Siegel spaces HypSPD$_n$ \cite{siegel1943symplectic}, a versatile family of non-compact RSS, which has not yet been explored in geometric deep learning. 
The simplicity and the versatility of the Siegel space make it particularly suited for representation learning. We highlight some of its main features.

 {\bf Models:}
HypSPD$_n$ admits concrete and tractable matrix models generalizing the Poincar\'e disk and the upper half plane model of the hyperbolic space. Both are open subsets of the space $\Sym(n,\C)$ of symmetric $n\times n$-matrices over $\C$. HypSPD$_n$ has $n(n+1)$ dimensions.
 
The bounded symmetric domain model for HypSPD$_n$  generalizes the Poincar\'e disk. It is given by:\footnote{For a real symmetric matrix $Y \in \Sym(n,\R)$ we write $Y>\!\!>0$ to indicate that $Y$ is positive definite.}
\vspace{0mm}
\begin{equation}
    \calB_n:=\{Z\in\Sym(n,\C)|\;\Id-Z^*Z>\!\!>0\};
\vspace{0mm}
\end{equation}
The Siegel upper half space model for HypSPD$_n$ generalizes the upper half plane model of the hyperbolic plane by:
\vspace{0mm}
\begin{equation}
\calS_n:=\{Z= X+iY\in\Sym(n,\C)|\;Y>\!\!>0\}.
\vspace{0mm}
\end{equation}
% Diagrams for both models can be seen in Figure~\ref{fig:model-diagrams}. %\ref{fig:BddDom} and \ref{fig:UHS}. 
{An explicit isomorphism from $\calB_n$ to $\calS_n$ is given by the Cayley transform, a matrix analogue of the familiar map from the Poincare disk to upper half space model of the hyperbolic plane:
$$
Z\mapsto i(Z+\Id)(Z-\Id)^{-1}.
$$}
{\bf Hyperbolic Plane over SPD:}
The Siegel space HypSPD$_n$ contains SPD$_n$ as a totally geodesic submanifold, and in fact, it can be considered as a hyperbolic plane over SPD. The role that real lines play in the hyperbolic plane, in HypSPD$_n$ is played by SPD$_n$. This is illustrated in Figure~\ref{fig:UHS}. %, in particular when it is endowed with the Finsler metric inducing the $\ell^1$ metric. 

%This additional structure can be leveraged when embedding data based on symmetric positive definite matrices, which has for example gained popularity in
%visual learning applications\todo{MS: what is this? also cite} \cite{}. 
%\todo[inline]{MS: Very nice but not very helpful, AW: better? } 

%  \begin{figure}[t]
%      \centering
%      \includegraphics[width=0.4\textwidth]{img/SiegelUHS.jpg}
%      \caption{Siegel Upper Half Space, an hyperbolic plane over SPD. $\mathcal S_2$  is a 6 dimensional manifold, the green lines represent totally geodesic submanifolds isometric to SPD that intersect in exactly one point. In dimension 2, SPD is isometric to the product of a hyperbolic plane and the line}
%      \label{fig:UHS}
%      \vspace{-4mm}
%  \end{figure}

%{\bf Matrix Models} 
%The Siegel space $\mathcal{H}_n$ is a  generalization of the hyperbolic plane which is $\mathcal{H}_1$.
%It admits matrix models of the Poincar\'e disc and the upper half-plane model of hyperbolic space in the space $\Sym(n,\C)$ of symmetric $n\times n$-matrices over $\C$. The dimension of $\mathcal{H}_n$ is $n(n+1)$. 
% The matrix models give rise to simple expressions for the gradient of a function, see Section~\ref{} 

{\bf Totally Geodesic Subspaces:}
The Siegel space HypSPD$_n$ contains $n$-dimensional Euclidean subspaces, products of $n$-copies of hyperbolic planes, SPD$_n$ as well as products of Euclidean and hyperbolic spaces as totally geodesic subspaces (see Figure~\ref{fig:model-diagrams}).
It thus has a richer pattern of submanifolds than, for example, SPD. In particular, HypSPD$_n$ contains more products of hyperbolic planes than SPD$_n$: in HypSPD$_n$ we need 6 real dimension to contain $\HH^2 \times \HH^2$ and 12 real dimension to contain $(\HH^2)^3$, whereas in SPD$_n$ we would need 9 (resp. 20) dimensions for this.

{\bf Finsler Metrics:} 
The Siegel space supports a Finsler metric $F_1$ that induces the $\ell^1$ metric on the Euclidean subspaces. As already remarked, the $\ell^1$ metric is particularly suitable for representing product graphs, or graphs that contain product subgraphs. Among all possible Finsler metrics supported by HypSPD$_n$, we focus  on  $F_1$ and $F_\infty$ (the latter induces the $\ell^\infty$ metric on the flat).

{\bf Scalability:} Like all RSS, HypSPD$_n$ has a \emph{dual} -- an  RSS with similar mathematical properties but reversed curvature -- generalizing the duality of $\HH^2$ and $\mathbb{S}^2$.
%is one member of a pair of RSS with similar mathematical properties, generalizing the relationship between $\HH^2$ and $S^2$.
We focus on HypSPD$_n$ over its dual for scalability reasons. 
The dual is a nonnegatively curved RSS of finite diameter, and thus does not admit isometric embeddings of arbitrarily large graphs.
HypSPD$_n$, being nonpositively curved and infinite diameter, does not suffer from this restriction.
 See Appendix \ref{a.cpt} for details on its implementation and experiments with the dual.

\section{Implementation}
\label{sec:implementation-of-matrix-models}
%{\bf Matrix models:} Siegel spaces generalize the hyperbolic plane which is $\mathcal{H}_1$. They admit a bounded domain model and an upper half-space model in the space $\Sym(n,\C)$ of $n \times n$ symmetric matrices with complex entries. When $n=1$, these are the Poincar\'e model and the upper half-space model of the hyperbolic plane. The dimension of the space $\mathcal{H}_n$ is $n(n+1)$, and it contains $n$-dimensional Euclidean subspaces, products of $n$-copies of hyperbolic planes, as well as products of Euclidean and hyperbolic spaces. 

A complex number $z\in\C$ can be written as $z=x+iy$ where $x,y\in\R$ and $i^2=-1$. Analogously a complex symmetric matrix $Z\in \Sym(n,\C)$ can be written as $Z = X+ iY$, where $X=\Re(Z), Y=\Im(Z) \in \Sym(n,\R)$ are symmetric matrices with real entries. 
We denote by $Z^*=X-iY$ the complex conjugate matrix.  

%\begin{figure}
%    \centering
%    \begin{tikzpicture}[scale=0.5]
    %bounded domain	
%    \fill [fill=blue!40!white] (0,0) circle [radius= 1cm];
%    \draw [thick] (0,0) circle [radius= 1cm];
%    \filldraw (0,0)  circle [radius= 1pt] node [above right]{$0$};    
%    \node at (1.1,1.4) {$\calB_1$};
%    \node at (3,.5) {Cayley};
%    \node at (3,0) {$\leftrightsquigarrow$};
%    
%    %siegel space
%    \fill[fill=blue!40!white] (5,-1) rectangle (8,1);
%    \draw [thick] (5,-1) to (8.5,-1);    
%    \filldraw (6.5,0)  circle [radius= 1pt] node [above right]{$i$};    
%    \node at (8.2,1.4) {$\calS_1$};
%    \draw (6.5, -1) to (6.5, 1.5);
%    \node at (5.75,1.5) {$\Im z$};
%    \node at (8.5,-1.5) {$\Re z$};
%    \node at (6.52, 1.25) [rotate=90] {$>$};
%    \node at (8.25, -1.05) {$>$};
%    \end{tikzpicture}
%    %\vspace{2mm}
%    \caption{Matrix models.}
%    \label{fig:spaces}
%\end{figure}

%\begin{minipage}{\linewidth}
 %\vspace{-2mm}
	\begin{algorithm}[!b]
		\caption{Computing Distances}
		\label{alg:Vdistances}
		\begin{algorithmic}[1]
			\small
		    \STATE Given two points $Z_1,Z_2\in \calS_n$:
			\STATE Define $Z_3=\sqrt{\Im(Z_1)}^{-1}(Z_2-\Re(Z_1))\sqrt{\Im(Z_1)}^{-1}\in\calS_n$
			\STATE Define $W=(Z_3 - i\Id) (Z_3 + i\Id)^{-1}\in\calB_n$
			\STATE Use the Takagi factorization to write
			$W=\ov K D K^*$ for $D$ real diagonal, and $K$ unitary.
			\STATE Define $v_i=\log\frac{1 + d_i}{1 - d_i}$ for $d_i$ the diagonal entries of $D$.
			\STATE Order the $v_i$ so that $v_1\geq v_2\geq\cdots\geq 0$.  The {\bf Vector-valued Distance} is
			$\mathrm{vDist}(Z_1,Z_2)=(v_1,v_2,\ldots,v_n)$.
						\STATE The {\bf Riemannian distance} is 
			$d^R(Z_1,Z_2):=\sqrt {\sum_{i=1}^n v_i^2}$.
			\STATE The {\bf Finsler distance inducing the $\ell^1$-metric} is $d^{F1}(Z_1,Z_2):=\sum_{i=1}^n v_i.$ 
		 \STATE The {\bf Finsler distance inducing the $\ell^\infty$-metric} is $d^{F\infty}(Z_1,Z_2):= \max \{v_i\}=v_1$.
		\end{algorithmic}
	\end{algorithm}
 %\end{minipage}%

{\bf Distance Functions:} To compute distances we  apply either Riemannian or Finsler distance functions to the vector-valued distance. These computations are described in  Algorithm~\ref{alg:Vdistances}, which is a concrete implementation of Toolkit~\ref{alg:distances}. 
{
Specifically, step 2 moves one point to the basepoint, step 4 moves the other into our chosen flat, step 5 identifies this with $\mathbb{R}^n$ and step 6 returns the vector-valued distance, from which all distances are computed.
}
We employ the Takagi factorization to obtain eigenvalues and eigenvectors of complex symmetric matrices in a tractable manner with automatic differentiation tools (see Appendix \ref{a.Takagi}).

\textbf{Complexity of Distance Algorithm:}
%\todo{ST: Maybe remove the heading here, as right now this is at the same "level" as the distance functions and Riemannian optimization.  Would make more sense if this was "part" of the distance section?. F: explicit is better than implicit }
\label{sec:complexity-of-distance}
Calculating distance between two points $Z_1,Z_2$ in either $\calS_n$ or $\calB_n$ spaces implies computing multiplications, inversions and diagonalizations of $n \times n$ matrices. We find that the cost of the distance computation with respect to the matrix dimensions is $\bigO(n^3)$. We prove this in Appendix~\ref{app:dist-alg-complexity}.

{\bf Riemannian Optimization with Finsler Distances:} With the proposed matrix models of the Siegel space, we optimize objectives based on the Riemannian or Finsler distance functions in the embeddings space. To overcome the lack of convexity of Finsler metrics, we combine the Riemannian and the Finsler structure, by using a Riemannian optimization scheme \cite{bonnabel2011rsgd} with a loss function based on the Finsler metric.
%To do so, we apply Riemannian optimization \cite{bonnabel2011rsgd}. 
In Algorithm~\ref{alg:riem-grad} we provide a way to compute the Riemannian gradient from the Euclidean gradient obtained via automatic differentiation.
{
This is a direct implementation of Toolkit~\ref{alg:backprop} Item 3.
}

{
To constrain the embeddings to remain within the Siegel space, we utilize a projection from the ambient space to our model.
More precisely, given $\epsilon$ and a point $Z\in\Sym(n,\C)$, we compute a point $Z_\epsilon^\calS$ (resp. $Z_\epsilon^\calB$) close to the original point lying in the $\epsilon$-interior of the model.
For $\calS_n$, starting from $Z=X+iY$ we orthogonally diagonalize $Y=K^tDK$, and then modify $D=\operatorname{diag}(d_i)$ by setting each diagonal entry to $\max\{d_i,\epsilon\}$.
An analogous projection is defined on the bounded domain $\calB_n$, see Appendix~\ref{a.project}. 
}
 
% \begin{minipage}{\linewidth}
% \vspace{-2mm}
	\begin{algorithm}[H]
		\caption{Computing Riemannian Gradient}
		\label{alg:riem-grad}
		\begin{algorithmic}[1]
			\small
		%	\STATE Given a function $f: \calS_n \to \R$ and a point $Z = X+iY \in \calS_n \subset \Sym(n,\C)$:
			\STATE Given $f: \calS_n \to \R$ and $Z = X+iY \in \calS_n$:
		\STATE Compute the Euclidean gradient ${\rm grad}_E(f)$ at $Z$ of $f$ obtained via automatic differentiation (see Appendix \ref{a.gradient}). 
		\STATE {\bf The Riemannian gradient} is ${\rm grad_R}(f) = Y \cdot {\rm grad_E}(f) \cdot Y$.
		\end{algorithmic}
	\end{algorithm}

\begin{table*}[!t]
\vspace{-1mm}
\small
\centering
\adjustbox{max width=\textwidth}{
\begin{tabular}{crrrrrrrrrrrr}
\toprule
 & \multicolumn{2}{c}{\textsc{4D Grid}} & \multicolumn{2}{c}{\textsc{Tree}} & \multicolumn{2}{c}{\textsc{Tree $\times$ Grid}} & \multicolumn{2}{c}{\textsc{Tree $\times$ Tree}} & \multicolumn{2}{c}{\textsc{Tree $\diamond$ Grids}} & \multicolumn{2}{c}{\textsc{Grid $\diamond$ Trees}} \\
$(|V|, |E|)$ & \multicolumn{2}{c}{$(625, 2000)$} & \multicolumn{2}{c}{$(364, 363)$} & \multicolumn{2}{c}{$(496, 1224)$} & \multicolumn{2}{c}{$(225, 420)$} & \multicolumn{2}{c}{$(775, 1270)$} & \multicolumn{2}{c}{$(775, 790)$} \\
 & \multicolumn{1}{c}{$D_{avg}$} & mAP & \multicolumn{1}{c}{$D_{avg}$} & mAP & \multicolumn{1}{c}{$D_{avg}$} & mAP & \multicolumn{1}{c}{$D_{avg}$} & mAP & \multicolumn{1}{c}{$D_{avg}$} & mAP & \multicolumn{1}{c}{$D_{avg}$} & mAP \\
 \cmidrule(lr){2-3}\cmidrule(lr){4-5}\cmidrule(lr){6-7}\cmidrule(lr){8-9}\cmidrule(lr){10-11}\cmidrule(lr){12-13}
$\mathbb{E}^{20}$ & 11.24$\pm$0.00 & \textbf{100.00} & 3.92$\pm$0.04 & 42.30 & 9.81$\pm$0.00 & 83.32 & 9.78$\pm$0.00 & 96.03 & 3.86$\pm$0.02 & 34.21 & 4.28$\pm$0.04 & 27.50 \\
$\mathbb{H}^{20}$ & 25.23$\pm$0.05 & 63.74 & \textbf{0.54$\pm$0.02} & \textbf{100.00} & 17.21$\pm$0.21 & 83.16 & 20.59$\pm$0.11 & 75.67 & 14.56$\pm$0.27 & 44.14 & 14.62$\pm$0.13 & 30.28 \\
$\mathbb{E}^{10} \times \mathbb{H}^{10}$ & 11.24$\pm$0.00 & \textbf{100.00} & 1.19$\pm$0.04 & \textbf{100.00} & 9.20$\pm$0.01 & \textbf{100.00} & 9.30$\pm$0.04 & 98.03 & 2.15$\pm$0.05 & 58.23 & 2.03$\pm$0.01 & \textbf{97.88} \\
$\mathbb{H}^{10} \times \mathbb{H}^{10}$ & 18.74$\pm$0.01 & 78.47 & 0.65$\pm$0.02 & \textbf{100.00} & 13.02$\pm$0.91 & 88.01 & 8.61$\pm$0.03 & 97.63 & 1.08$\pm$0.06 & 77.20 & 2.80$\pm$0.65 & 84.88 \\
$\spd_{6}$ & 11.24$\pm$0.00 & \textbf{100.00} & 1.79$\pm$0.02 & 55.92 & 9.23$\pm$0.01 & 99.73 & 8.83$\pm$0.01 & 98.49 & 1.56$\pm$0.02 & 62.31 & 1.83$\pm$0.00 & 72.17 \\
$\mathcal S_{4}^R$ & 11.27$\pm$0.01 & \textbf{100.00} & 1.35$\pm$0.02 & 78.53 & 9.13$\pm$0.01 & 99.92 & 8.68$\pm$0.02 & 98.03 & 1.45$\pm$0.09 & 72.49 & 1.54$\pm$0.08 & 76.66 \\
$\mathcal S_{4}^{F_{\infty}}$ & 5.92$\pm$0.06 & 99.61 & 1.23$\pm$0.28 & 99.56 & 4.81$\pm$0.55 & 99.28 & 3.31$\pm$0.06 & 99.95 & 10.88$\pm$0.19 & 63.52 & 10.48$\pm$0.21 & 72.53 \\
$\mathcal S_{4}^{F_{1}}$ & \textbf{0.01$\pm$0.00} & \textbf{100.00} & 0.76$\pm$0.02 & 91.57 & \textbf{0.81$\pm$0.08} & \textbf{100.00} & \textbf{1.08$\pm$0.16} & \textbf{100.00} & \textbf{1.03$\pm$0.00} & \textbf{78.71} & \textbf{0.84$\pm$0.06} & 80.52 \\
$\mathcal B_{4}^R$ & 11.28$\pm$0.01 & \textbf{100.00} & 1.27$\pm$0.05 & 74.77 & 9.24$\pm$0.13 & 99.22 & 8.74$\pm$0.09 & 98.12 & 2.88$\pm$0.32 & 72.55 & 2.76$\pm$0.11 & 96.29 \\
$\mathcal B_{4}^{F_{\infty}}$ & 7.32$\pm$0.16 & 97.92 & 1.51$\pm$0.13 & 99.73 & 8.70$\pm$0.87 & 96.40 & 4.26$\pm$0.26 & 99.70 & 6.55$\pm$1.77 & 73.80 & 7.15$\pm$0.85 & 90.51 \\
$\mathcal B_{4}^{F_{1}}$ & 0.39$\pm$0.02 & \textbf{100.00} & 0.77$\pm$0.02 & 94.64 & 0.90$\pm$0.08 & \textbf{100.00} & 1.28$\pm$0.16 & \textbf{100.00} & 1.09$\pm$0.03 & 76.55 & 0.99$\pm$0.01 & 81.82 \\
\bottomrule
\end{tabular}
}
\caption{Results for synthetic datasets. Lower $D_{avg}$ is better. Higher mAP is better. Metrics are given as percentage.}
\label{tab:synthetic-results}
\vspace{-3mm}
\end{table*}

%%%%%%%%%%%%%%%%%%%%%%% EXPs %%%%%%%%%%%%%%%%%%%%%%%%%%%%%
%\section{Experiments}

%We evaluate the representation capabilities of the proposed approach for graph reconstruction, and two downstream tasks: recommender systems and node classification.
\section{Graph Reconstruction}
\label{s.graphrec}
We evaluate the representation capabilities of the proposed approach for the task of graph reconstruction.\footnote{Code available at \url{https://github.com/fedelopez77/sympa}.}

%\subsection{Graph Reconstruction}
%\label{s.graphrec}
\textbf{Setup:} We embed graph nodes in a transductive setting. As input and evaluation data we take the shortest distance in the graph between every pair of connected nodes. Unlike previous work \cite{gu2019lmixedCurvature, cruceru20matrixGraph} we do not apply any scaling, neither in the input graph distances nor in the distances calculated on the space. We experiment with the loss proposed in \citet{gu2019lmixedCurvature}, which minimizes the relation between the distance in the space, compared to the distance in the graph, and captures the average distortion. 
%For the Bounded model, we initialize the symmetric matrices with values taken from a uniform distribution $\mathcal{U}(-0.001, 0.001)$. 
%Since the point $\textbf{0}$ is not part of the Siegel upper-half space, 
We initialize the matrix embeddings in the Siegel upper half space by adding small symmetric perturbations to the matrix basepoint $i\Id$. For the Bounded model, we additionally map the points with the Cayley transform (see Appendix~\ref{a.random}). In all cases we optimize with \textsc{Rsgd} \cite{bonnabel2011rsgd} and report the average of $5$ runs.

\textbf{Baselines:} We compare our approach to constant-curvature baselines, such as Euclidean ($\mathbb{E}$) and hyperbolic ($\mathbb{H}$) spaces (we compare to the Poincar\'e model \cite{nickel2017poincare} since the Bounded Domain model is a generalization of it), Cartesian products thereof ($\mathbb{E} \times \mathbb{H}$ and $\mathbb{H} \times \mathbb{H}$) \cite{gu2019lmixedCurvature}, and symmetric positive definite matrices ($\spd$) \cite{cruceru20matrixGraph} in low and high dimensions. Preliminary experiments on the \textit{dual} of HypSPD$_n$ and on spherical spaces showed poor performance thus we do not compare to them (see Appendix~\ref{sec:compact-dual-results}). To establish a fair comparison, each model has the same number of free parameters. This is, the spaces $\calS_n$ and $\calB_n$ have $n (n + 1)$ parameters, thus we compare to baselines of the same dimensionality.\footnote{We also consider comparable dimensionalities for $\spd_n$, which has $\nicefrac{n (n + 1)}{2}$ parameters.} All implementations are taken from Geoopt \cite{geoopt2019kochurov}.

\textbf{Metrics:} Following previous work \cite{deSa18tradeoffs, gu2019lmixedCurvature}, we measure the quality of the learned embeddings by reporting \textit{average distortion} $D_{avg}$, a global metric that considers the explicit value of all distances, and \textit{mean average precision} mAP, a ranking-based measure for local neighborhoods (local metric) as fidelity measures.

\textbf{Synthetic Graphs:}
As a first step, we investigate the representation capabilities of different geometric spaces on synthetic graphs. Previous work has focused on graphs with pure geometric features, such as grids, trees, or their Cartesian products \cite{gu2019lmixedCurvature, cruceru20matrixGraph}, which mix the grid- and tree-like features globally. We expand our analysis to rooted products of trees and grids.
These graphs mix features at different levels and scales. 
Thus, they reflect to a greater extent the complexity of intertwining and varying structure in different regions, making them a better approximation of real-world datasets.
We consider the rooted product \textsc{Tree $\diamond$ Grids} of a tree and  2D grids, and \textsc{Grid $\diamond$ Trees}, of a 2D grid and trees. 
More experimental details, hyperparameters, formulas and statistics about the data are present in Appendix~\ref{sec:appendix-graph-reco}.
%Consider knowledge graphs as a motivating example. The same entity can exhibit multiple local logical patterns given by hierarchical, symmetric and anti-symmetric relations with other entities \cite{chami2020lowdimkge}. A similar case can be found in WordNet \cite{miller1995wordnet}. Words inside a synset show a full-mesh layout with synonyms, and tree-like hypernym relations with other synsets. As illustrated in Figure~\ref{fig:synthetic-graphs}, Cartesian products of trees and grids mix the tree- and grid-like features globally. On the other hand, rooted products of trees with grids and of grids with trees mix these features at different levels and scales. 
%Thus, these graphs reflect to a greater extent the complexity of intertwining and varying structure in different regions, making them a better approximation of real-world datasets. Here we consider the rooted product \textsc{Tree $\diamond$ Grids} of a tree and  2D grids as well as the rooted product \textsc{Grid $\diamond$ Trees} of a 2D grid and trees. 
%Iterating the procedure of rooted products, taking for example the rooted products with different graphs, or a rooted product of a rooted product, one can generate graphs with diverse structures at varying scales.

We report the results for synthetic graphs in Table~\ref{tab:synthetic-results}. 
We find that the Siegel space with Finsler metrics significantly outperform constant curvature baselines in all graphs, except for the tree, where they have competitive results with the hyperbolic models.
We observe that Siegel spaces with the Riemannian metric perform on par with the matching geometric spaces or with the best-fitting product of spaces across graphs of pure geometry (grids and Cartesian products of graphs). However, the $\fone$ metric outperforms the Riemannian and $\finf$ metrics in all graphs, for both models. This is particularly noticeable for the \textsc{4D Grid}, where the distortion achieved by $\fone$ models is almost null, matching the intuition of less distorted grid representations through the taxicab metric.

\begin{table*}[!t]
\vspace{-1mm}
\small
\centering
\adjustbox{max width=0.85\textwidth}{
\begin{tabular}{crrrrrrrrr}
\toprule
\multicolumn{1}{l}{} & \multicolumn{1}{c}{\textsc{USCA312}} & \multicolumn{2}{c}{\textsc{bio-diseasome}} & \multicolumn{2}{c}{\textsc{csphd}} & \multicolumn{2}{c}{\textsc{EuroRoad}} & \multicolumn{2}{c}{\textsc{Facebook}} \\
$(|V|, |E|)$ & \multicolumn{1}{c}{$(312, 48516)$} & \multicolumn{2}{c}{$(516, 1188)$} & \multicolumn{2}{c}{$(1025, 1043)$} & \multicolumn{2}{c}{$(1039, 1305)$} & \multicolumn{2}{c}{$(4039, 88234)$} \\
& \multicolumn{1}{c}{$D_{avg}$} & \multicolumn{1}{c}{$D_{avg}$} & \multicolumn{1}{c}{mAP} & \multicolumn{1}{c}{$D_{avg}$} & \multicolumn{1}{c}{mAP} & \multicolumn{1}{c}{$D_{avg}$} & \multicolumn{1}{c}{mAP} & \multicolumn{1}{c}{$D_{avg}$} & \multicolumn{1}{c}{mAP} \\
\cmidrule(lr){2-2}\cmidrule(lr){3-4}\cmidrule(lr){5-6}\cmidrule(lr){7-8}\cmidrule(lr){9-10}
$\mathbb{E}^{20}$ & \textbf{0.18$\pm$0.01} & 3.83$\pm$0.01 & 76.31 & 4.04$\pm$0.01 & 47.37 & 4.50$\pm$0.00 & 87.70 & 3.16$\pm$0.01 & 32.21 \\
$\mathbb{H}^{20}$ & 2.39$\pm$0.02 & 6.83$\pm$0.08 & 91.26 & 22.42$\pm$0.23 & 60.24 & 43.56$\pm$0.44 & 54.25 & 3.72$\pm$0.00 & 44.85 \\
$\mathbb{E}^{10} \times \mathbb{H}^{10}$ & \textbf{0.18$\pm$0.00} & 2.52$\pm$0.02 & 91.99 & 3.06$\pm$0.02 & 73.25 & 4.24$\pm$0.02 & 89.93 & 2.80$\pm$0.01 & 34.26 \\
$\mathbb{H}^{10} \times \mathbb{H}^{10}$ & 0.47$\pm$0.18 & 2.57$\pm$0.05 & \textbf{95.00} & 7.02$\pm$1.07 & \textbf{79.22} & 23.30$\pm$1.62 & 75.07 & 2.51$\pm$0.00 & 36.39 \\
$\spd_{6}$ & 0.21$\pm$0.02 & 2.54$\pm$0.00 & 82.66 & 2.92$\pm$0.11 & 57.88 & 19.54$\pm$0.99 & 92.38 & 2.92$\pm$0.05 & 33.73 \\
$\mathcal S_{4}^R$ & 0.28$\pm$0.03 & 2.40$\pm$0.02 & 87.01 & 4.30$\pm$0.18 & 59.95 & 29.21$\pm$0.91 & 84.92 & 3.07$\pm$0.04 & 30.98 \\
$\mathcal S_{4}^{F_{\infty}}$ & 0.57$\pm$0.08 & 2.78$\pm$0.49 & 93.95 & 27.27$\pm$1.00 & 59.45 & 46.82$\pm$1.02 & 72.03 & \textbf{1.90$\pm$0.11} & \textbf{45.58} \\
$\mathcal S_{4}^{F_{1}}$ & \textbf{0.18$\pm$0.02} & 1.55$\pm$0.04 & 90.42 & \textbf{1.50$\pm$0.03} & 64.11 & \textbf{3.79$\pm$0.07} & \textbf{94.63} & 2.37$\pm$0.07 & 35.23 \\
$\mathcal B_{4}^R$ & 0.24$\pm$0.07 & 2.69$\pm$0.10 & 89.11 & 28.65$\pm$3.39 & 62.66 & 53.45$\pm$2.65 & 48.75 & 3.58$\pm$0.10 & 30.35 \\
$\mathcal B_{4}^{F_{\infty}}$ & 0.21$\pm$0.04 & 4.58$\pm$0.63 & 90.36 & 26.32$\pm$6.16 & 54.94 & 52.69$\pm$2.28 & 48.75 & 2.18$\pm$0.18 & 39.15 \\
$\mathcal B_{4}^{F_{1}}$ & \textbf{0.18$\pm$0.07} & \textbf{1.54$\pm$0.02} & 90.41 & 2.96$\pm$0.91 & 67.58 & 21.98$\pm$0.62 & 91.63 & 5.05$\pm$0.03 & 39.87 \\
\bottomrule
\end{tabular}
}
\caption{Results for real-world datasets. Lower $D_{avg}$ is better. Higher mAP is better. Metrics are given as percentage.}
\label{tab:real-world-results}
\vspace{-4mm}
\end{table*}

Even when the structure of the data conforms to the geometry of baselines, the Siegel spaces with the Finsler-Riemannian approach are able to outperform them by automatically adapting to very dissimilar patterns without any a priori estimates of the curvature or other features of the graph. This showcases the flexibility of our models, due to its enhanced geometry and higher expressivity.
%This showcases how the model, due to its rich geometry, is able to automatically accommodate the points without any apriori estimates of the curvature or other features of the graph. 
% \todo[inline]{Make comment about achieving perfect distortion for the grid, and ref to sq(2) distortion distance made before maybe}

For graphs with mixed geometric features (rooted products), Cartesian products of spaces cannot arrange these compound geometries into separate Euclidean and hyperbolic subspaces. RSS, on the other hand, offer a less distorted representation of these tangled patterns by exploiting their richer geometry which mixes hyperbolic and Euclidean features. Moreover, they reach a competitive performance on the local neighborhood reconstruction, as the mean precision shows.
Results for more dimensionalities are given in Appendix~\ref{sec:appendix-results}.

% \textbf{Learning the Metric:} We consider a more general approach to Finsler metrics and introduce a model to learn a weighted sum:  $d^{F_{W}}(Z_1,Z_2):=\sum_{i=1}^n \alpha_i v_i$,   
% where $\alpha_i \in \mathbb{R}$ are model parameters. 
% We observed that the model recovers the $\fone$ metric in the cases where it is the most convenient, whereas for \textsc{Tree $\diamond$ Grids}, it finds a more optimal solution (see Appendix~\ref{a.weightLearning}).

\begin{table}[!b]
\vspace{-2mm}
\small
\centering
\adjustbox{max width=\linewidth}{
\begin{tabular}{crrrrrr}
\toprule
\multicolumn{1}{l}{} & \multicolumn{2}{c}{\textsc{Tree $\times$ Grid}} & \multicolumn{2}{c}{\textsc{Grid $\diamond$ Trees}} & \multicolumn{2}{c}{\textsc{bio-diseasome}} \\
\multicolumn{1}{l}{} & \multicolumn{1}{c}{$D_{avg}$} & \multicolumn{1}{c}{mAP} & \multicolumn{1}{c}{$D_{avg}$} & \multicolumn{1}{c}{mAP} & \multicolumn{1}{c}{$D_{avg}$} & \multicolumn{1}{c}{mAP} \\
\cmidrule(lr){2-3}\cmidrule(lr){4-5}\cmidrule(lr){6-7}
$\mathcal S_{4}^R$ & 9.13 & 99.92 & 1.54 & 76.66 & 2.40 & 87.01 \\
$\mathcal S_{4}^{F_{\infty}}$ & 4.81 & 99.28 & 10.48 & 72.53 & 2.78 & 93.95 \\
$\mathcal S_{4}^{F_{1}}$ & \underline{0.81} & \textbf{100.00} & \underline{0.84} & 80.52 & 1.55 & 90.42 \\
\midrule
$\mathbb{E}^{306}$ & 9.80 & 85.14 & 2.81 & 67.69 & 3.52 & 88.45 \\
$\mathbb{H}^{306}$ & 17.31 & 82.97 & 15.92 & 27.14 & 7.04 & 91.46 \\
$\mathbb{S}^{306}$ & 73.78 & 35.36 & 81.67 & 58.26 & 70.91 & 84.61 \\
$\mathbb{E}^{153} \times \mathbb{H}^{153}$ & 9.14 & \textbf{100.00} & 1.52 & \underline{97.85} & 2.36 & 95.65 \\
$\mathbb{S}^{153} \times \mathbb{S}^{153}$ & 60.71 & 6.93 & 70.00 & 5.64 & 55.51 & 19.51 \\
$\mathcal S_{17}^R$ & 9.19 & 99.89 & 1.31 & 75.45 & 2.13 & 93.14 \\
$\mathcal S_{17}^{F_{\infty}}$ & 4.82 & 97.45 & 11.45 & 94.09 & \underline{1.50} & \underline{98.27} \\
$\mathcal S_{17}^{F_{1}}$ & \textbf{0.03} & \textbf{100.00} & \textbf{0.27} & \textbf{99.23} & \textbf{0.73} & \textbf{99.09} \\
\bottomrule
\end{tabular}
}
\caption{Results for different datasets in high-dimensional spaces. Best result is \textbf{bold}, second best \underline{underlined}.}
\label{tab:high-dim-results}
\vspace{-2mm}
\end{table}

\textbf{Real-world Datasets:} We compare the models on two road networks, namely \textsc{USCA312} of distances between North American cities and \textsc{EuroRoad} between European cities, \textsc{bio-diseasome}, a network of human disorders and diseases with reference to their genetic origins \cite{goh2007human}, a graph of computer science Ph.D. advisor-advisee relationships \cite{nooy2011csphdDataset}, and a dense social network from Facebook \cite{mcauley2012fbGraphDataset}. These graphs have been analyzed in previous work as well \cite{gu2019lmixedCurvature, cruceru20matrixGraph}.

We report the results in Table~\ref{tab:real-world-results}.
On the \textsc{USCA312} dataset, which is the only weighted graph under consideration, the Siegel spaces perform on par with the compared target manifolds. For all other datasets, the model with Finsler metrics outperforms all baselines. In line with the results for synthetic datasets, the $\fone$ metric exhibits an outstanding performance across several datasets.

Overall, these results show the strong reconstruction capabilities of RSS for real-world data as well. It also indicates that vertices in these real-world dataset form networks with a more intricate geometry, which the Siegel space is able to unfold to a better extent.

\textbf{High-dimensional Spaces:} In Table~\ref{tab:high-dim-results} we compare the approach in high-dimensional spaces (rank $17$ which is equal to $306$ free parameters), also including spherical spaces $\mathbb{S}$. The results show that our models operate well with larger matrices, where we see further improvement in our distortion and mean average precision over the low dimensional spaces of rank $4$.
We observe that even though we notably increase the dimensions of the baselines to $306$, the Siegel models of rank $4$ (equivalent to $20$ dimensions) significantly outperform them.
These results %that the Siegel space significantly outperforms {\bf product spaces} 
match the expectation that the richer variable curvature geometry of RSS better adapts to graphs with intricate geometric structures.

\section{Analysis of the Embedding Space}
\label{sec:vectorialdistance}

\begin{figure}[!b]
    \vspace{-3mm}
    \centering
    \subfloat{\includegraphics[width=0.33\linewidth, height=2.9cm]{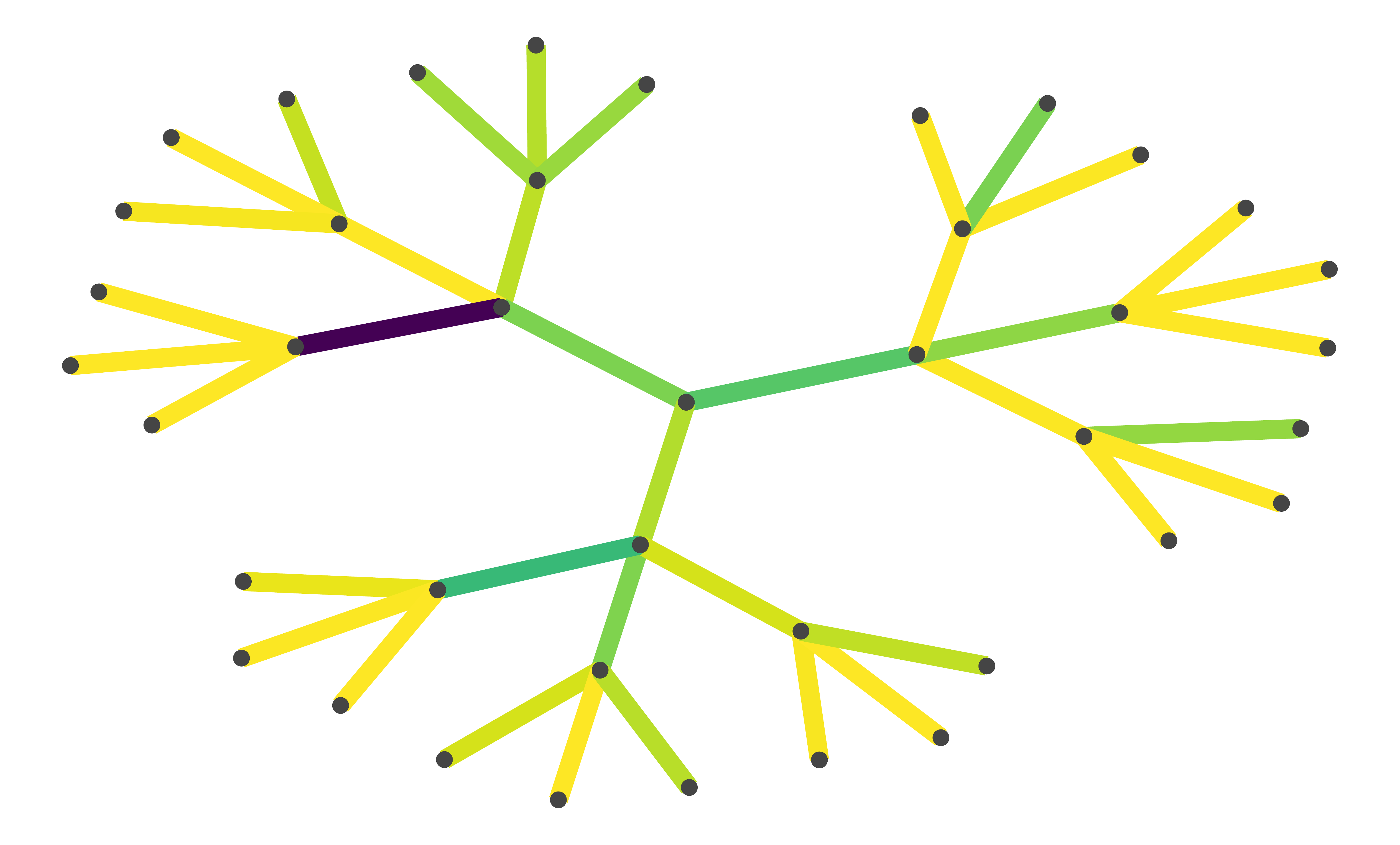}}
    \hfill
    \subfloat{\includegraphics[width=0.33\linewidth, height=2.9cm]{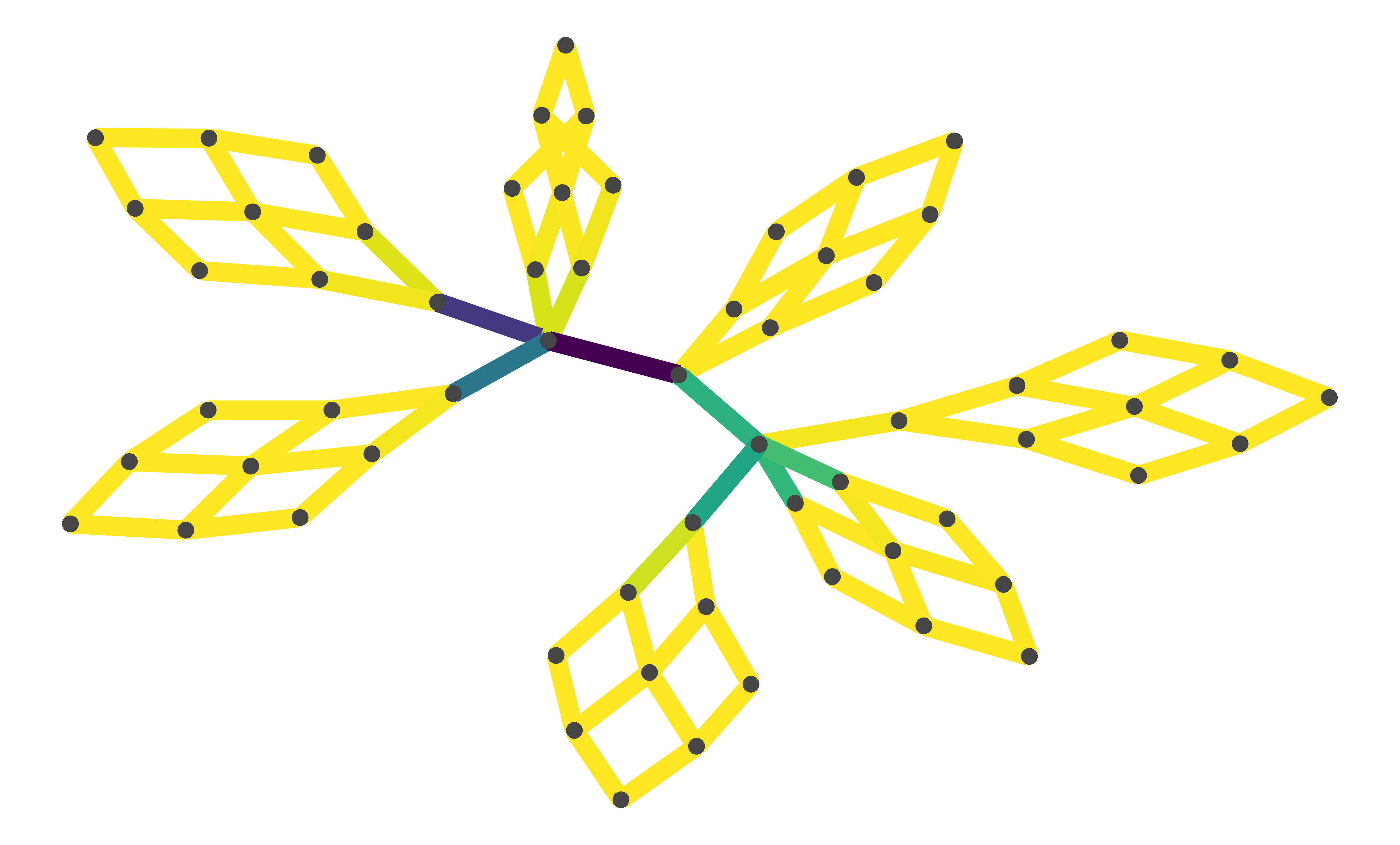}}
    \hfill
    \subfloat{\includegraphics[width=0.34\linewidth, height=2.9cm]{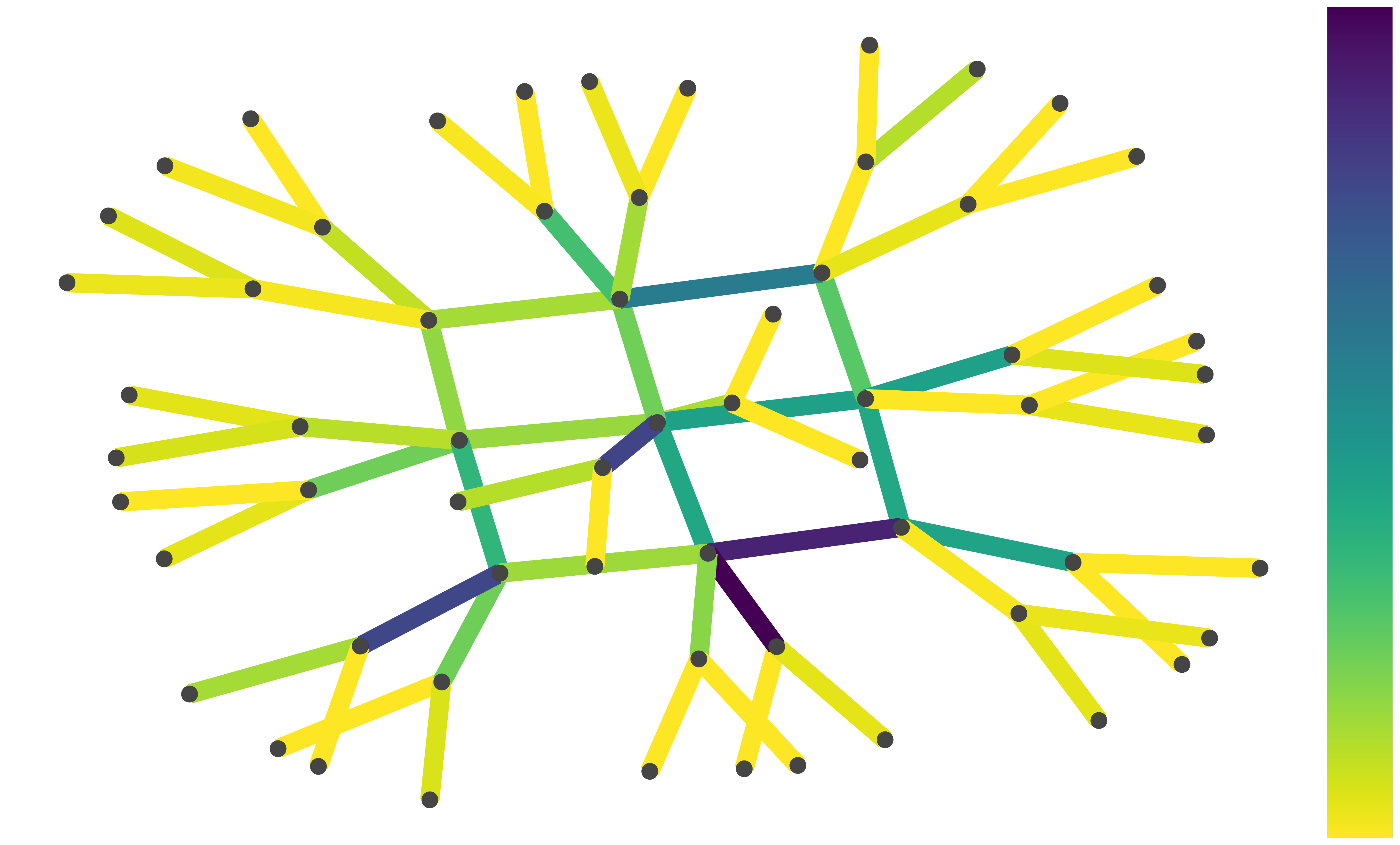}}
    \vspace{-1mm}
    \caption{Edge coloring of $\mathcal S_{2}^{F1}$ for a tree (left), and a rooted product of \textsc{tree} $\diamond$ \textsc{grids} (center), and of \textsc{grid}  $\diamond$ \textsc{trees}.}
    \label{fig:angle-analysis-trees}
    \vspace{-1mm}
\end{figure}

\begin{figure*}[!t]
\vspace{-4mm}
    \centering
    \subfloat{\includegraphics[width=0.33\textwidth, height=4.1cm]{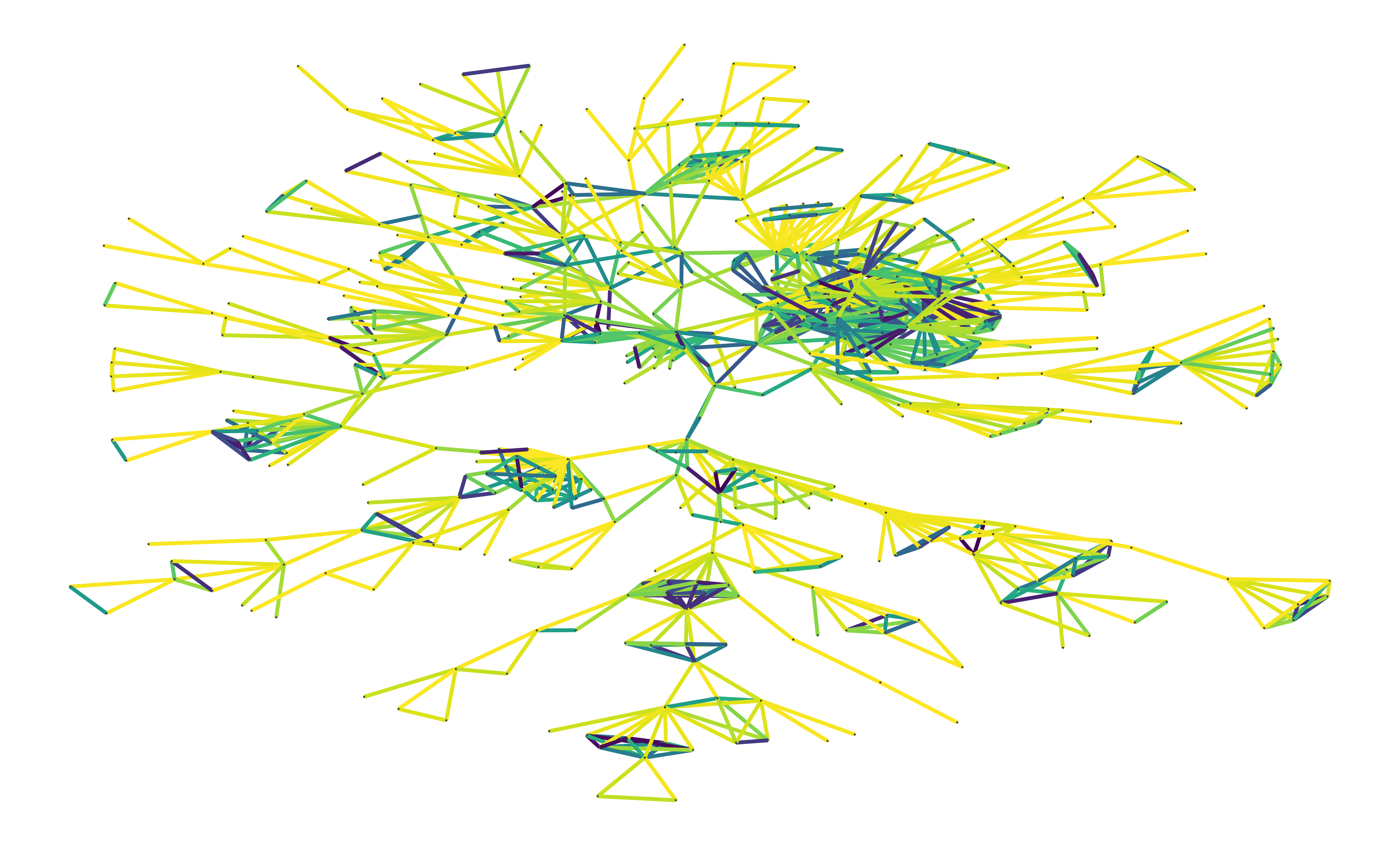}}
    %\hfill
    \subfloat{\includegraphics[width=0.33\textwidth, height=4.1cm]{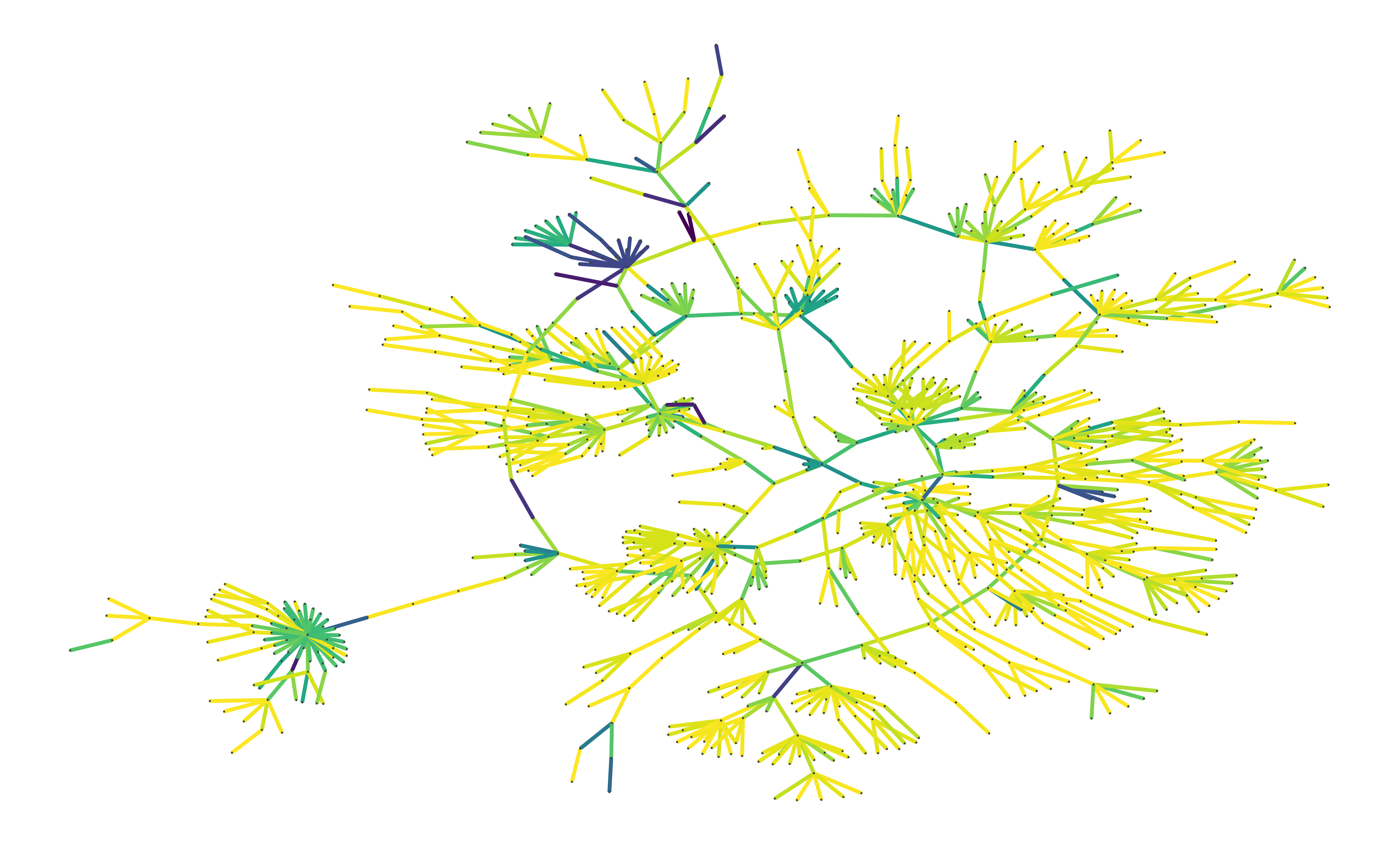}}
   % \hfill
    \subfloat{\includegraphics[width=0.33\textwidth, height=4.1cm]{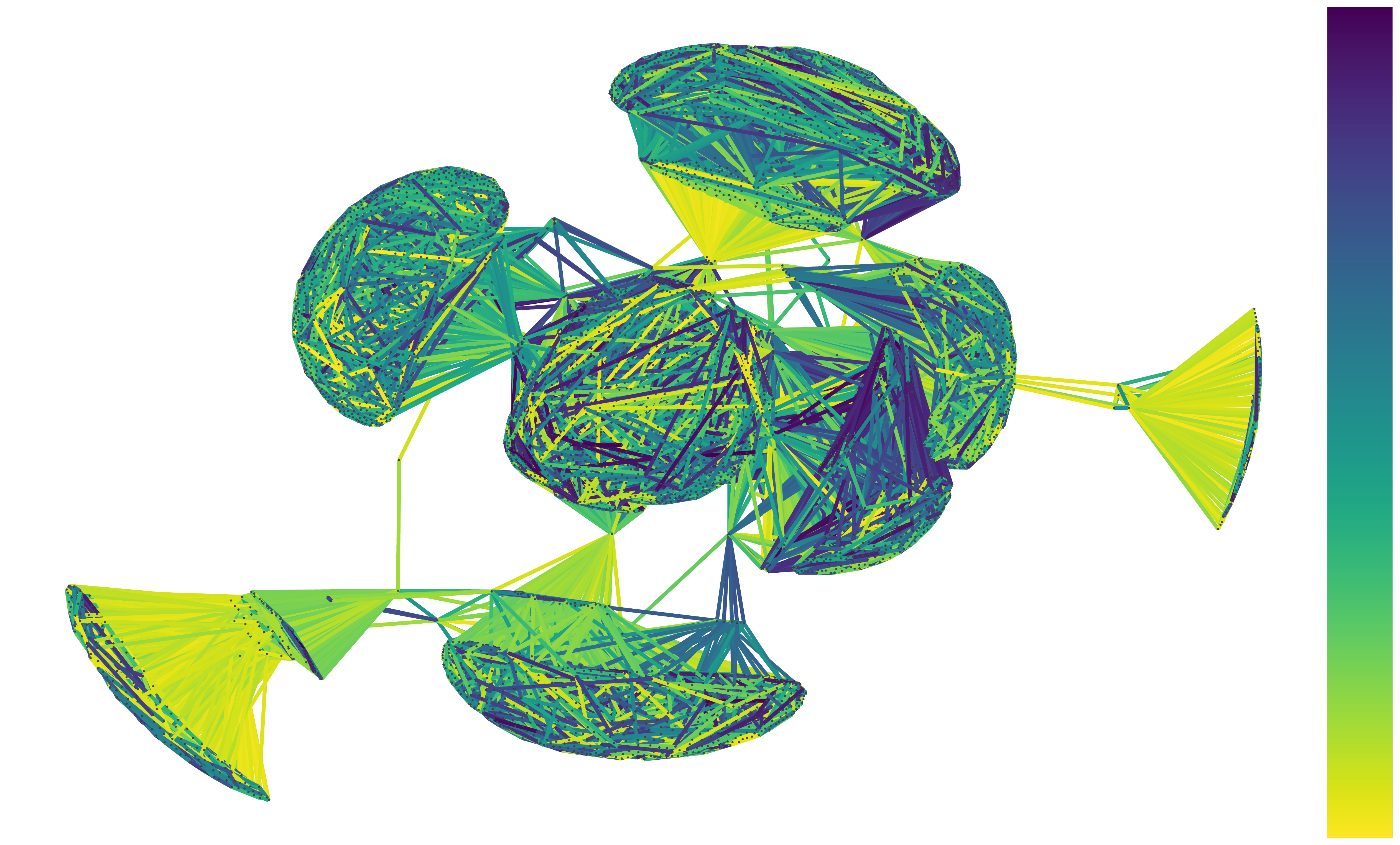}}
    \caption{Edge coloring of $\mathcal S_{2}^{F1}$ for \textsc{bio-diseasome} (left) and \textsc{csphd} (center) and \textsc{facebook} (right). Edge colors indicate the angle of the vector-valued distance for each edge, on a linear scale from 0 (yellow) to $\pi/4$ (blue).}
    \label{fig:angle-analysis-realworld}
    \vspace{-3mm}
\end{figure*}

One reason to embed graphs into Riemannian manifolds is to use geometric properties of the manifold to analyze the structure of the graph. Embeddings into hyperbolic spaces, for example, have been used to infer and visualize hierarchical structure in data sets \cite{nickel2018lorentz}. Visualizations in RSS are difficult due to their high dimensionality. As a solution we use the vector-valued distance function in the RSS to develop a new tool to visualize and to analyze structural properties of the graphs.

We focus on HypSPD$_2$, the Siegel space of rank $k=2$, where the vector-valued distance is just a vector in a cone in $\R^2$. We take edges $(Z_i, Z_j)$ and assign the angle of the vector $\mathrm{vDist}(Z_i,Z_j)=(v_1,v_2)$ (see Algorithm~\ref{alg:Vdistances}, step $6$) to each edge in the graph.
This angle assignment provides a continuous edge coloring that can be leveraged to find structure in graphs.

We see in Figure \ref{fig:angle-analysis-trees} that the edge coloring makes the large-scale structure of the tree (blue/green edges) and the leaves (yellow edges) visible. This is even more striking for the rooted products. In \textsc{tree} $\diamond$ \textsc{grids} the edge coloring distinguishes the hyperbolic parts of the graph (blue edges) and the Euclidean parts (yellow edges). For the \textsc{grid}  $\diamond$ \textsc{trees}, the Euclidean parts are labelled by blue/green edges and the hyperbolic parts by yellow edges. Thus, even though we trained the embedding only on the metric, it automatically adapts to other features of the graph.

In the edge visualizations for real-world datasets (Figure \ref{fig:angle-analysis-realworld}), the edges in the denser connected parts of the graph have a higher angle, as it can be seen for the \textsc{bio-diseasome} and \textsc{facebook} data sets.
For \textsc{csphd}, the tree structure is emphasized by the low angles.

This suggests that the continuous values that we assign to edges are a powerful tool to automatically discover dissimilar patterns in graphs. This can be further used in efficient clustering of the graph. In Appendix \ref{a.vvplot} we give similar visualizations for the Riemannian metric and the $F_\infty$ Finsler metric, showing that also with respect to exhibiting structural properties, the $\fone$ metric performs best. 

\section{Downstream Tasks}
We also evaluate the representation capabilities of Siegel spaces on two downstream tasks: recommender systems and node classification.

\subsection{Recommender Systems}
Our method can be applied in downstream tasks that involve embedding graphs, such as recommender systems. These systems mine user-item interactions and recommend items to users according to the distance/similarity between their respective embeddings \cite{hsieh2017collaborativeML}.
% Recommender systems can be interpreted as a particular case of a link prediction task over a bipartite graph of users and items \cite{li2014rslinkpred}.

% To rank the items $v \in \mathcal{V}$ for each user $u \in \mathcal{U}$ we adapt the scoring function $\phi: \mathcal{U} \times \mathcal{V} \rightarrow \R$ from \citet{balazevic2019murp} as follows:
% \begin{equation}
%     \phi(u, v) = -d_{\mathbb{K}}(u, v)^{2} + b_{u} + b_{v}
% \end{equation}
% where $\mathbb{K}$ is the target manifold, $u, v \in \mathbb{K}; b_{u}, b_{v} \in \R$ are scalar biases for the user and item respectively, and $d_{\mathbb{K}}$ is the corresponding distance in $\mathbb{K}$.
% \todo{Formula in Appendix?}{We} train by minimizing the cross-entropy loss with uniform negative sampling:
% \begin{equation}
% \mathcal{L} = \sum_{(u,v) \in \mathcal{T}} \log(\sigma(\phi(u, v))) + \log(1 - \sigma(\phi(u, w)))
% \end{equation}
% where $\mathcal{T}$ is the set of observed interactions and $w$ is a negative sample (an item the user has not interacted with).
\textbf{Setup:} Given a set of observed user-item interactions $\mathcal{T} = \{(u, v)\}$, we follow a metric learning approach \cite{vinh2018hyperRecommenderSystems} and learn embeddings 
% for user $u \in \mathcal{U}$ and items $v \in \mathcal{V}$ 
by optimizing the following hinge loss function:
\begin{equation}
    \mathcal{L} = \sum_{(u,v) \in \mathcal{T}} \sum_{(u,w) \not\in \mathcal{T}} [m + d_{\mathbb{K}}(\textbf{u}, \textbf{v})^{2} - d_{\mathbb{K}}(\textbf{u}, \textbf{w})^{2}]_{+}
    \label{eq:hinge_loss}
\end{equation}
where $\mathbb{K}$ is the target space, $w$ is an item the user has not interacted with, $\textbf{u}, \textbf{v}, \textbf{w} \in \mathbb{K}$, $m > 0$ is the hinge margin and $[z]_{+} = max(0, z)$.
To generate recommendations, for each user $u$ we rank the items $\textbf{v}_{i}$ according to their distance to $\textbf{u}$. Since it is very costly to rank all the available items, we randomly select $100$ samples which the user has not interacted with, and rank the ground truth amongst these samples \cite{he2017neuralCF}.
We adopt normalized discounted cumulative gain (nDG) %, mean reciprocal ranking (MRR) 
and hit ratio (HR), both at $10$, as ranking evaluation metrics for recommendations. More experimental details and data stats in Appendix~\ref{sec:appen-recosys-details}.

\textbf{Data:}
We evaluate the different models over two MovieLens datasets (\textsc{ml-1m} and \textsc{ml-100k}) \cite{harper2015movieLensDatasets}, \textsc{last.fm}, a dataset of artist listening records \cite{Cantador2011lastFmDataset}, and \textsc{MeetUp}, crawled from Meetup.com \cite{pham2015meetupDataset}.
%, and two branches of the Amazon dataset \cite{nimcauley2019extamazondata}: "Musical Instruments" (\textit{musins}) and "Prime Pantry" (\textit{pripan}).
To generate evaluation splits, the penultimate and last item the user has interacted with are withheld as dev and test set respectively.

\textbf{Results:}
We report the performance for all analyzed models in Table~\ref{tab:recosys-results}. 
%The Siegel models obtain the best performance across all datasets. 
While in the Movies datasets, the Riemannian model marginally outperforms the baselines, in the other two cases the $\fone$ model achieves the highest performance by a larger difference. These systems learn to model users' preferences, and embeds users and items in the space, in a way that is exploited for the task of generating recommendations.
In this manner we demonstrate how downstream tasks can profit from the enhanced graph representation capacity of our models, and we highlight the flexibility of the method, in this case applied in combination with a collaborative metric learning approach \cite{hsieh2017collaborativeML}.

\begin{table}[!t]
\vspace{-1mm}
\small
\centering
\adjustbox{max width=\linewidth}{
\begin{tabular}{crrrrrrrr}
\toprule
\multicolumn{1}{l}{} & \multicolumn{2}{c}{\textsc{ml-1m}} & \multicolumn{2}{c}{\textsc{ml-100k}} & \multicolumn{2}{c}{\textsc{lastfm}} & \multicolumn{2}{c}{\textsc{MeetUp}} \\
\multicolumn{1}{l}{} & HR@10 & nDG & HR@10 & nDG & HR@10 & nDG & HR@10 & nDG \\
\cmidrule(lr){2-3}\cmidrule(lr){4-5}\cmidrule(lr){6-7}\cmidrule(lr){8-9}
$\mathbb{E}^{20}$ & 46.9$\pm$0.6 & 22.7 & 54.6$\pm$1.0 & 28.7 & 55.4$\pm$0.3 & 24.6 & 69.8$\pm$0.4 & 46.4 \\
$\mathbb{H}^{20}$ & 46.0$\pm$0.5 & 23.0 & 53.4$\pm$1.0 & 28.2 & 54.8$\pm$0.5 & 24.9 & 71.8$\pm$0.5 & 48.5 \\
$\mathbb{E}^{10} \times \mathbb{H}^{10}$ & 52.0$\pm$0.7 & 27.4 & 53.1$\pm$1.3 & 27.9 & 45.5$\pm$0.9 & 18.9 & 70.7$\pm$0.2 & 47.5 \\
$\mathbb{H}^{10} \times \mathbb{H}^{10}$ & 46.7$\pm$0.6 & 23.0 & 54.8$\pm$0.9 & 29.1 & 55.0$\pm$0.9 & 24.6 & 71.7$\pm$0.1 & 48.8 \\
\multicolumn{1}{l}{$\mathcal{SPD}_6$} & 45.8$\pm$1.0 & 22.1 & 53.3$\pm$1.4 & 28.0 & 55.4$\pm$0.2 & 25.3 & 70.1$\pm$0.6 & 46.5 \\
$\mathcal S_{4}^R$ & \textbf{53.8$\pm$0.3} & \textbf{27.7} & \textbf{55.7$\pm$0.9} & 28.6 & 53.1$\pm$0.5 & 24.8 & 65.8$\pm$1.2 & 43.4 \\
$\mathcal S_{4}^{F_{\infty}}$ & 45.9$\pm$0.9 & 22.7 & 52.5$\pm$0.3 & 27.5 & 53.8$\pm$1.7 & 32.5 & 69.0$\pm$0.5 & 46.4 \\
$\mathcal S_{4}^{F_{1}}$ & 52.9$\pm$0.6 & 27.2 & 55.6$\pm$1.3 & \textbf{29.4} & \textbf{61.1$\pm$1.2} & \textbf{38.0} & \textbf{74.9$\pm$0.1} & \textbf{52.8} \\
\bottomrule
\end{tabular}
}
\caption{Results for recommender system datasets.}
\label{tab:recosys-results}
\vspace{-4mm}
\end{table}

\subsection{Node Classification}
Our proposed graph embeddings can be used in conjunction with standard machine learning pipelines, such as downstream classification. To demonstrate this, and following the procedure of \citet{chami2020trees}, we embed three hierarchical clustering datasets based on the cosine distance between their points, and then use the learned embeddings as input features for a Euclidean logistic regression model.
Since the node embeddings lie in different metric spaces, we apply the corresponding logarithmic map to obtain a "flat" representation before classifying. 
For the Siegel models of dimension $n$, we first map each complex matrix embedding $Z = X + iY$ to $[(Y+XY^{-1}X, XY^{-1}), (Y^{-1}X, Y^{-1})] \in {\rm SPD}_{2n}$, this is the natural realisation of HypSPD$_n$ as a totally geodesic submanifold of ${\rm SPD}_{2n}$, and then we apply the LogEig map \cite{huang2017riemannianNetForSPDMatrix}, which yields a representation in a flat space. More experimental details in Appendix~\ref{sec:appen-nodecls-details}.

\begin{table}[!b]
\vspace{-4mm}
\small
\centering
\adjustbox{max width=0.7\linewidth}{
\begin{tabular}{cccc}
\toprule
\multicolumn{1}{l}{Dataset} & \textsc{Iris} & \textsc{Zoo} & \textsc{Glass} \\
\midrule
$\mathbb{E}^{20}$ & 83.3$\pm$1.1 & 88.7$\pm$1.8 & 67.2$\pm$2.5 \\
$\mathbb{H}^{20}$ & 84.0$\pm$0.6 & 87.3$\pm$1.5 & 62.8$\pm$2.0 \\
$\mathbb{E}^{10} \times \mathbb{H}^{10}$ & 85.6$\pm$1.1 & 88.0$\pm$1.4 & 64.8$\pm$4.3 \\
$\mathbb{H}^{10} \times \mathbb{H}^{10}$ & 87.8$\pm$1.4 & 87.3$\pm$1.5 & 63.4$\pm$3.4 \\
$\operatorname{SPD}_{6}$ & 88.0$\pm$1.6 & 88.7$\pm$2.2 & 66.9$\pm$2.0 \\
$\mathcal S_{4}^R$ & 88.0$\pm$0.5 & 88.7$\pm$2.2 & 66.6$\pm$2.4 \\
$\mathcal S_{4}^{F_{\infty}}$ & 89.1$\pm$0.5 & 88.7$\pm$2.5 & 65.2$\pm$3.0 \\
$\mathcal S_{4}^{F_{1}}$ & \textbf{89.3$\pm$1.1} & \textbf{90.7$\pm$1.5} & \textbf{67.5$\pm$3.9} \\
$\mathcal B_{4}^R$ & 86.0$\pm$1.9 & 88.7$\pm$1.4 & 65.5$\pm$3.1 \\
$\mathcal B_{4}^{F_{\infty}}$ & 84.4$\pm$0.0 & 87.3$\pm$1.9 & 65.6$\pm$1.7 \\
$\mathcal B_{4}^{F_{1}}$ & 85.6$\pm$1.4 & 89.3$\pm$2.8 & 64.2$\pm$1.7 \\
\bottomrule
\end{tabular}
}
\vspace{-2mm}
\caption{Accuracy for node classification based on its embedding.}
\label{tab:nodecls-results}
\vspace{-2mm}
\end{table}

Results are presented in Table~\ref{tab:nodecls-results}.
In all cases we see that the embeddings learned by our models capture the structural properties of the dataset, so that a simple classifier can separate the nodes into different clusters. They offer the best performance in the three datasets.
% suggesting that clustering learns meaningful partitions of the input similarity graph.
This suggests that embeddings in Siegel spaces learn meaningful representations that can be exploited into downstream tasks. Moreover, we showcase how to map these embeddings to "flat" vectors; in this way they can be integrated with classical Euclidean network layers. 

% AND it showcases a way to integrate them with euclidean and vectorized tools

%%%%%%%%%%%%%%%%%%%% CONCLUSION %%%%%%%%%%%%%%%%%%%%%%%%%%%%%%%%
\section{Conclusions \& Future Work}
Riemannian manifold learning has regained attention due to appealing geometric properties that allow methods to represent non-Euclidean data arising in several domains. 
We propose the systematic use of symmetric spaces to encompass previous work in representation learning, and develop a toolkit that allows 
%We develop a general framework that allows 
practitioners to choose a Riemannian symmetric space and implement the mathematical tools required to learn graph embeddings.
We introduce the use of Finsler metrics integrated with a Riemannian optimization scheme, which provide a significantly less distorted representation over several data sets.
As a new tool to discover structure in the graph, we leverage the vector-valued distance function on a RSS. 
We implement these ideas on Siegel spaces, a rich class of RSS that had not been explored in geometric deep learning, and we develop tractable and mathematically sound algorithms to learn embeddings in these spaces through gradient-descent methods.
We showcase the effectiveness of the proposed approach on conventional as well as new datasets for the graph reconstruction task, and in two downstream tasks. 
Our method ties or outperforms constant-curvature baselines without requiring any previous assumption on geometric features of the graphs. This shows the flexibility and enhanced representation capacity of Siegel spaces, as well as the versatility of our approach.

As future directions, we consider applying the vector-valued distance in clustering and structural analysis of graphs, and
%Other directions lines opened by our work are the introduction of powerful tools from harmonic analysis, such as the Poisson transformation, in geometric deep learning, and 
the development of deep neural network architectures adapted to the geometry of RSS, specifically Siegel spaces.
%Since the Siegel space embeds as a subspace of its compact dual, there is an explicit geometric transition, generalizing the geometric transition between spherical and hyperbolic geometry (see Appendix \ref{a.transition}). 
A further interesting research direction is to use geometric transition between symmetric spaces to extend the approach demonstrated by curvature learning {\`a} la \citet{gu2019lmixedCurvature}.
We plan to leverage the structure of the Siegel space of a hyperbolic plane over SPD to analyze medical imaging data, which is often given as symmetric positive definite matrices, see \citet{Pennec:2020aa}.

%%%%%%%%%%%%%%%%%%%%%%%%%%%%% ACK %%%%%%%%%%%%%%%%%%%%%%%%%%%%%%
\section*{Acknowledgements}
This work has been supported by the German Research Foundation (DFG) as part of the Research Training Group AIPHES under grant No. GRK 1994/1, as well as under Germany’s Excellence Strategy EXC-2181/1 - 390900948 (the Heidelberg STRUCTURES Cluster of Excellence), and by the Klaus Tschira Foundation, Heidelberg, Germany. 

\bibliography{mybib}

\begin{thebibliography}{53}
\providecommand{\natexlab}[1]{#1}
\providecommand{\url}[1]{\texttt{#1}}
\expandafter\ifx\csname urlstyle\endcsname\relax
  \providecommand{\doi}[1]{doi: #1}\else
  \providecommand{\doi}{doi: \begingroup \urlstyle{rm}\Url}\fi

\bibitem[Bachmann et~al.(2020)Bachmann, B{\'{e}}cigneul, and
  Ganea]{bachmann2020ccgcn}
Bachmann, G., B{\'{e}}cigneul, G., and Ganea, O.-E.
\newblock Constant curvature graph convolutional networks.
\newblock In \emph{37th International Conference on Machine Learning (ICML)},
  2020.

\bibitem[B\'ecigneul \& Ganea(2019)B\'ecigneul and
  Ganea]{becigneul2019riemannianMethods}
B\'ecigneul, G. and Ganea, O.-E.
\newblock Riemannian adaptive optimization methods.
\newblock In \emph{7th International Conference on Learning Representations,
  {ICLR}}, New Orleans, LA, USA, May 2019.
\newblock URL \url{https://openreview.net/forum?id=r1eiqi09K7}.

\bibitem[Boland \& Newberger(2001)Boland and
  Newberger]{boland2001minimalEntropyFinslerMetrics}
Boland, J. and Newberger, F.
\newblock Minimal entropy rigidity for {F}insler manifolds of negative flag
  curvature.
\newblock \emph{Ergodic Theory and Dynamical Systems}, 21\penalty0
  (1):\penalty0 13–23, 2001.
\newblock \doi{10.1017/S0143385701001055}.

\bibitem[Bonnabel(2011)]{bonnabel2011rsgd}
Bonnabel, S.
\newblock Stochastic gradient descent on {R}iemannian manifolds.
\newblock \emph{IEEE Transactions on Automatic Control}, 58, 11 2011.
\newblock \doi{10.1109/TAC.2013.2254619}.

\bibitem[{Bronstein} et~al.(2017){Bronstein}, {Bruna}, {LeCun}, {Szlam}, and
  {Vandergheynst}]{bronstein2018geomdeeplearning}
{Bronstein}, M.~M., {Bruna}, J., {LeCun}, Y., {Szlam}, A., and {Vandergheynst},
  P.
\newblock Geometric deep learning: Going beyond {E}uclidean data.
\newblock \emph{IEEE Signal Processing Magazine}, 34\penalty0 (4):\penalty0
  18--42, 2017.

\bibitem[Cantador et~al.(2011)Cantador, Brusilovsky, and
  Kuflik]{Cantador2011lastFmDataset}
Cantador, I., Brusilovsky, P., and Kuflik, T.
\newblock 2nd {W}orkshop on {I}nformation {H}eterogeneity and {F}usion in
  {R}ecommender {S}ystems ({HetRec} 2011).
\newblock In \emph{Proceedings of the 5th ACM Conference on Recommender
  Systems}, RecSys 2011, New York, NY, USA, 2011. ACM.

\bibitem[Cayley(1846)]{cayley1846transform}
Cayley, A.
\newblock Sur quelques propriétés des déterminants gauches.
\newblock \emph{Journal für die reine und angewandte {M}athematik},
  32:\penalty0 119--123, 1846.
\newblock URL
  \url{http://www.digizeitschriften.de/dms/img/?PID=GDZPPN002145308}.

\bibitem[Chamberlain et~al.(2017)Chamberlain, Deisenroth, and
  Clough]{Chamberlain2017}
Chamberlain, B., Deisenroth, M., and Clough, J.
\newblock Neural embeddings of graphs in hyperbolic space.
\newblock In \emph{Proceedings of the 13th International Workshop on Mining and
  Learning with Graphs (MLG)}, 2017.

\bibitem[Chami et~al.(2019)Chami, Ying, R\'{e}, and Leskovec]{chami2019hgcnn}
Chami, I., Ying, Z., R\'{e}, C., and Leskovec, J.
\newblock Hyperbolic graph convolutional neural networks.
\newblock In \emph{Advances in Neural Information Processing Systems 32}, pp.\
  4869--4880. Curran Associates, Inc., 2019.
\newblock URL
  \url{https://proceedings.neurips.cc/paper/2019/file/0415740eaa4d9decbc8da001d3fd805f-Paper.pdf}.

\bibitem[Chami et~al.(2020)Chami, Gu, Chatziafratis, and Ré]{chami2020trees}
Chami, I., Gu, A., Chatziafratis, V., and Ré, C.
\newblock From trees to continuous embeddings and back: Hyperbolic hierarchical
  clustering.
\newblock In Larochelle, H., Ranzato, M., Hadsell, R., Balcan, M., and Lin, H.
  (eds.), \emph{Advances in Neural Information Processing Systems 33: Annual
  Conference on Neural Information Processing Systems 2020, NeurIPS 2020,
  December 6-12, 2020, virtual}, 2020.

\bibitem[Cruceru et~al.(2020)Cruceru, B{\'{e}}cigneul, and
  Ganea]{cruceru20matrixGraph}
Cruceru, C., B{\'{e}}cigneul, G., and Ganea, O.-E.
\newblock Computationally tractable {R}iemannian manifolds for graph
  embeddings.
\newblock In \emph{37th International Conference on Machine Learning (ICML)},
  2020.

\bibitem[Defferrard et~al.(2020)Defferrard, Milani, Gusset, and
  Perraudin]{defferrard2020DeepSphere}
Defferrard, M., Milani, M., Gusset, F., and Perraudin, N.
\newblock Deep{S}phere: A graph-based spherical {CNN}.
\newblock In \emph{International Conference on Learning Representations}, 2020.
\newblock URL \url{https://openreview.net/forum?id=B1e3OlStPB}.

\bibitem[Donoho \& Tsaig(2008)Donoho and Tsaig]{Donoho:2008aa}
Donoho, D.~L. and Tsaig, Y.
\newblock Fast solution of $\ell_1$-norm minimization problems when the
  solution may be sparse.
\newblock \emph{IEEE Trans. Information Theory}, 54\penalty0 (11):\penalty0
  4789--4812, 2008.

\bibitem[Dua \& Graff(2017)Dua and Graff]{dua2017uciMLRepo}
Dua, D. and Graff, C.
\newblock {UCI} machine learning repository, 2017.
\newblock URL \url{http://archive.ics.uci.edu/ml}.

\bibitem[Falkenberg(2007)]{falkenberg2007matrixinverse}
Falkenberg, A.
\newblock Method to calculate the inverse of a complex matrix using real matrix
  inversion.
\newblock 2007.

\bibitem[Ganea et~al.(2018)Ganea, B\'ecigneul, and
  Hofmann]{ganea2018hyperEntailmentCones}
Ganea, O.-E., B\'ecigneul, G., and Hofmann, T.
\newblock Hyperbolic entailment cones for learning hierarchical embeddings.
\newblock In Dy, J. and Krause, A. (eds.), \emph{Proceedings of the 35th
  International Conference on Machine Learning}, volume~80 of \emph{Proceedings
  of Machine Learning Research}, pp.\  1646--1655, Stockholmsmässan, Stockholm
  Sweden, 10--15 Jul 2018. PMLR.
\newblock URL \url{http://proceedings.mlr.press/v80/ganea18a.html}.

\bibitem[Goh et~al.(2007)Goh, Cusick, Valle, Childs, Vidal, and
  Barab{\'a}si]{goh2007human}
Goh, K.-I., Cusick, M.~E., Valle, D., Childs, B., Vidal, M., and Barab{\'a}si,
  A.-L.
\newblock The human disease network.
\newblock \emph{Proceedings of the National Academy of Sciences}, 104\penalty0
  (21):\penalty0 8685--8690, 2007.

\bibitem[Grattarola et~al.(2020)Grattarola, Zambon, Livi, and
  Alippi]{grattarola2020constantCurvatureGraphEmbeds}
Grattarola, D., Zambon, D., Livi, L., and Alippi, C.
\newblock Change detection in graph streams by learning graph embeddings on
  constant-curvature manifolds.
\newblock \emph{{IEEE} Trans. Neural Networks Learn. Syst.}, 31\penalty0
  (6):\penalty0 1856--1869, 2020.
\newblock \doi{10.1109/TNNLS.2019.2927301}.
\newblock URL \url{https://doi.org/10.1109/TNNLS.2019.2927301}.

\bibitem[Gu et~al.(2019)Gu, Sala, Gunel, and Ré]{gu2019lmixedCurvature}
Gu, A., Sala, F., Gunel, B., and Ré, C.
\newblock Learning mixed-curvature representations in product spaces.
\newblock In \emph{International Conference on Learning Representations}, 2019.
\newblock URL \url{https://openreview.net/forum?id=HJxeWnCcF7}.

\bibitem[Hagberg et~al.(2008)Hagberg, Schult, and Swart]{networkx2008hagberg}
Hagberg, A.~A., Schult, D.~A., and Swart, P.~J.
\newblock Exploring network structure, dynamics, and function using {NetworkX}.
\newblock In Varoquaux, G., Vaught, T., and Millman, J. (eds.),
  \emph{Proceedings of the 7th Python in Science Conference}, pp.\  11 -- 15,
  Pasadena, CA USA, 2008.

\bibitem[Harper \& Konstan(2015)Harper and
  Konstan]{harper2015movieLensDatasets}
Harper, F.~M. and Konstan, J.~A.
\newblock The {MovieLens} datasets: History and context.
\newblock \emph{ACM Trans. Interact. Intell. Syst.}, 5\penalty0 (4), December
  2015.
\newblock ISSN 2160-6455.
\newblock \doi{10.1145/2827872}.
\newblock URL \url{https://doi.org/10.1145/2827872}.

\bibitem[He et~al.(2017)He, Liao, Zhang, Nie, Hu, and Chua]{he2017neuralCF}
He, X., Liao, L., Zhang, H., Nie, L., Hu, X., and Chua, T.-S.
\newblock Neural collaborative filtering.
\newblock In \emph{Proceedings of the 26th International Conference on World
  Wide Web}, WWW '17, pp.\  173–182, Republic and Canton of Geneva, CHE,
  2017. International World Wide Web Conferences Steering Committee.
\newblock ISBN 9781450349130.
\newblock \doi{10.1145/3038912.3052569}.
\newblock URL \url{https://doi.org/10.1145/3038912.3052569}.

\bibitem[Helgason(1978)]{helgason1078diffGeom}
Helgason, S.
\newblock \emph{Differential geometry, Lie groups, and symmetric spaces}.
\newblock Academic Press New York, 1978.
\newblock ISBN 0123384605.

\bibitem[Hsieh et~al.(2017)Hsieh, Yang, Cui, Lin, Belongie, and
  Estrin]{hsieh2017collaborativeML}
Hsieh, C.-K., Yang, L., Cui, Y., Lin, T.-Y., Belongie, S., and Estrin, D.
\newblock Collaborative metric learning.
\newblock In \emph{Proceedings of the 26th International Conference on World
  Wide Web}, WWW '17, pp.\  193–201, Republic and Canton of Geneva, CHE,
  2017. International World Wide Web Conferences Steering Committee.
\newblock ISBN 9781450349130.
\newblock \doi{10.1145/3038912.3052639}.
\newblock URL \url{https://doi.org/10.1145/3038912.3052639}.

\bibitem[Huang \& Gool(2017)Huang and Gool]{huang2017riemannianNetForSPDMatrix}
Huang, Z. and Gool, L.~V.
\newblock A {R}iemannian network for {SPD} matrix learning.
\newblock In \emph{Proceedings of the Thirty-First AAAI Conference on
  Artificial Intelligence}, AAAI'17, pp.\  2036–2042. AAAI Press, 2017.

\bibitem[Huang et~al.(2018)Huang, Wu, and
  Gool]{huang2018DeepNetsOnGrassmannManifolds}
Huang, Z., Wu, J., and Gool, L.~V.
\newblock Building deep networks on {G}rassmann manifolds.
\newblock In McIlraith, S.~A. and Weinberger, K.~Q. (eds.), \emph{Proceedings
  of the Thirty-Second {AAAI} Conference on Artificial Intelligence, (AAAI-18),
  the 30th Innovative Applications of Artificial Intelligence (IAAI-18), and
  the 8th {AAAI} Symposium on Educational Advances in Artificial Intelligence
  (EAAI-18), New Orleans, Louisiana, USA, February 2-7, 2018}, pp.\
  3279--3286. {AAAI} Press, 2018.
\newblock URL
  \url{https://www.aaai.org/ocs/index.php/AAAI/AAAI18/paper/view/16846}.

\bibitem[Kochurov et~al.(2020)Kochurov, Karimov, and
  Kozlukov]{geoopt2019kochurov}
Kochurov, M., Karimov, R., and Kozlukov, S.
\newblock Geoopt: Riemannian optimization in {PyTorch}.
\newblock \emph{ArXiv}, abs/2005.02819, 2020.

\bibitem[Krioukov et~al.(2009)Krioukov, Papadopoulos, Vahdat, and
  Boguñá]{krioukov2009tempNets}
Krioukov, D., Papadopoulos, F., Vahdat, A., and Boguñá, M.
\newblock {On Curvature and Temperature of Complex Networks}.
\newblock \emph{Physical Review E}, 80\penalty0 (035101), Sep 2009.

\bibitem[Krioukov et~al.(2010)Krioukov, Papadopoulos, Kitsak, Vahdat, and
  Boguñá]{krioukov2010hypernetworks}
Krioukov, D., Papadopoulos, F., Kitsak, M., Vahdat, A., and Boguñá, M.
\newblock Hyperbolic geometry of complex networks.
\newblock \emph{Physical review. E, Statistical, nonlinear, and soft matter
  physics}, 82:\penalty0 036106, 09 2010.
\newblock \doi{10.1103/PhysRevE.82.036106}.

\bibitem[Law \& Stam(2020)Law and Stam]{law2020ultrahyperbolic}
Law, M.~T. and Stam, J.
\newblock Ultrahyperbolic representation learning.
\newblock In Larochelle, H., Ranzato, M., Hadsell, R., Balcan, M., and Lin, H.
  (eds.), \emph{Advances in Neural Information Processing Systems 33: Annual
  Conference on Neural Information Processing Systems 2020, NeurIPS 2020,
  December 6-12, 2020, virtual}, 2020.

\bibitem[{Liu} et~al.(2017){Liu}, {Wen}, {Yu}, {Li}, {Raj}, and
  {Song}]{liu2017sphereface}
{Liu}, W., {Wen}, Y., {Yu}, Z., {Li}, M., {Raj}, B., and {Song}, L.
\newblock {SphereFace}: Deep hypersphere embedding for face recognition.
\newblock In \emph{2017 IEEE Conference on Computer Vision and Pattern
  Recognition (CVPR)}, pp.\  6738--6746, 2017.
\newblock \doi{10.1109/CVPR.2017.713}.

\bibitem[L{\'o}pez \& Strube(2020)L{\'o}pez and Strube]{lopez2020fullyhyper}
L{\'o}pez, F. and Strube, M.
\newblock A fully hyperbolic neural model for hierarchical multi-class
  classification.
\newblock In \emph{Findings of the Association for Computational Linguistics:
  EMNLP 2020}, pp.\  460--475, Online, November 2020. Association for
  Computational Linguistics.
\newblock URL \url{https://www.aclweb.org/anthology/2020.findings-emnlp.42}.

\bibitem[L{\'o}pez et~al.(2019)L{\'o}pez, Heinzerling, and
  Strube]{lopez2019figetinHS}
L{\'o}pez, F., Heinzerling, B., and Strube, M.
\newblock Fine-grained entity typing in hyperbolic space.
\newblock In \emph{Proceedings of the 4th Workshop on Representation Learning
  for NLP (RepL4NLP-2019)}, pp.\  169--180, Florence, Italy, August 2019.
  Association for Computational Linguistics.
\newblock \doi{10.18653/v1/W19-4319}.
\newblock URL \url{https://www.aclweb.org/anthology/W19-4319}.

\bibitem[McAuley \& Leskovec(2012)McAuley and
  Leskovec]{mcauley2012fbGraphDataset}
McAuley, J. and Leskovec, J.
\newblock Learning to discover social circles in ego networks.
\newblock In \emph{Proceedings of the 25th International Conference on Neural
  Information Processing Systems - Volume 1}, NIPS'12, pp.\  539–547, Red
  Hook, NY, USA, 2012. Curran Associates Inc.

\bibitem[Meng et~al.(2019)Meng, Huang, Wang, Zhang, Zhuang, Kaplan, and
  Han]{meng2019sphericalTextEmbed}
Meng, Y., Huang, J., Wang, G., Zhang, C., Zhuang, H., Kaplan, L., and Han, J.
\newblock Spherical text embedding.
\newblock In Wallach, H., Larochelle, H., Beygelzimer, A., d\textquotesingle
  Alch\'{e}-Buc, F., Fox, E., and Garnett, R. (eds.), \emph{Advances in Neural
  Information Processing Systems}, volume~32, pp.\  8208--8217. Curran
  Associates, Inc., 2019.
\newblock URL
  \url{https://proceedings.neurips.cc/paper/2019/file/043ab21fc5a1607b381ac3896176dac6-Paper.pdf}.

\bibitem[Nickel \& Kiela(2017)Nickel and Kiela]{nickel2017poincare}
Nickel, M. and Kiela, D.
\newblock Poincar\'{e} embeddings for learning hierarchical representations.
\newblock In Guyon, I., Luxburg, U.~V., Bengio, S., Wallach, H., Fergus, R.,
  Vishwanathan, S., and Garnett, R. (eds.), \emph{Advances in Neural
  Information Processing Systems 30}, pp.\  6341--6350. Curran Associates,
  Inc., 2017.
\newblock URL
  \url{https://proceedings.neurips.cc/paper/2017/file/59dfa2df42d9e3d41f5b02bfc32229dd-Paper.pdf}.

\bibitem[Nickel \& Kiela(2018)Nickel and Kiela]{nickel2018lorentz}
Nickel, M. and Kiela, D.
\newblock Learning continuous hierarchies in the {L}orentz model of hyperbolic
  geometry.
\newblock In Dy, J. and Krause, A. (eds.), \emph{Proceedings of the 35th
  International Conference on Machine Learning}, volume~80 of \emph{Proceedings
  of Machine Learning Research}, pp.\  3779--3788, Stockholmsmässan, Stockholm
  Sweden, 10--15 Jul 2018. PMLR.
\newblock URL \url{http://proceedings.mlr.press/v80/nickel18a.html}.

\bibitem[Nielsen \& Sun(2019)Nielsen and Sun]{Nielsen2019clusteringWithFinsler}
Nielsen, F. and Sun, K.
\newblock \emph{Clustering in Hilbert's Projective Geometry: The Case Studies
  of the Probability Simplex and the Elliptope of Correlation Matrices}, pp.\
  297--331.
\newblock Springer International Publishing, Cham, 2019.
\newblock ISBN 978-3-030-02520-5.
\newblock \doi{10.1007/978-3-030-02520-5_11}.
\newblock URL \url{https://doi.org/10.1007/978-3-030-02520-5_11}.

\bibitem[Nooy et~al.(2011)Nooy, Mrvar, and Batagelj]{nooy2011csphdDataset}
Nooy, W.~D., Mrvar, A., and Batagelj, V.
\newblock \emph{Exploratory Social Network Analysis with Pajek}.
\newblock Cambridge University Press, USA, 2011.
\newblock ISBN 0521174805.

\bibitem[Paszke et~al.(2019)Paszke, Gross, Massa, Lerer, Bradbury, Chanan,
  Killeen, Lin, Gimelshein, Antiga, Desmaison, Kopf, Yang, DeVito, Raison,
  Tejani, Chilamkurthy, Steiner, Fang, Bai, and
  Chintala]{paszke2019pytorchNeurips}
Paszke, A., Gross, S., Massa, F., Lerer, A., Bradbury, J., Chanan, G., Killeen,
  T., Lin, Z., Gimelshein, N., Antiga, L., Desmaison, A., Kopf, A., Yang, E.,
  DeVito, Z., Raison, M., Tejani, A., Chilamkurthy, S., Steiner, B., Fang, L.,
  Bai, J., and Chintala, S.
\newblock Pytorch: An imperative style, high-performance deep learning library.
\newblock In Wallach, H., Larochelle, H., Beygelzimer, A., d\textquotesingle
  Alch\'{e}-Buc, F., Fox, E., and Garnett, R. (eds.), \emph{Advances in Neural
  Information Processing Systems 32}, pp.\  8024--8035. Curran Associates,
  Inc., 2019.

\bibitem[Pennec(2020)]{Pennec:2020aa}
Pennec, X.
\newblock {Manifold-Valued Image Processing with SDP Matrices}.
\newblock In \emph{{Riemannian Geometric Statistics in Medical Image
  Analysis}}, pp.\  75--134. Academic Press, 2020.

\bibitem[{Pham} et~al.(2015){Pham}, {Li}, {Cong}, and
  {Zhang}]{pham2015meetupDataset}
{Pham}, T.~N., {Li}, X., {Cong}, G., and {Zhang}, Z.
\newblock A general graph-based model for recommendation in event-based social
  networks.
\newblock In \emph{2015 IEEE 31st International Conference on Data
  Engineering}, pp.\  567--578, 2015.
\newblock \doi{10.1109/ICDE.2015.7113315}.

\bibitem[Ratliff et~al.(2020)Ratliff, Wyk, Xie, Li, and
  Rana]{ratliff2020finslerRobotics}
Ratliff, N.~D., Wyk, K.~V., Xie, M., Li, A., and Rana, M.~A.
\newblock Generalized nonlinear and {Finsler} geometry for robotics.
\newblock \emph{CoRR}, abs/2010.14745, 2020.
\newblock URL \url{https://arxiv.org/abs/2010.14745}.

\bibitem[Rossi \& Ahmed(2015)Rossi and Ahmed]{networkrepository2015rossi}
Rossi, R.~A. and Ahmed, N.~K.
\newblock The network data repository with interactive graph analytics and
  visualization.
\newblock In \emph{AAAI}, 2015.
\newblock URL \url{http://networkrepository.com}.

\bibitem[Rubin-Delanchy(2020)]{rubindelanchy2020manifold}
Rubin-Delanchy, P.
\newblock Manifold structure in graph embeddings, 2020.
\newblock URL \url{https://arxiv.org/abs/2006.05168}.

\bibitem[Sala et~al.(2018)Sala, De~Sa, Gu, and Re]{deSa18tradeoffs}
Sala, F., De~Sa, C., Gu, A., and Re, C.
\newblock Representation tradeoffs for hyperbolic embeddings.
\newblock In Dy, J. and Krause, A. (eds.), \emph{Proceedings of the 35th
  International Conference on Machine Learning}, volume~80 of \emph{Proceedings
  of Machine Learning Research}, pp.\  4460--4469, Stockholmsmässan, Stockholm
  Sweden, 10--15 Jul 2018. PMLR.
\newblock URL \url{http://proceedings.mlr.press/v80/sala18a.html}.

\bibitem[Siegel(1943)]{siegel1943symplectic}
Siegel, C.~L.
\newblock Symplectic geometry.
\newblock \emph{American Journal of Mathematics}, 65\penalty0 (1):\penalty0
  1--86, 1943.
\newblock ISSN 00029327, 10806377.
\newblock URL \url{http://www.jstor.org/stable/2371774}.

\bibitem[Skopek et~al.(2020)Skopek, Ganea, and Becigneul]{skopek2020mixedva}
Skopek, O., Ganea, O.-E., and Becigneul, G.
\newblock Mixed-curvature variational autoencoders.
\newblock In \emph{8th International Conference on Learning Representations
  (ICLR)}, April 2020.
\newblock URL \url{https://openreview.net/pdf?id=S1g6xeSKDS}.

\bibitem[Takagi(1924)]{takagiFactorization1924}
Takagi, T.
\newblock On an algebraic problem related to an analytic theorem of
  carath\'{e}odory and fej\'{e}r and on an allied theorem of {Landau}.
\newblock \emph{Japanese journal of mathematics :transactions and abstracts},
  1:\penalty0 83--93, 1924.
\newblock \doi{10.4099/jjm1924.1.0_83}.

\bibitem[Tifrea et~al.(2019)Tifrea, B{\'{e}}cigneul, and
  Ganea]{tifrea2018poincareGlove}
Tifrea, A., B{\'{e}}cigneul, G., and Ganea, O.-E.
\newblock Poincare {G}lo{V}e: Hyperbolic word embeddings.
\newblock In \emph{7th International Conference on Learning Representations,
  {ICLR}}, New Orleans, LA, USA, May 2019.
\newblock URL \url{https://openreview.net/forum?id=Ske5r3AqK7}.

\bibitem[Vinh~Tran et~al.(2020)Vinh~Tran, Tay, Zhang, Cong, and
  Li]{vinh2018hyperRecommenderSystems}
Vinh~Tran, L., Tay, Y., Zhang, S., Cong, G., and Li, X.
\newblock {HyperML}: A boosting metric learning approach in hyperbolic space
  for recommender systems.
\newblock In \emph{Proceedings of the 13th International Conference on Web
  Search and Data Mining}, WSDM '20, pp.\  609–617, New York, NY, USA, 2020.
  Association for Computing Machinery.
\newblock ISBN 9781450368223.
\newblock \doi{10.1145/3336191.3371850}.
\newblock URL \url{https://doi.org/10.1145/3336191.3371850}.

\bibitem[{Wilson} et~al.(2014){Wilson}, {Hancock}, {Pekalska}, and
  {Duin}]{wilson2014sphere}
{Wilson}, R.~C., {Hancock}, E.~R., {Pekalska}, E., and {Duin}, R. P.~W.
\newblock Spherical and hyperbolic embeddings of data.
\newblock \emph{IEEE Transactions on Pattern Analysis and Machine
  Intelligence}, 36\penalty0 (11):\penalty0 2255--2269, 2014.

\bibitem[Xu \& Durrett(2018)Xu and Durrett]{xu2018sphericalVAE}
Xu, J. and Durrett, G.
\newblock Spherical latent spaces for stable variational autoencoders.
\newblock In \emph{Proceedings of the 2018 Conference on Empirical Methods in
  Natural Language Processing}, 2018.

\end{thebibliography}
\bibliographystyle{icml2021}

\newpage
\appendix
\section{Symmetric Spaces: a Short Overview}
\label{app:appendix-a}

\begin{table*}[!t]
\centering
\adjustbox{max width=\textwidth}{
%\begin{table}[ht]
 %  \renewcommand*{\arraystretch}{1.5}
  %  \centering
\begin{tabular}{|l|l|l|l|l|}
\hline
Type&Non-compact&Compact&$\rk_\R$&$\dim$\\
\hline
AI&$\SL(n,\R)/\SO(n,\R)$&$\SU(n)/\SO(n)$& $n-1$&$\frac{(n-1)(n+2)}2$\\
\hline
A&$\SL(n,\C)/\SU(2)$&$(\SU(n)\times\SU(n))/\SU(n)$&$n-1$&$(n+1)(n-1)$\\
\hline
BDI&$\SO(p,q)/\SO(p)\times\SO(q)$&$\SO(p+q)/\SO(p)\times\SO(q)$&$\min\{p,q\}$&$p q$\\
\hline
AIII&$\SU(p,q)/\SU(p)\times\SU(q)$&$\SU(p+q)/\SU(p)\times\SU(q)$&$\min\{p,q\}$&$2p q$\\
\hline
CI&$\Sp(2n,\R)/\UU(n)$&$\Sp(2n)/\UU(n)$&$n$&$2n(n+1)$\\
\hline 
DIII&$\SO^*(2n)/\UU(n)$&$\SO(2n)/\UU(n)$&$\lfloor\frac n2\rfloor$&$n(n-1)$\\
\hline
CII&$\Sp(p,q)/\Sp(p)\times\Sp(q)$&$\Sp(p+q)/\Sp(p)\times\Sp(q)$&$\min\{p,q\}$&$4pq$\\
\hline
AII&$\SL(n,\HH)/\Sp(n)$&$\SU(2n)/\Sp(n)$&$n-1$&$(n-1)(2n+1)$\\
\hline
D&$\SO(2n,\C)/\SO(2n)$&$(\SO(2n)\times\SO(2n))/\SO(2n)$&$n$&$n(2n-1)$\\
\hline
B&$\SO(2n+1,\C)/\SO(2n+1)$&$(\SO(2n+1)\times\SO(2n+1))/\SO(2n+1)$&$n$&$n(2n+1)$\\
\hline
C&$\Sp(n,\C)/\Sp(n)$&$(\Sp(n)\times\Sp(n))/\Sp(n)$&$n$&\\
\hline
\end{tabular}}
\caption{The classical symmetric spaces. Row CI represents the Siegel spaces and their compact duals.}\label{tab.1}
%\end{table}
\end{table*}

Riemannian symmetric spaces have been extensively studied by mathematicians, and there are many ways to characterize them. They can be described as simply connected Riemannian manifolds, for which the curvature is covariantly constant, or Riemannian manifolds, for which the geodesic reflection in each point defines a global isometry of the space. A key consequence is that symmetric spaces are homogeneous manifolds, which means in particular that the neighbourhood of any point in the space looks the same, and moreover that they can be efficiently described by the theory of semisimple Lie groups. 

 To be more precise a \emph{symmetric space} is a Riemannian manifold $(M,g)$ such that for any point $p\in M$, the geodesic reflection at $p$ is induced by a global isometry of $M$. A direct consequence is that the group of isometries $\mathrm{Isom}(M,g)$ acts transitively on $M$, i.e. given $p,q \in M$ there exists $g \in \mathrm{Isom}(M,g)$ such that $g(p) = q$. 
Thus symmetric spaces are homogeneous manifolds, which means in particular that the neighbourhood of any point in the space looks the same.
This leads to an efficient description by the theory of semisimple Lie groups:   $M=\sf G/\sf K$ where $\sf G={\rm Isom}_0(M)$ and $\sf K$, a compact Lie group, is the stabilizer of a point $p \in M$.

\subsection{Classification}
Every symmetric space $(M,g)$ can be decomposed into an (almost) product $M = M_1 \times \cdots \times M_k$ of symmetric spaces. A symmetric space is \emph{irreducible}, if it cannot be further decomposed into a Riemannian product $M=M_1\times M_2$. We restrict our discussion to these fundamental building blocks, the irreducible symmetric spaces.

Irreducible symmetric spaces can be distinguished in two classes, the symmetric spaces of compact type, and the symmetric spaces of non-compact type, with an interesting {\bf duality} between them. 
Apart from twelve exceptional examples, there are eleven infinite families of pairs of symmetric spaces $X$ of compact and non-compact type, which we summarize in Table \ref{tab.1}.
We refer the reader to \citet{helgason1078diffGeom} for more details and a list of the exceptional examples. 

\begin{remark}
Observe that, due to isomorphisms in low dimensions, the first cases of each of the above series is a hyperbolic space (of the suitable dimension). Using this one can construct many natural hyperbolic spaces as totally geodesic submanifolds of the symmetric spaces above. We listed them in Table \ref{tab.2} for the reader's convenience. 
\end{remark}

{\bf Rank:} An important invariant of a symmetric space $M$ is its rank, which is the maximal dimension of an (isometrically embedded) Euclidean submanifold. In a rank $r$ non-compact symmetric space, such submanifolds are isometric to $\mathbb{R}^n$, and called \emph{maximal flats}.
In a compact symmetric space, they are compact Euclidean manifolds such as tori.

Some of the rich symmetry of symmetric spaces is visible in the distribution of flats.
As homogeneous spaces, each point of a symmetric space $M$ must lie in \emph{some} maximal flat, but in fact for every pair $p,q$ of points in $M$, one may find some maximal flat containing them.
The ability to move any pair of points into a fixed maximal flat by symmetries renders many quantities (such as the metric distances described below) computationally feasible.

\subsection{Duality}
Compactness provides a useful dichotomy for irreducible symmetric spaces.
Symmetric spaces of compact type are compact and of non-negative sectional curvature. The basic example being the sphere $S^n$. 
Symmetric spaces of non-compact type are non-compact, in fact they are homeomorphic to $\R^n$ and of non-positive sectional curvature. The basic example being the hyperbolic spaces $\mathbb{H}^n$. 

There is a duality between the symmetric spaces of non-compact type and those of compact type, pairing every noncompact symmetric space with its compact 'partner' or dual.

 \begin{figure}[h]
     \centering
    \includegraphics[width=0.4\textwidth]{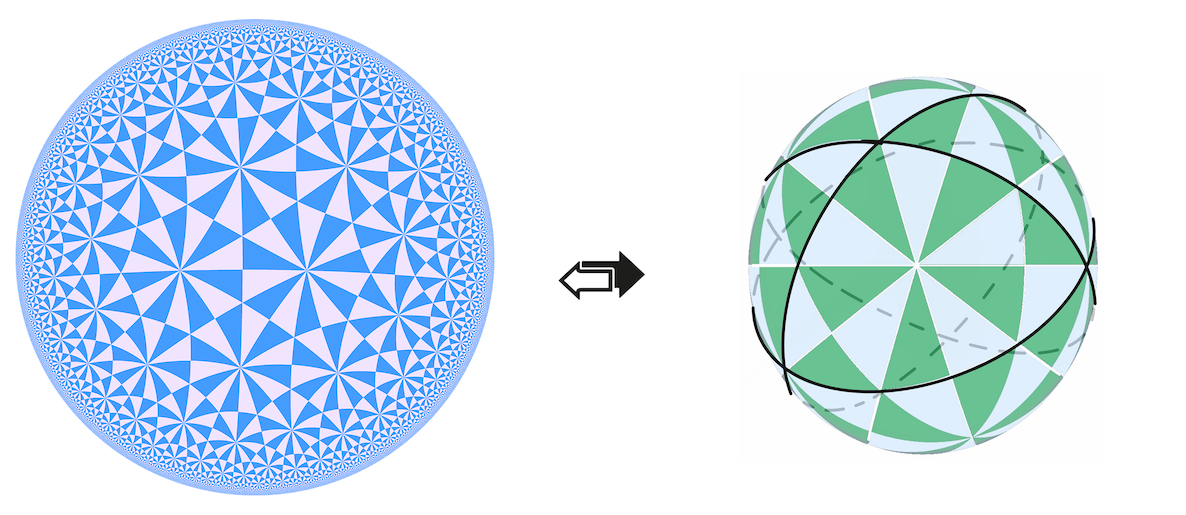}
     \caption{The duality between the hyperbolic plane and sphere is the basic example of the duality between symmetric spaces of compact and noncompact type.}
     \label{fig:Duality}
 \end{figure}

%The basic example of this duality pairs the two aforementioned examples: the compact dual to  hyperbolic $n$-space $\HH^n$ is the sphere $S^n$.
Duality for symmetric spaces generalizes the relationship between spheres and hyperbolic spaces, as well as between classical and hyperbolic trigonometric functions.
In the reference Table \ref{tab.1}, we provide for each family of symmetric spaces an explicit realization of both the noncompact symmetric space and its compact dual as coset spaces $\mathsf{G}/\mathsf{K}$.

% of compact and of non-compact type, depending on their sectional curvature, which is, respectively, non-negative and non-positive \cite[]{Helgason}. Among the two classes holds a duality,

\begin{table}[ht]
   \renewcommand*{\arraystretch}{1.5}
    \centering

\begin{tabular}{|l|l|l|l|}
\hline
Type&Parameters&Symmetric space& \\
\hline
AI&$n=2$&$\SL(2,\R)/\SO(2,\R)$&$\mathbb H^2$\\
\hline
A&$n=2$&$\SL(2,\C)/\SU(2)$&$\mathbb H^3$\\
\hline
BDI&$p=1$&$\SO(1,q)/\SO(q)$&$\mathbb H^q$\\
\hline
AIII&$p=1, q=1$&$\SU(1,1)/\SU(1)\times\SU(1)$&$\mathbb H^2$\\
\hline
CI&$n=1$&$\Sp(2,\R)/\UU(1)$&$\mathbb H^2$\\
\hline
DIII&$n=2$&$\SO^*(4)/\UU(1)$&$\mathbb H^2$\\
\hline
CII&$p=q=1$&$\Sp(1,1)/\Sp(1)\times\Sp(1)$&$\mathbb H^4$\\
\hline
AII&$n=1$&$\SL(2,\HH)/\Sp(1)$&$\mathbb H^5$\\
\hline
D&$n=1$&$\SO(2,\C)/\SO(2)$&$\mathbb R^*$\\
\hline
B&$n=1$&$\SO(3,\C)/\SO(3)$&$\mathbb H^3$\\
\hline
C&$n=1$&$\Sp(1,\C)/\Sp(1)$&$\mathbb H^3$\\
\hline
\end{tabular}
\caption{Hyperbolic spaces for low parameters}\label{tab.2}
\end{table}

\subsection{Vector-Valued Distance}
\label{a.VValuedDist}
The familiar geometric invariant of pairs of points is simply the distance between them.
For rank $n$ symmetric spaces, this one dimensional invariant is superseded by an $n$-dimensional invariant: the \emph{vector valued distance}.

Abstractly, one computes this invariant as follows: for a symmetric space $M$ with ${\rm Isom}_0(M)=G$, choose a distinguished basepoint $m\in M$, and let $K<G$ be the subgroup of symmetries fixing $m$.
Additionally choose a distinguished maximal flat $F\subset M$ containing $m$, and an identification of this flat with $\R^n$.
Given any pair of points $p,q\in M$, one may find an isometry $g\in G$ moving $p$ to $m$, and $q$ to some other point $g(q)=v\in F$ in the distinguished flat.
Under the identification of $F$ with $\R^n$, the difference vector $v-m$ is a vector-valued invariant of the original two points, and determines the vector valued distance. (In practice we may arrange so that $m$ is identified with ${\bf 0}$, so this difference is simply $v$).

In rank 1, the flat $F$ identifies with $\R^1$, and this difference vector $v-m$ with a number.
This number encodes all geometric information about the pair $(p,q)$ invariant under the symmetries of $M$.  Indeed, the distance from $p$ to $q$ is simply its absolute value!

In rank $n$, ``taking the absolute value" has an $n$-dimensional generalization, via a finite a finite group of symmetries of called the Weyl group.
This group $W<K$ acts on the flat $F$, and abstractly, \emph{the vector valued distance $\rm{vDist}(p,q)$ from $p$ to $q$ is this difference vector up to the action of the Weyl group}.  This vector valued distance $\rm{vDist}(p,q)$ is the complete invariant for pairs of points in $M$ - it contains all geometric information about the pair which is invariant under all symmetries.
In particular, given the vector valued distance $\rm{vDist}(p,q)$, the (Riemannian) distance from $p$ to $q$ is trivial to compute - it is given by the length of $\rm{vDist}(p,q)$ in $\R^n$.

The identification of $F$ with $\R^n$ makes this more explicit.  Here the Weyl group acts as a group of linear transformations, which divide $\R^n$ into a collection of conical fundamental domains for the action, known as Weyl chambers.
Choosing a fixed Weyl chamber $C$, we may use these symmetries to move our originally found difference vector $v-m$ into $C$.
The vector valued distance is this resulting vector $\rm{vDist}(p,q)\in C$.

 \begin{figure}[h]
     \centering
    \includegraphics[width=0.45\textwidth]{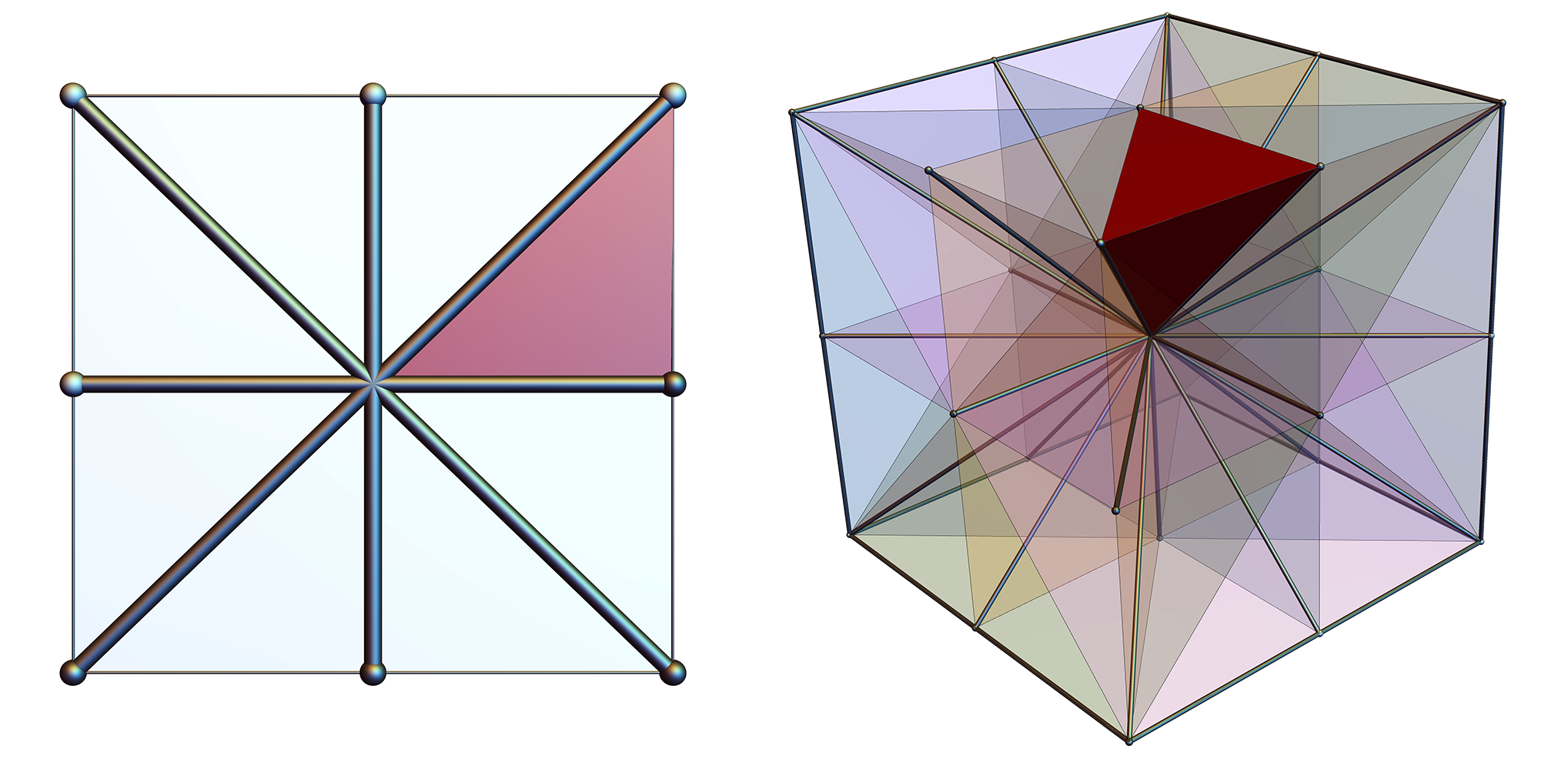}
     \caption{A choice of Weyl chamber the Siegel spaces of rank $n$ is given by $C=\{(v_i)\in\R^n\mid v_1\geq v_2\geq\cdots \geq v_n\geq 0\}$.  In rank 1, this is the nonnegative reals. Illustrated here are ranks $n=2,3$.}
     \label{fig:WeylChamber}
 \end{figure}

For example, in rank $n$ Siegel space, the Weyl group acts on $\R^2$ by the reflection symmetries of a cube, and a choice of Weyl chamber amounts to a choice of linear ordering of the vector components with respect to zero.  One choice is shown in Figure \ref{fig:WeylChamber}.  
In rank 2, this chamber is used to display the vector valued distances associated to edges and nodes of an embedded graph in Figures
\ref{fig:d-values-usca}-\ref{fig:d-values-tree}.
Note that once a Weyl chamber is picked it may be possible to find the vector valued distance corresponding to a vector in $\R^n$ without explicit use of the Weyl group: for the Siegel spaces this is by  sorting the vector components in nondecreasing order.

{\bf Computing Distance:}
The process for computing the vector valued distance is summarized below. It is  explicitly carried out for the Siegel spaces and their compact duals in Appendix \ref{a.Siegel}.

Let $M,G,K,F,m$ be as in the previous section.  Choose an identification $\phi\colon F\to\R^n$ which sends the basepoint $m$ to $\bf{0}$, and a Weyl chamber $C\subset\R^n$ for the Weyl group $W$.  For any pair of points $p,q\in M$;
\begin{enumerate}
    \item {\bf Move $p$ to the basepoint: }\\\noindent Compute $g\in G$ such that $g(p)=m$.
    \item {\bf Move $q$ into the flat:}\\\noindent 
    Compute $k\in K$ such that $k(g(q))\in F$. Now both $g(p)=m$ and $k(g(q))$ lie in the distinguished flat $F$.
    \item {\bf Identify the flat with $\R^n$: }\\\noindent
    Compute $u=\phi(k(g(q)))\in \R^n$. The points ${\bf 0}$ and $u$ represent $p,q$ after being moved into the flat, respectively.
    \item {\bf Return the Vector Valued Distance: }\\\noindent
    Compute $v\in C$ such that $v=Au$ for some element $A\in W$.  This is the vector valued distance $\rm{vDist}(p,q)$
    
\end{enumerate}
The {\bf Riemannian distance} is computed directly from the vector valued distance as its Euclidean norm, ${\rm dist}(p,q)=\|{\rm vDist}(p,q)\|$.

\subsection{Finsler Metrics}\label{a.finsler}

A Riemannian metric on a manifold $M$ is defined by a smooth choice of inner product on the tangent bundle.
Finsler metrics generalize this by requiring only a smoothly varying choice of norm $\|\cdot \|_F$.  
The length of a curve $\gamma$ is defined via integration of this norm along the path
$$\mathrm{Length}_F(\gamma)=\int_I\|\gamma'\|_F dt,$$
and the distance between points by the infimum of this over all rectifiable curves joining them 
$$d_F(p,q)=\inf\{\mathrm{Length}_F(\gamma)\mid \gamma(0)=p,\;\gamma(1)=q\}$$

The geometry of symmetric spaces allows the computation of Finsler distances, like much else, to take place in a chosen maximal flat.
On such flat spaces, the ability to identify all tangent spaces allow particularly simple Finsler metrics to be defined by choosing a single norm on $\R^n$.
We quickly review this theory below.

{\bf \noindent Finsler Metrics on $\R^n$:}
Any norm on $\R^n$ defines a Finsler metric.
As norms on a vector space are uniquely determined by their unit spheres, the data of a Finsler metric is given by a convex polytope $S$ containing $\bf{0}$.
An important example in this work is the $\ell^1$ Finsler metric on $\R^n$, given by the norm $\|(x_i)\|_{\ell^1}=\sum_i|x_i|$.  Its unit sphere is the boundary of the dual to the $n$-dimensional cube (in $\R^2$, this is again a square, but oriented at $45^{\circ}$ with respect to the coordinate axes).

Given such an $P$, the Finsler norm $\|v\|_F$ of a vector $v\in \R^n$ is the unique positive $\ell$ such that $\frac{1}{\ell}v\in\partial P$.
Figure \ref{fig:Taxicab} below shows the spheres of radius $1$ and $2$ with respect to the $\ell^1$ metric on the plane.

 \begin{figure}[h]
     \centering
    \includegraphics[width=0.4\textwidth]{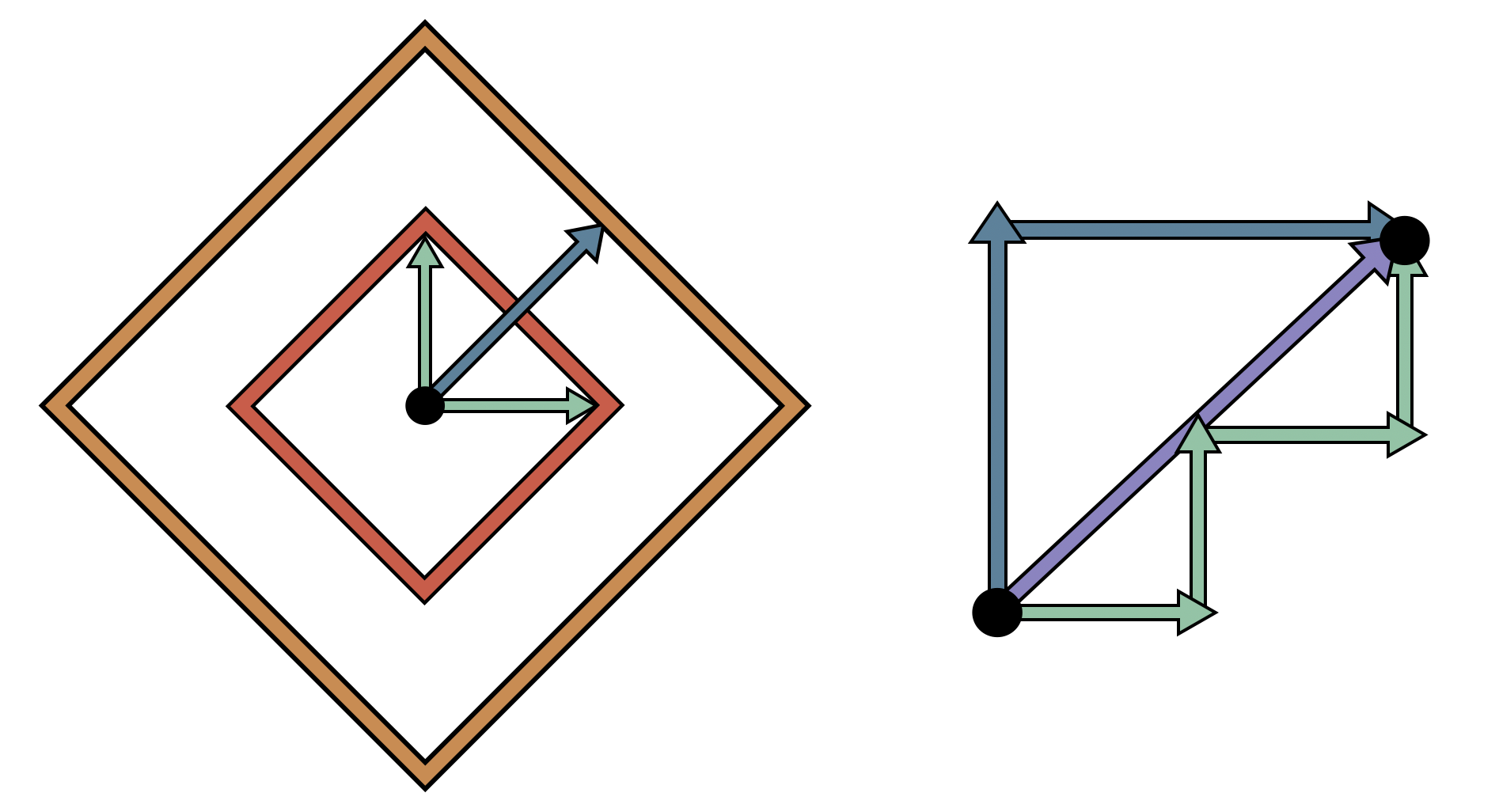}
     \caption{Left: Vectors of length 1 and 2 with respect to the $\ell^1$ norm on $\R^2$. Right: three geodesics of length 4 in the $\ell^1$ metric (to same scale as left image).}
     \label{fig:Taxicab}
 \end{figure}

While affine lines are geodesics in Finsler geometry, they need not be the unique geodesics between a pair of points.  Consider again Figure \ref{fig:Taxicab}: the vector sum of the two unit vectors in is exactly the diagonal vector, which lies on the $\ell^1$ sphere of radius $2$.
That is, in $\ell^1$ geometry traveling along the diagonal, or along the union of a vertical and horizontal side of a square both are distance minimizing paths of length 2.
The $\ell^1$ metric is often called the `taxicab' metric for this reason: much as in a city with a grid layout of streets, there are many shortest paths between a generic pair of points, as you may break your path into different choices of horizontal and vertical segments without changing its length.
See Figure \ref{fig:Finsler} in the main text for another example of this.

{\bf \noindent Finsler Metrics on Symmetric Spaces:}
To define a Finsler metric on a symmetric space $M$, it suffices to define it on a chosen maximal flat, and evaluate on arbitrary pairs of points with the help of the vector valued distance.
To induce a well defined Finsler metric $M$, a norm on this designated flat need only be invariant under the Weyl group $W$.
Said geometrically, the unit sphere of the norm $\|\cdot \|_F$ needs to contain it as a subgroup of its symmetries.
Given such a norm, the Finsler distance between two points is simply the Finsler norm of their vector valued distance 
$$d_F(p,q)=\|{\rm vDist}(p,q)\|_F.$$
Consequentially once the vector valued distance is known, any selection of Riemannian or Finsler distances may be computed at marginal additional cost.

 \begin{figure}[h]
     \centering
    \includegraphics[width=0.45\textwidth]{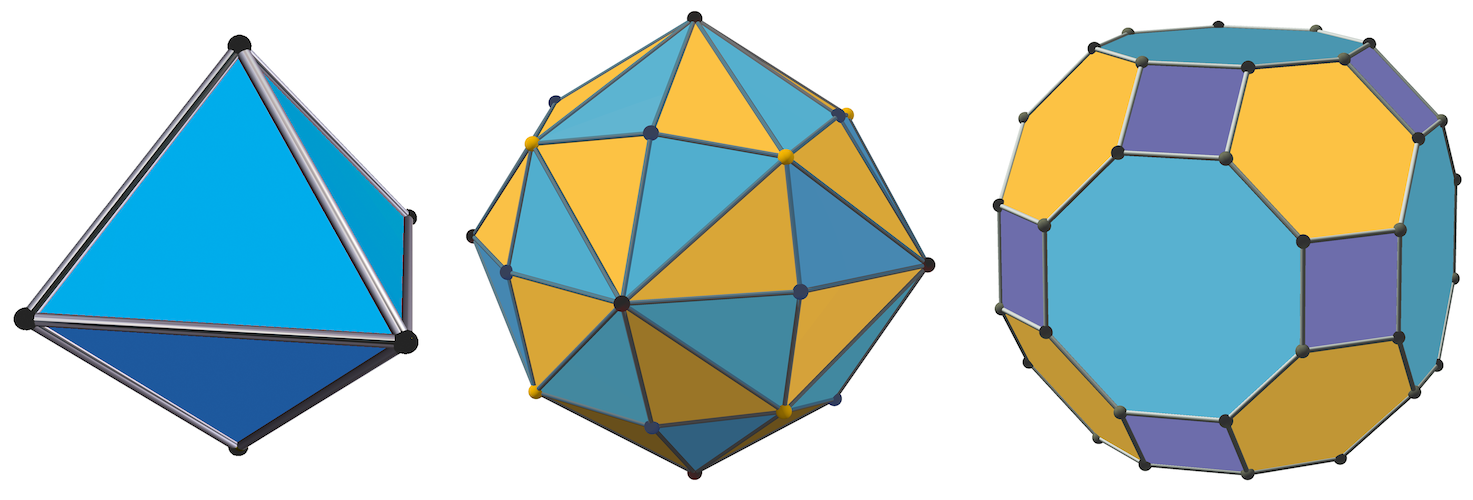}
     \caption{The unit spheres of several Finsler metrics on $\R^3$ invariant under the Weyl group of the rank 3 Siegel space.  The octahedron induces the $\ell^1$ metric.}
     \label{fig:FinslerNorms}
 \end{figure}

\subsection{Local Geometry for Riemannian Optimization}

Different Riemannian optimization methods require various input from the local geometry - here we describe a computation of the Riemannian gradient, parallel transport and the exponential map for general irreducible symmetric spaces.  

{\bf Riemannian Gradient}
Given a function $f\colon M\to\R$, the \emph{differential} of $f$ is a 1-form which measures how infinitesimal changes in the input affects (infinitesimally) the output.
More precisely at each point $p\in M$, $df$ is a linear functional on $T_pM$ sending a vector $v$ to the directional derivative $df_p(v)$ of $f$ in direction $v$.

In Euclidean space, this data is conveniently expressed as a vector: \emph{the gradient} $\nabla f$ defined such that $(\nabla f(p))\cdot v=df_p(v)$.
This extends directly to any Riemannian manifold, where the dot product is replaced with the Riemannian metric.  That is, the \emph{Riemannian gradient} of a function $f\colon M\to \R$ is the vector field $grad_R(f)$ on $M$ such that
$$g_p(\mathrm{grad}_R(f),v)=df_p(v)$$
for every $p\in M$, $v\in T_pM$.  Given a particular model (and thus, a particular coordinate system and metric tensor) one may use this implicit definition to give a formula for $\mathrm{grad}_R$.  See Appendix \ref{a.gradient} for an explicit example, deriving the Riemannian gradient for Siegel space from its metric tensor.

{\bf Parallel Transport}
\label{a.parallel}

While the lack of curvature in Euclidean space allows all tangent spaces to be identified, in general symmetric spaces the result of transporting a vector from one tangent space to another is a nontrivial, path dependent operation.
This \emph{parallel transport} assigns to a path $\gamma$ in $M$ from $p$ to $q$ an isomorphism $P_\gamma\colon T_pM\to T_qM$ interpreted as taking a vector $v\in T_pM$ at $p$ to $P_\gamma(v)\in T_q M$ by ``moving without turning" along $\gamma$.

The computation of parallel transport along geodesics in a symmetric space is possible directly from the isometry group.
To fix notation, for each $m\in M$ let $\sigma_m\in G$ be the geodesic reflection fixing $m$.
Let $\gamma$ be a geodesic in $M$ through $p$ at $t=0$.  As $t$ varies, the isometries $\tau_t=\sigma_{\gamma(t/2)}\circ\sigma_p$, called  \emph{transvections}, form the 1-parameter subgroup of translations along $\gamma$.
%A quick computation shows $\tau_t(\gamma(s))=\gamma(s+t)$.
If $p,q\in M$ are two points at distance $L$ apart along the the geodesic $\gamma$, the transvection $\tau_L$ takes $p$ to $q$, and its derivative $(d\tau_L)_p=P_\gamma\colon T_pM\to T_qM$ performs the parallel transport for $\gamma$.

% PICTURE?

% As an explicit example, consider the points $iI$ and $iY$ in the upper half space model of Siegel space, for $Y>>0$ a real symmetric matrix.  The transvection $\tau$ taking $iI$ to $iY$ along the geodesic $\gamma$ joining them is given by $Z\mapsto \sqrt{Y}Z\sqrt{Y}$, and so for a complex symmetric matrix $V$ thought of as a tangent vector to $iI$, the matrix $P_\gamma(V)=\sqrt{Y}V\sqrt{Y}$ is its parallel transport to the tangent space at $iY$.

{\bf The Exponential Map \& Lie Algebra}
The exponential map for a Riemannian manifold $M$ is the map $\exp\colon TM\to M$ such that if $v\in T_p(X)$,  $\exp(v)$ is the point in $M$ reached by traveling distance $\|v\|$ along the geodesic on $M$ through $p$ with initial direction parallel to $v$.  

When $M$ is a symmetric space with symmetry group $G$, this may be computed using the Lie group exponential $\exp\colon \mathfrak{g}\to G$ (the matrix exponential, when $G$ is a matrix Lie group).
Choose a point $p\in M$ and let $\sigma_p$ be the geodesic reflection in $p$.  Then $\sigma_p$ defines an involution $G\to G$ by $g\mapsto \sigma_p\circ g\circ \sigma_p$ (where composition is as isometries of $M$), and the eigenspaces of the differential of this involtuion give a decomposition $\mathfrak{g}=\mathfrak{k}\oplus\mathfrak{p}$ into the $+1$ eigenspace $\mathfrak{k}$ and the $-1$ eigenspace $\mathfrak{p}$.
Here $\mathfrak{k}$ is the Lie algebra of the stabilizer $K=\mathfrak{stab}(p)<G$, and so $\mathfrak{p}$ identifies with $T_pM$ under the differential of the quotient $G\to G/K\cong M$.

Let $\phi\colon T_pM \to \mathfrak{p}$ be the inverse of this identification.  Then for a vector $v\in T_pM$, we may find the point $q=\exp_p(v)\in M$ as follows:
\begin{enumerate}
    \item Compute $\phi(v)=A\in \mathfrak{p}$.  This is the tangent vector $v$, viewed as a matrix in the Lie algebra to $G$.
    \item Compute $g=\exp(A)$, where $\exp$ is the matrix exponential.
    \item Use the action of $G$ on $M$ by isometries to compute $q=g(p)$.
\end{enumerate}

\section{Explicit Formulas for Siegel Spaces}
\label{a.Siegel}
This section gives the calculations mathematics required to implement two models of Siegel space (the bounded domain model and upper half space) as well as a model of its compact dual.

\subsection{Linear Algebra Conventions}
A few clarifications from linear algebra can be useful:
\begin{enumerate}
\item The inverse of a matrix $X^{-1}$, the product of two matrices $XY$, the square $X^2$ of a square matrix are understood with respect to the matrix operations. Unless all matrices are diagonal these are different than doing the same operation to each entry of the matrix.
\item If $Z=X+iY$ is a complex matrix, 
\begin{itemize}
\item$Z^t$ denotes the transpose matrix, i.e.  $(Z^t)_{ij}=Z_{ji}$, 
\item $\ov Z=X-iY$ denotes the complex conjugate
\item $X^*$ denotes its transpose conjugate, i.e. $X^*=\ov{X^t}$.
\end{itemize}
\item A complex square matrix $Z$ is \emph{Hermitian} if $Z^*=Z$. In this case its eigenvalues are real and positive. It is \emph{unitary} if $Z^*=Z^{-1}$. In this case its eigenvalues are complex numbers of absolute value 1 (i.e. points in the unit circle).
\item If $X$ is a real symmetric, or complex Hermitian matrix, $X>\!\!>0$ means that $X$ is positive definite, equivalently all its eigenvalues are bigger than zero.
\end{enumerate}

\subsection{Takagi Factorization}\label{a.Takagi}
Given a complex symmetric matrix $A$, the Takagi factorization is an algorithm that computes a real diagonal matrix $D$ and a complex unitary matrix $K$ such that 
$$A=\ov K D K^*.$$
This will be useful to work with the bounded domain model. It is done in a few steps
\begin{enumerate}
	\item Find $Z_1$ unitary, $D$ real diagonal such that
	$$A^*A=Z_1^* D^2 Z_1$$
	\item Find $Z_2$ orthogonal, $B$ complex diagonal such that 
	$$\ov Z_1AZ_1^*=Z_2BZ_2^t$$
	This is possible since the real and imaginary parts of $\ov Z_1AZ_1^*$ are symmetric and commute, and are therefore diagonalizable in the same orthogonal basis.
	\item Set $Z_3$ be the diagonal matrix with entries
	$$(Z_3)_{ii}=\left(\sqrt\frac{b_i}{|b_i|}\right)^{-1}$$
	where $b_i=(B)_{ii}$
	\item Set $K=Z_1^*Z_2Z_3$, $D$ as in Step 1. It then holds 
	$$A=\ov K D K^*.$$
\end{enumerate}

\subsection{Siegel Space and its Models}\label{a.models}
We consider two models for the symmetric space, the bounded domain
$$\calB_n:=\{Z\in\Sym(n,\C)|\;\Id-Z^*Z>\!\!>0\}$$
and the upper half space
$$\calS_n:=\{X+iY\in\Sym(n,\C)|\;Y>\!\!>0\}.$$
An explicit isomorphism between the two domains is given by the Cayley transform
$$\begin{array}{cccc}
c:&\calB_n&\to&\calS_n\\
&Z&\mapsto&i(Z+\Id)(Z-\Id)^{-1}
\end{array}$$ 
whose inverse $c^{-1}=t$ is given by
$$\begin{array}{cccc}
t:&\calS_n&\to&\calB_n\\
&X&\mapsto&(X-i\Id)(X+i\Id)^{-1}
\end{array}$$

When needed, a choice of {\bf basepoint} for these models is $i\Id\in \calS_n$ for upper half space and the zero matrix ${\bf 0}\in \calB_n$ for the bounded domain.
A convenient choice of {\bf maximal flats} containing these basepoints are the subspaces $\{iD\mid D={\rm diag}(d_i),\;d_i>0\}\subset \calS_n$ and $\{D={\rm diag}(d_i)\mid d_i\in(-1,1)\}\subset\calB_n$.

The group of symmetries of the Siegel space $\calS_n$ is  
$\Sp(2n,\R)$, the subgroup of $\SL(2n,\R)$ preserving a symplectic form: a non-degenerate antisymmetric bilinear form on $\R^{2n}$. In this text we will choose the symplectic form represented, with respect to the standard basis, by the matrix $\bsm0&\Id_n\\-\Id_n&0\esm$ so that the symplectic group is given by the matrices that have the block expression
$$\left\{\bpm A&B\\C&D\epm\left|\begin{array}{l} A^tD-C^tB=\Id\\ A^tC=C^tA\\B^tD=D^tB\end{array}\right.\right\}$$
where $A,B,C,D$ are real $n\times n$ matrices.

The symplectic group $\Sp(2n,\R)$ acts on $\calS_n$ by non-commutative fractional linear transformations
$$\begin{pmatrix}A&B\\C&D\end{pmatrix}\cdot Z=(AZ+B)(CZ+D)^{-1}.$$
The action of $\Sp(2n,\R)$ on $\calB_n$ can be obtained through the Cayley transform.

\subsection{Computing the Vector-Valued Distance}\label{a.dist}

The Riemannian metric, as well as any desired Finsler distance, are computable directly from the vector-valued distance as explained in Appendix \ref{a.VValuedDist}.
Following those steps, we give an explicit implementation for the upper half space model below, and subsequently use the Cayley transform to extend this to the bounded domain model.

Given as input two points $Z_1,Z_2\in \calS_n$ we perform the following computations:

{\bf\noindent 1) Move $Z_1$ to the basepoint: }
 Compute the image of $Z_2$ under the transformation taking $Z_1$ to $iI$, defining
$$Z_3:=\sqrt{\Im Z_1}^{-1}(Z_2-\Re Z_1)\sqrt{\Im Z_1}^{-1}\in\calS_n$$

{\bf\noindent 2) Move $Z_2$ into the chosen flat: }
 Define
$$W=t(Z_3)\in\calB,$$
and use the Takagi factorization to write
$$W=\ov K D K^*$$ for some real diagonal matrix $D$ with eigenvalues between 0 and 1, and some unitary matrix $K$.
\emph{Note: to make computations easier, we are leveraging the geometry of both models here, so in fact $i(I+D)(I-D)^{-1}$ is the matrix lying in the standard flat containing $iI$.}

{\bf\noindent 3) Identify the flat with $\R^n$: }
 Define the vector $v=(v_i)\in\R^n$ with 
 $$v_i=\log\frac{1+d_i}{1-d_i},$$
 for $d_i$ the $i^{th}$ diagonal entry of the matrix $D$ from the last step.
 
 {\bf\noindent 4) Return the Vector Valued Distance: }
 Sort the absolute values of the entries of $v$ to be in nonincreasing order, and set
 $\mathrm{vDist}(Z_1,Z_2)$ equal to the resulting list.
$$
 \mathrm{vDist}=(|v_{i_1}|,|v_{i_2}|,\ldots,|v_{i_n}|)$$
 $$|v_{i_1}|\geq |v_{i_2}|\geq \cdots\geq |v_{i_n}|
$$
%  \emph{Note: if the goal is to only compute Riemanian or Finsler distances, this last step may be omitted, and the desired distance computed directly from the vector $v$.}
 
% \begin{enumerate}
% \item Compute $$Z_3:=\sqrt{\Im Z_1}^{-1}(Z_2-\Re Z_1)\sqrt{\Im Z_1}^{-1}\in\calS_n$$
% \item Define
% $$W=t(Z_3)\in\calB$$
% \item Use the Takagi factorization to write
% $$W=\ov K D K^*$$ for some real diagonal matrix $D$ with eigenvalues between 0 and 1, and some unitary matrix $K$.
% \item The distance is given by 
% $$d^R(Z_1,Z_2):=\sqrt {\sum_{i=1}^n \left[\log{\frac{1+d_i}{1-d_i}}\right]^2}$$
% Here, as before, $d_i$ denote the diagonal entries of the diagonal  matrix $D$.
% \end{enumerate}

{\bf \noindent Bounded domain:}
In this case, given $W_1,W_2\in\calB$ we consider the pair $Z_1,Z_2\in\calS_n$ obtained applying the Cayley transform $Z_i=t (W_i)$. Then we can apply the previous algorithm, indeed 
$$\mathrm{vDist}(W_1,W_2)=\mathrm{vDist}(Z_1,Z_2).$$

\subsection{Riemannian \& Finsler Distances}

The Riemannian distance between two points $X,Y$ in the Siegel space (either the upper half space or bounded domain model) is induced by the Euclidean metric on its maximal flats.  This is calculable directly from the vector valued distance $\mathrm{vDist}(X,Y)=(v_1,v_2,\ldots,v_n)$ as
$$d^R(X,Y)=\sqrt {\sum_{i=1}^n v_i^2}.$$

The Weyl group for the rank $n$ Siegel space is the symmetry group of the $n$ cube.  Thus, any Finsler metric on $\R^n$ whose unit sphere has these symmetries has these symmetries induces a Finsler metric on Siegel space.
The class of such finsler metrics includes many well-known examples such as the $\ell^p$ metrics
$$\|(v_1,\ldots,v_n)\|_{\ell^p}=\left(\sum_i |v_i|^p\right)^{\frac{1}{p}},$$
which is one of the reasons the Siegel space is an attractive avenue for experimentation.

Of particular interest are the $\ell^1$ and $\ell^\infty$ Finsler metrics.  The distance functions induced on the Siegel space by them are given below
$$d^{F_1}(X,Y)=\sum_{i=1}^n v_i \qquad d^{F_\infty}(X,Y)=v_1.$$
Where $X,Y$ are points in Siegel space (again, either in the upper half space or bounded domain models), and the $v_i$ are the component of the vector valued distance $\mathrm{vDist}(X,Y)=(v_1,v_2,\ldots, v_n)$.

There are explicit bounds between the distances, for example 
\begin{align}\label{e.dist}\frac 1{\sqrt n} d^{F_1}(X,Y)\leq d^R(X,Y)\leq d^{F_1}(X,Y)\end{align}
Furthermore, we have 
\begin{equation}
\begin{split}
d^{F_1}(X,Y)=\log\det(\sqrt{R(X,Y)}+\Id)- \\ 
\log\det(\Id-\sqrt{R(X,Y)})        
\end{split}
\end{equation}
which, in turn, allows to estimate the Riemannian distance using \eqref{e.dist}.

\subsection{Riemannian Gradient}\label{a.gradient}
We consider on $\Sym(n,\C)$ the Euclidean metric given by 
$$\|V\|_E^2=\tr(V\ov V),$$
here $\tr$ denotes the trace, and, as above,  $V\ov V$ denotes the matrix product of the matrix $V$ and its conjugate. 

{\bf \noindent Siegel upperhalf space:}
The Riemannian metric at a point $Z \in \calS_n$, where $Z = X + iY$ is given by \cite{siegel1943symplectic} 
$$\|V\|_R^2=\tr(Y^{-1}VY^{-1}\ov V).$$
As a result we deduce that 
$${\rm grad}(f(Z)) = Y \cdot {\rm grad_E}(f(Z)) \cdot Y$$
{\bf \noindent Bounded domain:}
In this case we have 
$${\rm grad}(f(Z)) = A \cdot {\rm grad_E}(f(Z)) \cdot A$$
where $A = \Id - \overline{Z}Z$

\subsection{Embedding Initialization}\label{a.random}
Different embeddings methods initialize the points close to a fixed basepoint. In this manner, no a priori bias is introduced in the model, since all the embeddings start with similar values.

We use the {\bf basepoints} specified previously: $i\Id$ for Siegel upper half space and ${\bf 0}$ for the bounded domain model.

In order to produce a random point we generate a random symmetric matrix with small entries and add it to our basepoint. As soon as all entries of the perturbation are smaller than $1/n$ the resulting matrix necessarily belongs to the model.
In our experiments, we generate random symmetric matrices with entries taken from a uniform distribution $\mathcal{U}(-0.001, 0.001)$.

\subsection{Projecting Back to the Models}\label{a.project}
The goal of this section is to explain two algorithms that, given $\epsilon$ and a point $Z\in\Sym(n,\C)$, return a point $Z_\epsilon^\calS$ (resp. $Z_\epsilon^\calB$), a point close to the original point lying in the $\epsilon$-interior of the model. This is the equivalent of the projection applied in \citet{nickel2017poincare} to constrain the embeddings to remain within the Poincar\'e ball, but adapted to the structure of the model. Observe that the projections are not conjugated through the Cayley transform.

{\bf \noindent Siegel upperhalf space:}
In the case of the Siegel upperhalf space $\calS_n$ given a point $Z=X+iY\in\Sym(n,\C)$
\begin{enumerate}
\item Find a real $n$-dimensional diagonal matrix $D$ and an orthogonal matrix $K$ such that $$Y=K^tDK$$
\item Compute the diagonal matrix $D_\epsilon$ with the property that
$$(D_\epsilon)_{ii}=\begin{cases} D_{ii} &\text{ if } D_{ii}>\epsilon\\\epsilon &\text{ otherwise}\end{cases}$$
\item The projection is given by
$$Z_\epsilon^\calS:=X+i K^tD_\epsilon K$$
\end{enumerate}

{\bf \noindent Bounded Domain:} In the case of the bounded domain $\calB$ given a point $Z=X+iY\in\Sym(n,\C)$
\begin{enumerate}
\item Use the Takagi factorization to find a real $n$-dimensional diagonal matrix $D$ and an unitary matrix $K$ such that $$Y=\ov KDK^*$$
\item Compute the diagonal matrix $D^\calB_\epsilon$ with the property that
$$(D_\epsilon^\calB)_{ii}=
\begin{cases} 
D_{ii} &\text{ if } D_{ii}<1-\epsilon\\
1-\epsilon &\text{ otherwise
}\end{cases}$$
\item The projection is given by
$$Z_\epsilon^\calB:=\ov KD^{\calB}_\epsilon K^*$$
\end{enumerate}

\subsection{Crossratio and Distance}\label{a.cr}

Given two points $X,Y$ in Siegel space, there is an alternative means of calculating the vector valued distance (and thus any Riemannian or Finsler distance one wishes) via an invariant known as the cross ratio.

{\bf \noindent Siegel upperhalf space:}
Given two points $X,Y\in\calS_n$ their crossratio is given by the complex $n\times n$-matrix 
$$R_\calS(X,Y)=(X-Y)(X-\ov Y)^{-1}(\ov X-\ov Y)(\ov X-Y)^{-1}.$$

It was established by Siegel \cite{siegel1943symplectic} %(see also Burger-P.\cite[Section ]{BP}) 
that if $r_1,\ldots, r_n$ denote the eigenvalues of $R$ (which  are necessarily real greater than or equal to 1)
and we denote by $v_i$ the numbers
$$v_i=\log{\frac{1 + \sqrt r_i}{1-\sqrt r_i}}$$
then the $v_i$ are the components of the vector-valued distandce $\mathrm{vDist}(X,Y)$.
Thus, the Riemannian distance is
$$d^R(X,Y)=\sqrt {\sum_{i=1}^n v_i^2}.$$
The Finsler distances $d^{F_1}$ and $d^{F_\infty}$ are likewise given by the same formulas as previously.

In general it is computationally difficult to compute the eigenvalues, or the squareroot, of a general complex matrix. However, we can use the determinant $\det_R$ of the matrix $R(X,Y)$ to give a lower bound on the distance:
$$\log{\frac{1 + \sqrt {\det_R}}{1-\sqrt {\det_R}}} \leq d^R(X,Y).$$

{\bf \noindent Bounded domain:}
The same study applies to pairs of points  $X,Y\in\calB$, but their crossratio should be replaced by the expression
\begin{equation}
\begin{split}
R_\calB(X,Y)=(X-Y)(X-\ov {Y^{-1}})^{-1} \\
(\ov{ X^{-1}}-\ov {Y^{-1}})(\ov{ X^{-1}}-Y)^{-1}    
\end{split}
\end{equation}

\subsection{The Compact Dual of the Siegel Space}\label{a.cpt}

The compact dual to the (non-positively) curved Siegel space is a compact non-negatively curved symmetric space; in rank $1$ this is just the 2-sphere.
Many computations in the compact dual are analogous to those for the Siegel spaces, and are presented below.

\subsubsection*{Model}
Abstractly, the compact dual is the space of complex structures on quaternionic $n$-dimensional space compatible with a fixed inner product.
%This is a closed Riemannian manifold; of finite diameter and with no boundary.
It is convenient to work with a coordinate chart, or \emph{affine path} covering all but a measure zero subset of this space.  We denote this patch by $\calD_n$, which consists of all $n\times n$ complex symmetric matrices:
$$\calD_n=\Sym(n;\mathbb{C})$$
With this choice of model, tangent vectors to $\calD_n$ are also represented by complex symmetric matrices.  More precisely, for each $W\in \calD_n$ we may identify the tangent space $T_W\calD_n$ with $\Sym(n,\C)$.

{\bf \noindent Basepoint:}
The basepoint of $\calD_n$ is the zero matrix $\bf{0}$.

{\bf \noindent Maximal Flat:}
A useful choice of maximal flat is the subspace of real diagonal matrices.

{\bf \noindent Projection:}
The model $\calD_n$ is a linear subspace of the space of $n\times n$ complex matrices.  Orthogonal projection onto this subspace is given by symmetrization,
$$W\mapsto \frac{1}{2}(W+W^t).$$

{\bf\noindent Isometries:}
The symmetries of the compact dual are given by the compact symplectic group $\Sp(n)$.  With respect to the model $\calD_n$, we may realize this as the intersection of the complex symplectic group $\Sp(2n,\C)$ and the unitary group $U(2n,\C)$

$$\left\{\bpm A&B\\C&D\epm\left|\begin{array}{l} A^tD-C^tB=\Id\\ A^tC=C^tA\\B^tD=D^tB\\
A^\ast A+C^\ast C=\Id\\
B^\ast B+D^\ast D=\Id\\
A^\ast B+C^\ast D=0
\end{array}\right.\right\}$$
where $A,B,C,D$ are \emph{complex} $n\times n$ matrices.
The first four conditions are analogs of those defining $\Sp(2n;\R)$, and the final three come from the defining property that a unitary matrix $M$ satisfies $M^\ast M=\Id$.

This group acts on $\calD_n$ by non-commutative fractional linear transformations
$$\begin{pmatrix}A&B\\C&D\end{pmatrix}\cdot W=(AW+B)(CW+D)^{-1}.$$

{\bf \noindent Riemannian Metric \& Gradient:}
The Riemannian metric at a point $W\in\calD_n$ is given by
  $$\langle U,V\rangle_W=(\Id+\overline{W}W)^{-1}U(\Id+\overline{W}W)^{-1}\overline{V},$$
where $U,V$ are tangent vectors at $W$.

The gradient of a function on the compact dual can be written in terms of its Euclidean gradient, via a formula very similar to that for the Bounded Domain model of the Siegel space.
In this case we have 
$${\rm grad}(f(W)) = A \cdot {\rm grad_E}(f(W)) \cdot A$$
where (the only difference from the bounded domain version being that the $-$ sign in the definition of $A$ has been replaced with a $+$).

\subsubsection*{Vector Valued Distance}
% This allows a direct computation of the distance between the basepoint $\bf{0}$ and a real diagonal matrix $D\in \calD_n:$

% $$\mathrm{dist}({\bf 0},D)=\sqrt{\sum_{i} \arctan(d_i)^2}$$

% \noindent where $d_i$ are the diagonal entries of the diagonal matrix $D$.

We again give an explicit implementation of the abstract procedure described in Appendix \ref{a.VValuedDist}, to calculate the vector valued distance associated to an arbitrary pair $W_1,W_2\in\calD_n$ as follows:

{\bf\noindent Move $W1$ to the basepoint:}
 \begin{enumerate}
 \item Use the Takagi factorization to write 
 $$W_1=\overline{U}PU^\ast$$
 for a unitary matrix $U$ and real diagonal matrix $P$.
 \item 	From $P$, we build the diagonal matrix $A=(\Id+P^2)^{-1/2}$.  That is, the diagonal entries of $A$ are $a_i=\frac{1}{\sqrt{1+p_i^2}}$ for $p_i$ the diagonal entries of $P$.
 \item From $A,U$ we build the following elements $M,R\in\Sp(n)$ of the compact symplectic group:
$$M=\begin{pmatrix}A&-AP\\AP&A\end{pmatrix}
\hspace{0.5cm}
R=\begin{pmatrix}U^t&0\\0&U^\ast\end{pmatrix}$$
 \end{enumerate}
 We now use the transformation $M\cdot R$ to move the pair $(W_1,W_2)$ to a pair $({\bf 0}, Z)$.  Because $W_1$ ends at the basepoint by construction, we focus on $W_2$.
\begin{enumerate}
 \setcounter{enumi}{3}	
 \item Compute $X=R.W_2$, that is $X=U^tW_2U$.
 \item Compute $Z=M.X$, that is $Z=(AX-AP)(APX-A)^{-1}$.  Alternatively, this simplifies to the conjugation $Z=AYA^{-1}$ by $A$ of the matrix $Y=(X-P)(PX-\Id)^{-1}$ 
\end{enumerate}

{\bf\noindent Move $Z$ into the chosen flat:}
% \noindent As these transformations are symmetries of the space, $\mathrm{dist}(W_1,W_2)=\mathrm{dist}({\bf 0},Z)$ so it suffices to compute the disatance from the basepoint to $Z$.
Use the Takagi factorization to write
 $$Z=\overline{K}DK^\ast$$
 \noindent for a unitary matrix $K$ and real diagonal matrix $D$.

 {\bf\noindent Identify the Flat with $\R^n$:}
 Produce from $D$ the $n$-vector 
 $$v=(\arctan(d_1),\ldots \arctan(d_n))$$ 
 Where $(d_1,\ldots d_n)$ are the diagonal entries of $D$.

 {\bf\noindent Return the Vector Valued Distance:}
 Order the the entries of $v$ in nondecreasing order.  This is the vector valued distance.
 $$
 \mathrm{vDist}=(v_{i_1},v_{i_2},\ldots,v_{i_n})$$
 $$v_{i_1}\geq v_{i_2}\geq \cdots\geq v_{i_n}\geq 0
$$

\subsubsection*{Riemannian and Finsler Distances:}

 The Riemannian distance between two points $X,Y$ in the compact dual is calculable directly from the vector valued distance $\mathrm{vDist}(X,Y)=(v_1,v_2,\ldots,v_n)$ as
$$d^R(X,Y)=\sqrt {\sum_{i=1}^n v_i^2}.$$

The Weyl group for the compact dual is the same as for Seigel space, the symmetries of the $n$-cube.
Thus the same collection of Finsler metrics induce distance functions on the compact dual, and their formulas in terms of the vector valued distance are unchanged.  

$$d^{F_1}(X,Y)=\sum_{i=1}^n v_i \qquad d^{F_\infty}(X,Y)=v_1.$$

\subsection{Interpolation between Siegel Space and its Compact Dual}\label{a.transition}

\begin{figure}[h]
     \centering
    \includegraphics[width=0.35\textwidth]{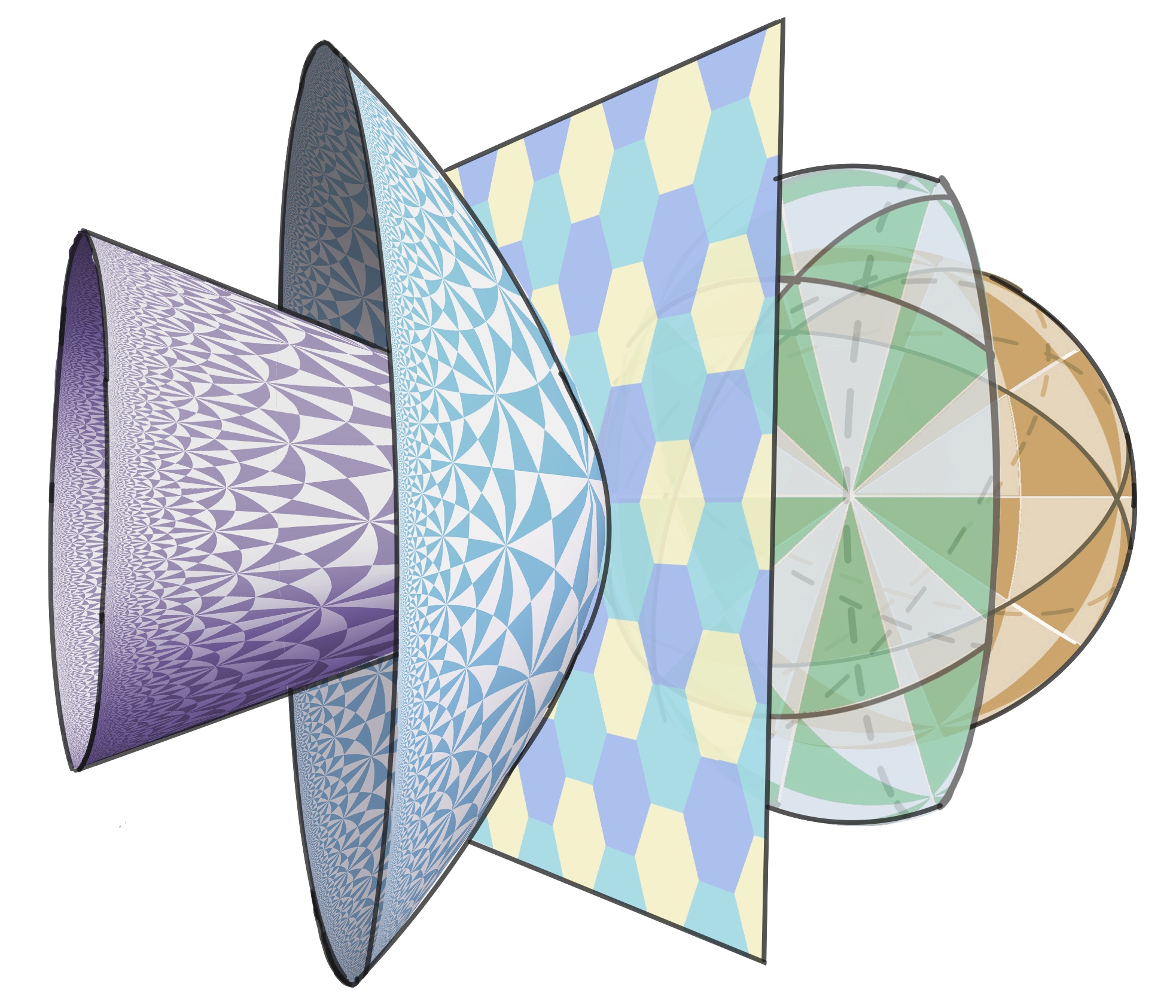}
     \caption{A 1-parameter family of spaces interpolating between the Siegel space and its compact dual, here illustrated in rank 1 ($\HH^2$ transitioning $S^2$ through the Euclidean plane with $k=0$).}
     \label{fig:Transition}
 \end{figure}

The Siegel space and its compact dual are part of a 1-parameter family of spaces indexed by a real parameter $k\in\R$.
When $n=1$ the symmetric spaces are two-dimensional, and this $k$ is interpreted as their (constant) curvature.
That is, this family represents an interpolation between the hyperbolic plane ($k=-1$) and the sphere ($k=1$) through Euclidean space $(k=0)$ as schematically represented in Figure \ref{fig:Transition}.
Below we describe the generalization of this to all $n$, by giving the model, symmetries, and distance functions in terms of the parameter $k\in\R$.

{\bf \noindent Model:}
Our models are most similar to the Bounded domain model of Siegel space, and so we use notation to match.
For each $k\in\R$ we define the subset $\calB_n^k$ of $\Sym(n,\C)$ as follows:
$$\calB_n^k=\left\{\begin{matrix}
\{W\mid \Id+k W^\ast W>>0\}& k<0\\
\Sym(n,\C)&k\geq 0
\end{matrix}\right.$$
The {\bf basepoint} for $\calB_n^k$ is the zero matrix ${\bf 0}$ for all $k$.
{\bf Projection back to the model} is analogous to what is done for the bounded domain when $k<0$, and is just symmetrization for $k\geq 0$.

{\bf\noindent Isometries:}
 Denote by $G^k$ the isometry group of $\calB_n^k$.
 A uniform description of $G^k$ can be given in close analogy to the description of the symmetries of the compact dual.
 For each $k\in\R$, $G^k=\Sp(2n,\C)\cap \mathcal{U}^k$ where $\mathcal{U}^k$ is a generalization of the usual unitary group
 $$\mathcal{U}^k=\left\{M\mid M^\ast\left(\begin{smallmatrix}k\Id &0\\0&\Id\end{smallmatrix}\right)M=\left(\begin{smallmatrix}k\Id &0\\0&\Id\end{smallmatrix}\right)\right\}$$

{\bf\noindent Riemannian Geometry:}
The Riemannian metric at a point $W\in\calB_n^k$ is given by the formula
$$\langle U,V\rangle_W^k=\tr\left(A^{-1}UA^{-1}\overline{V}\right)$$
Where $A=\Id+k\overline{W}W$.  
As before, this allows us to compute the {\bf Riemannian Gradient} in terms of the Euclidean gradient on $\calB^k_n$:
$${\rm grad}(f(W)) = A\cdot {\rm grad_E}(f(W)) \cdot A$$
From the Riemannian metric we may explicitly compute the distance function from the basepoint ${\bf 0}$ to a real diagonal matrix $D\in \calB_n^k$:
$$\mathrm{dist}^k({\bf 0}, D)=\left\{
\begin{matrix}\sqrt{\sum_i \frac{\mathrm{arctanh}\left(d_i\sqrt{|k|}\right)^2}{\sqrt{|k|}}}&k<0\\
\sqrt{\sum_id_i^2}&k=0\\
\sqrt{\sum_i \frac{\arctan\left(d_i\sqrt{k}\right)^2}{\sqrt{k}}}&k>0
\end{matrix}
\right.$$

{\bf \noindent Distance:}
The seven step procedure for calculating distance in the compact dual can be modified to give a procedure for the distance in $\calB_n^k$.
To calculate the Riemannian distance, Step 7 must be replaced with the distance formula above.
The only other changes involve the construction of the matrix called $M$ 
\begin{itemize}
\item In step 2, the computation of $P$ is unchanged but $A$ is replaced with $A=(\Id+kP^2)^{-1/2}$.
\item In step 3, the matrix $M$ is replaced with $$M=\begin{pmatrix}A&-AP\\\mathrm{sgn}(k)AP&A\end{pmatrix}$$

\end{itemize}

Where $\mathrm{sgn}(k)=0$ if $k=0$.  Note the computation of $M.X$ in Step 5 also changes, as $M$ has changed.  Now $M.X=(AX-AP)(\mathrm{sgn}(k)APX-A)^{-1}$.

\subsection{Experiments on the Compact Dual}
\label{sec:compact-dual-results}
We perform experiments on small synthetic datasets to compare the performance of the dual model to the upper half Siegel space and the Bounded domain model.
Results are reported in Table~\ref{tab:dual-cmp}. We can observe the reduced representation capabilities of the compact dual model, even on small datasets.

\begin{table}[!t]
\centering
\adjustbox{max width=\linewidth}{
\begin{tabular}{crrrr}
\toprule
\multicolumn{1}{l}{} & \multicolumn{2}{c}{\textsc{2D Grid}} & \multicolumn{2}{c}{\textsc{Tree}} \\
$(|V|, |E|)$ & \multicolumn{2}{c}{$(36, 60)$} & \multicolumn{2}{c}{$(40, 39)$} \\
\multicolumn{1}{l}{} & \multicolumn{1}{l}{$D_{avg}$} & \multicolumn{1}{l}{mAP} & \multicolumn{1}{l}{$D_{avg}$} & \multicolumn{1}{l}{mAP} \\
\cmidrule(lr){2-3}\cmidrule(lr){4-5}
$\mathcal S_{3}^R$ & 12.29 & 100.00 & 4.27 & 95.00 \\
$\mathcal S_{3}^{F_{\infty}}$ & 0.21 & 100.00 & \textbf{2.01} & 100.00 \\
$\mathcal S_{3}^{F_{1}}$ & 0.02 & 100.00 & 2.10 & 100.00 \\
$\mathcal B_{3}^R$ & 12.26 & 100.00 & 4.14 & 94.17 \\
$\mathcal B_{3}^{F_{\infty}}$ & 0.29 & 100.00 & 2.04 & 100.00 \\
$\mathcal B_{3}^{F_{1}}$ & \textbf{0.01} & 100.00 & 2.06 & 99.17 \\
$\mathcal D_{3}^R$ & 47.59 & 54.35 & 69.65 & 29.05 \\
$\mathcal D_{3}^{F_{\infty}}$ & 63.85 & 18.94 & 75.33 & 15.18 \\
$\mathcal D_{3}^{F_{1}}$ & 28.68 & 82.96 & 38.84 & 55.28 \\
\midrule
$\mathcal S_{4}^R$ & 12.27 & 100.00 & 4.20 & 98.33 \\
$\mathcal S_{4}^{F_{\infty}}$ & 0.49 & 100.00 & 1.72 & 100.00 \\
$\mathcal S_{4}^{F_{1}}$ & \textbf{0.01} & 100.00 & 1.58 & 100.00 \\
$\mathcal B_{4}^R$ & 12.24 & 100.00 & 4.10 & 100.00 \\
$\mathcal B_{4}^{F_{\infty}}$ & 0.17 & 100.00 & \textbf{1.18} & 100.00 \\
$\mathcal B_{4}^{F_{1}}$ & \textbf{0.01} & 100.00 & 1.48 & 100.00 \\
$\mathcal D_{4}^R$ & 41.82 & 78.20 & 65.95 & 31.76 \\
$\mathcal D_{4}^{F_{\infty}}$ & 53.31 & 79.34 & 74.19 & 19.16 \\
$\mathcal D_{4}^{F_{1}}$ & 13.38 & 100.00 & 23.64 & 71.94 \\
\bottomrule
\end{tabular}
}
\caption{Comparison of the compact dual model to the upper half space and bounded domain model on two synthetic datasets.}
\label{tab:dual-cmp}
\end{table}

\section{Experimental Setup}
\label{sec:appendix-expdetails}

\subsection{Implementation of Complex Operations}

All models and experiments were implemented in PyTorch \cite{paszke2019pytorchNeurips} with distributed data parallelism, for high performance on clusters of CPUs/GPUs.

Given a complex matrix $Z \in \C^{n \times n}$, we model real and imaginary components $Z = X + iY$ with $X, Y \in \R^{n \times n}$ separate matrices with real entries.
We followed standard complex math to implement basic arithmetic matrix operations. For complex matrix inversion we implemented the procedure detailed in \citet{falkenberg2007matrixinverse}.

\paragraph{Hardware:}
All experiments were run on Intel Cascade Lake CPUs, with microprocessors Intel Xeon Gold 6230 (20 Cores, 40 Threads, 2.1 GHz, 28MB Cache, 125W TDP). Although the code supports GPUs, we did not utilize them due to higher availability of CPU's.

\subsection{Optimization}
\label{sec:appendix-optimization}

As stated before, the models under consideration are Riemannian manifolds, therefore they can be optimized via stochastic Riemannian optimization methods such as \textsc{Rsgd} \cite{bonnabel2011rsgd} (we adapt the Geoopt implementation \cite{geoopt2019kochurov}).
Given a function $f(\theta)$ defined over the set of embeddings (parameters) $\theta$ and let $\nabla_{R}$ denote the Riemannian gradient of $f(\theta)$, the parameter update according to \textsc{Rsgd} is of the form:
$$\theta_{t + 1} = \mathcal{R}_{\theta_{t}}(-\eta_{t} \nabla_{R}f(\theta_{t}))$$

where $\mathcal{R}_{\theta_{t}}$ denotes the retraction onto space at $\theta$ and $\eta_{t}$ denotes the learning rate at time $t$. Hence, to apply this type of optimization we require the Riemannian gradient (described in Appendix~\ref{a.gradient}) and a suitable retraction.

\begin{figure*}[!t]
	\centering
	\subfloat{\label{fig:prod-cart-treegrid}{\includegraphics[clip, width=.25\linewidth,keepaspectratio]{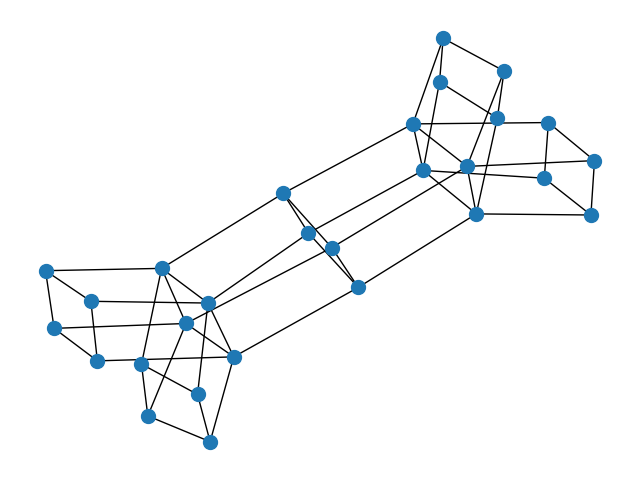}}}\hfill
	\subfloat{\label{fig:prod-cart-treetree}{\includegraphics[clip, width=.25\linewidth,keepaspectratio]{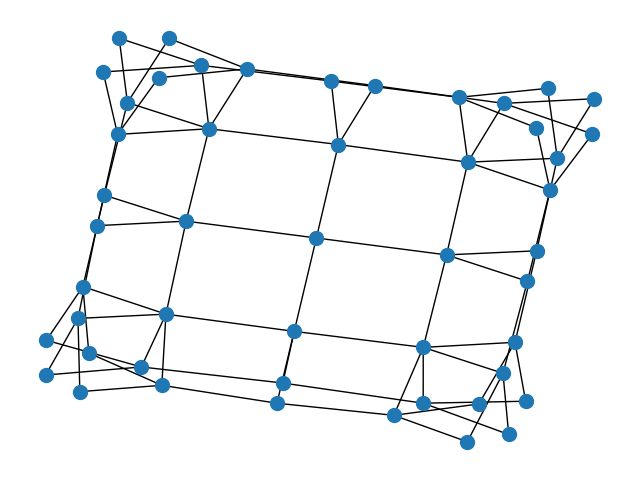}}}\hfill
	\subfloat{\label{fig:prod-root-treegrids}{\includegraphics[clip, width=.25\linewidth,keepaspectratio]{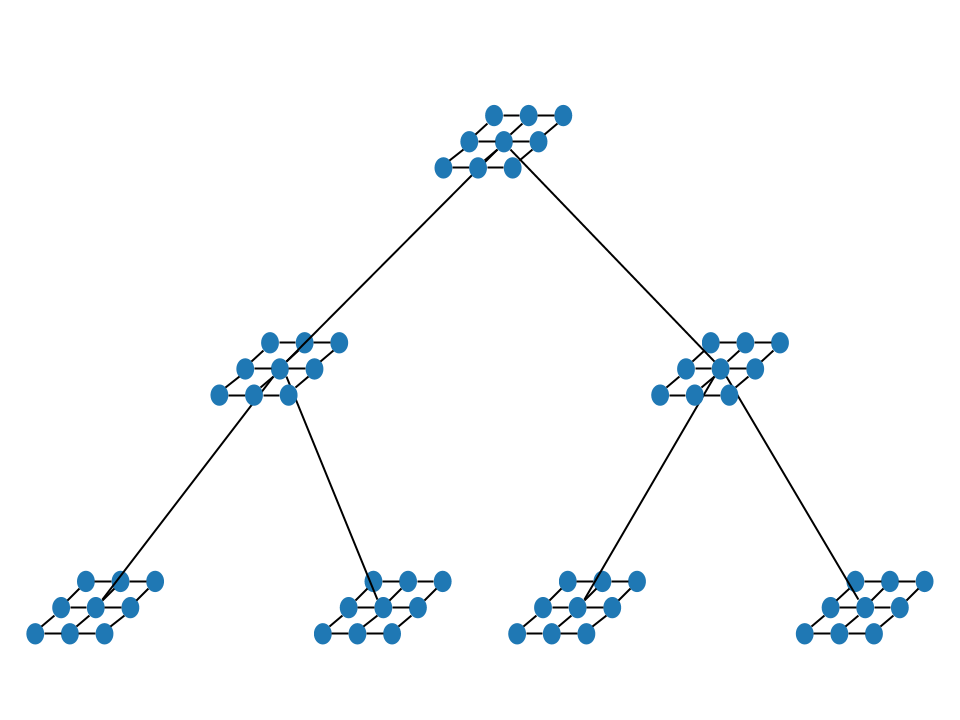}}}
	\subfloat{\label{fig:prod-root-gridtrees}{\includegraphics[clip, width=.25\linewidth,keepaspectratio]{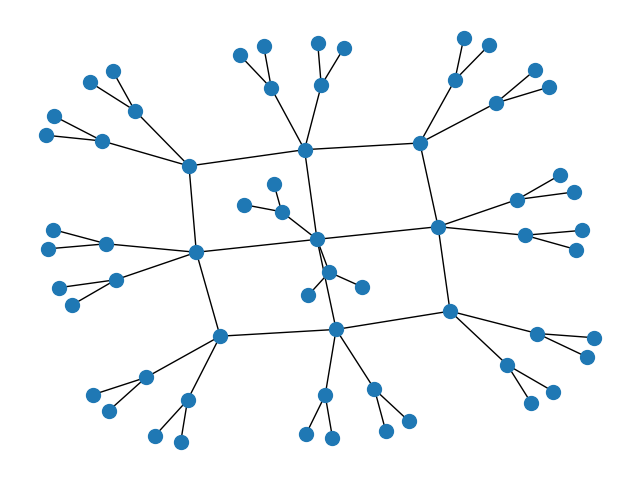}}}
	\caption{a) Cartesian product of tree and 2D grid. b) Cartesian product of tree and tree. c) Rooted product of tree and 2D grids. d) Rooted product of 2D grid and trees.}
	\label{fig:synthetic-graphs}
\end{figure*}

\textbf{Retraction:}
Following \citet{nickel2017poincare} we experiment with a simple retraction:
$$\mathcal{R}_{\theta_{t}}(v) = \theta + v$$

\subsection{Graph Reconstruction}
\label{sec:appendix-graph-reco}
\paragraph{Loss Function:}
\label{sec:appendix-loss}
To compute the embeddings, we optimize the distance-based loss function proposed in \citet{gu2019lmixedCurvature}. Given graph distances $\{d_{G}(X_i, X_j)\}_{ij}$ between all pairs of connected nodes, the loss is defined as:
$$\mathcal{L}(x) = \sum_{1 \leq i \leq j \leq n} \left| \left( \frac{d_{\mathcal{P}}(x_i, x_j)}{d_{G}(X_i, X_j)} \right)^2 - 1\right|$$

where $d_{\mathcal{P}}(x_i, x_j)$ is the distance between the corresponding node representations in the embeddings space.
This formulation of the loss function captures the average distortion.
We regard as future work experimenting with different loss functions, similar to the ones proposed on \citet{cruceru20matrixGraph}.

\paragraph{Evaluation Metrics:}
\label{sec:appendix-metrics}
To measure the quality of the learned embeddings we follow the same fidelity metrics applied in \citet{gu2019lmixedCurvature}, which are distortion and precision.
The distortion of a pair of connected nodes $a, b$ in the graph $G$, where $f(a), f(b)$ are their respective embeddings in the space $\mathcal{P}$ is given by:
$$\operatorname{distortion}(a,b) = \frac{|d_{\mathcal{P}}(f(a), f(b)) - d_{G}(a, b)|}{d_{G}(a, b)}$$
The average distortion $D_{avg}$ is the average over all pairs of points. Distortion is a global metric that considers the explicit value of all distances.

The other metric that we consider is the mean average precision (mAP). It is a ranking-based measure for local neighborhoods that does not track explicit distances. Let $G = (V, E)$ be a graph and node $a \in V$ have neighborhood $\mathcal{N}_a = \{b_1, . . . , b_{\operatorname{deg}(a)}\}$, where $\operatorname{deg}(a)$ is the degree of $a$. In the embedding $f$, define $R_{a,b_i}$
to be the smallest ball around $f(a)$ that contains $b_i$ (that is, $R_{a,b_i}$ is the smallest set of nearest points required to
retrieve the $i$-th neighbor of $a$ in $f$). Then: 
$$\operatorname{mAP}(f) = \frac{1}{|V|} \sum_{a \in V} \frac{1}{\operatorname{deg(a)}} \sum_{i=1}^{|\mathcal{N}_a|} \frac{|\mathcal{N}_a \cap R_{a,b_i}|}{|R_{a,b_i}|}$$

\paragraph{Data:}
\label{sec:app-data-stats}
We employ NetworkX \cite{networkx2008hagberg} to generate the synthetic datasets, and their Cartesian and rooted products. The statistics of the synthetic datasets reported in this work are presented in Table~\ref{tab:data-stat}, and a diagram of the graphs can be seen in Figure~\ref{fig:synthetic-graphs}. 

The real-world datasets were downloaded from the Network Repository \cite{networkrepository2015rossi}. Stats are presented in Table~\ref{tab:real-world-data-stat}. 

By triples we mean the 3-tuple $(\textbf{u}, \textbf{v}, d(\textbf{u}, \textbf{v}))$, where $\textbf{u, v}$ represent connected nodes in the graph, and $d(\textbf{u,v})$ is the shortest distance between them.

\begin{table}[!t]
	\centering
	\adjustbox{max width=\linewidth}{
	\begin{tabular}{ccccccc}
		\toprule
		\textbf{Graph} & \textbf{Nodes} & \textbf{Edges} & \textbf{Triples} & \textbf{\begin{tabular}[c]{@{}c@{}}Grid \\ Layout\end{tabular}} & \textbf{\begin{tabular}[c]{@{}c@{}}Tree\\ Valency\end{tabular}} & \textbf{\begin{tabular}[c]{@{}c@{}}Tree\\ Height\end{tabular}} \\
		\midrule
		\textsc{4D Grid} & 625 & 2000 & 195,000 & $(5)^4$ &  &  \\
		\textsc{Tree} & 364 & 363 & 66,066 &  & 3 & 5 \\
		\textsc{Tree $\times$ Grid} & 496 & 1,224 & 122,760 & $4 \times 4$ & 2 & 3 \\
		\textsc{Tree $\times$ Tree} & 225 & 420 & 25,200 &  & 2 & 3 \\
		\textsc{Tree $\diamond$ Grids} & 775 & 1,270 & 299,925 & $5 \times 5$ & 2 & 4 \\
		\textsc{Grid $\diamond$ Trees} & 775 & 790 & 299,925 & $5 \times 5$ & 2 & 4 \\
		\bottomrule
	\end{tabular}
	}
	\caption{Synthetic graph stats}
	\label{tab:data-stat}
\end{table}

\begin{table}[!t]
\centering
\adjustbox{max width=\linewidth}{
\begin{tabular}{lrrr}
\toprule
 \textbf{Graph} & \multicolumn{1}{c}{\textbf{Nodes}} & \multicolumn{1}{c}{\textbf{Edges}} & \multicolumn{1}{c}{\textbf{Triples}} \\
 \midrule
USCA312 & 312 & 48,516 & 48,516 \\
bio-diseasome & 516 & 1,188 & 132,870 \\
csphd & 1,025 & 1,043 & 524,800 \\
road-euroroad & 1,039 & 1,305 & 539,241 \\
facebook & 4,039 & 88,234 & 8,154,741 \\
\bottomrule
\end{tabular}
}
\caption{Real-world graph stats}
\label{tab:real-world-data-stat}
\end{table}

\paragraph{Setup Details:}
For all models and datasets we run the same grid search and optimize the distortion loss, applying \textsc{Rsgd}. We report the average of $5$ runs in all cases. The implementation of all baselines are taken from Geoopt \cite{geoopt2019kochurov}. We train for $3000$ epochs, reducing the learning rate by a factor of $5$ if the model does not improve the performance after $50$ epochs, and early stopping based on the average distortion if the model does not improve after $150$ epochs. We use the burn-in strategy \cite{nickel2017poincare, cruceru20matrixGraph} training with a 10 times smaller learning rate for the first 10 epochs.
We experiment with learning rates from $\{0.05, 0.01, 0.005, 0.001\}$, batch sizes from $\{512, 1024, 2048\}$ and max gradient norm from $\{10, 50, 250\}$.

\paragraph{Experimental Observations:}
We noticed that for some combinations of hyper-parameters and datasets, the learning process for the Bounded domain model becomes unstable. Points eventually fall outside of the space, and need to be projected in every epoch.
We did not observe this behavior on the Siegel model. We consider that these findings are in line with the ones reported on \citet{nickel2018lorentz}, where they observe that the Lorentz model, since it is unbounded, is more stable for gradient-based optimization than the Poincar\'e one.

\subsection{Recommender Systems}
\label{sec:appen-recosys-details}
\paragraph{Setup Details:}
For all models and datasets we run the same grid search and optimize the Hinge loss from Equation~\ref{eq:hinge_loss}, applying \textsc{Rsgd}. We report the average of $5$ runs in all cases. We train for $500$ epochs, reducing the learning rate by a factor of $5$ if the model does not improve the performance after $50$ epochs, and early stopping based on the dev set if the model does not improve after $150$ epochs. We use the burn-in strategy \cite{nickel2017poincare, cruceru20matrixGraph} training with a 10 times smaller learning rate for the first 10 epochs.
We experiment with learning rates from $\{0.1, 0.05, 0.01, 0.005, 0.001\}$, batch sizes from $\{1024, 2048\}$ and max gradient norm from $\{10, 50, 250\}$.

\paragraph{Data:}
\label{sec:appen-recosys-data}
We provide a brief description of the datasets used in the recommender systems experiments.
\begin{itemize}
    \item ml-1m and ml-100k: refers to the MovieLens datasets \cite{harper2015movieLensDatasets}.\footnote{\url{https://grouplens.org/datasets/movielens/}}
    \item last.fm: Dataset of artist listening records from 1892 users \cite{Cantador2011lastFmDataset}.\footnote{\url{https://grouplens.org/datasets/hetrec-2011/}}
    \item meetup: dataset crawled from Meetup.com, where the goal is to recommend events to users \cite{pham2015meetupDataset}. The dataset consists of the data from two regions, New York City (NYC) and state of California (CA), we only report results for NYC.
    % \item musins and pripan: branches from the Amazon dataset \cite{mcauley2013hft, nimcauley2019extamazondata}.\footnote{\url{https://nijianmo.github.io/amazon/index.html}} We adopt the 5-core split for the branches "Musical Instruments" (musins) and "Prime Pantry" (pripan)
\end{itemize}
 
Stats for the datasets are presented in Table~\ref{tab:reco-sys-data-stat}. 

To generate evaluation splits, the penultimate and last item the user has interacted with are withheld as dev and test set respectively.

\begin{table}[!t]
\centering
\adjustbox{max width=\linewidth}{
\begin{tabular}{lrrrc}
\toprule
\textbf{Dataset} & \multicolumn{1}{c}{\textbf{Users}} & \multicolumn{1}{c}{\textbf{Items}} & \multicolumn{1}{c}{\textbf{Interactions}} & \textbf{Density (\%)} \\
\midrule
ml-1m & 6,040 & 3,706 & 1,000,209 & 4.47 \\
ml-100k & 943 & 1,682 & 100,000 & 6.30 \\
last.fm & 1,892 & 17,632 & 92,834 & 0.28 \\
meetup-nyc & 46,895 & 16,612 & 277,863 & 0.04 \\
% musins & 27,530 & 10,620 & 231,392 & 0.08 \\
% pripan & 14,180 & 4,970 & 137,788 & 0.20 \\
\bottomrule
\end{tabular}
}
\caption{Recommender system dataset stats}
\label{tab:reco-sys-data-stat}
\end{table}

\subsection{Node Classification}
\label{sec:appen-nodecls-details}
\paragraph{Setup Details:}
In these experiments, for all datasets we use the cosine distance on the datapoints’
features to compute a complete input distance graph. We employ the available features and normalize them so that each attribute has mean zero and standard deviation one.
Once we have a graph, we embed it in the exact same way than in the graph reconstruction task.
Finally, we use the learned node embeddings as features to feed a logistic regression classifier

\paragraph{Matrix Mapping:} Since the node embeddings lie in different metric spaces, we apply the corresponding logarithmic map to obtain a "flat" representation before classifying. 
For the Siegel upper half-space model of dimension $n$, we apply the following mapping. From each complex matrix embedding $Z = X + iY$ we stack the result of the following operations in matrix form as:
$$
M = 
\begin{pmatrix}
Y+XY^{-1}X & XY^{-1}\\
Y^{-1}X & Y^{-1} \\
\end{pmatrix}    
$$
where $M \in \mathbb{R}^{2n \times 2n}$. This mapping is the natural realisation of HypSPD$_n$ as a totally geodesic submanifold of ${\rm SPD}_{2n}$. Since $M \in {\rm SPD}_{2n}$, finally we apply the LogEig map as proposed by  \citet{huang2017riemannianNetForSPDMatrix}, which yields a representation in a flat space. This operations results in new matrix of the form:
$$\operatorname{LogEig}(M) = 
\begin{pmatrix}
U & V\\
V & -U \\
\end{pmatrix}$$
where $U, V \in Sym(n)$. The final step is to take the upper triangular from $U$ and $V$, and concatenate them as a vector of $n(n+1)$ dimensions.

This procedure is implemented for the Upper half-space. In the case of the Bounded domain model, we first map the points to the upper half-space with the Cayley transform.

\paragraph{Datasets:}
All datasets were downloaded from the UCI Machine Learning Repository \cite{dua2017uciMLRepo}. \footnote{\url{https://archive.ics.uci.edu/ml/datasets.php}} Statistics about the datasets used are presented in Table~\ref{tab:nodecls-data-stat}.

\begin{table}[!t]
\centering
\begin{tabular}{lccr}
\toprule
\textbf{Dataset} & \textbf{Nodes} & \textbf{Classes} & \textbf{Triples} \\
\midrule
\textsc{Iris} & 150 & 3 & 11,175 \\
\textsc{Zoo} & 101 & 7 & 5,050 \\
\textsc{Glass} & 214 & 6 & 22,790 \\
\bottomrule
\end{tabular}
\caption{Machine learning datasets used for node classification.}
\label{tab:nodecls-data-stat}
\end{table}

\subsection{Learning the Weights for Finsler Distances}
\label{a.weightLearning}
Both Finsler metrics $\fone$ and $\finf$ exhibit outstanding results in our experiments. However, there are significant differences in their relative performances depending on the target dataset. $\fone$ and $\finf$ are two metrics among many variants in the family of Finsler metrics. Thus, we propose an alternative of our Finslerian models, by learning weights for the summation of Algo~\ref{alg:distances}, step 5, according to:
$d^{F_{W}}(Z_1,Z_2):\sum_{i=1}^n \left[\alpha_i \operatorname{log}(\nicefrac{1 + d_i}{1 - d_i})\right]$   
where $\alpha_i \in \mathbb{R}$ are model parameters. 
% Max Welling: "the whole endeavor of ML is defining the right inductive bias, and leaving what you don't know to the data"
The intuition behind this variant is to impose an inductive bias through the family of Finsler metrics, while allowing the model to learn from the data which particular metric is more suitable in each case. The model has flexibility to represent $\fone$ or $\finf$, and also explore different variations, such as Finsler metrics of minimum entropy \cite{boland2001minimalEntropyFinslerMetrics}. We report results in Table~\ref{tab:fweightedavg-results}.
We observe that the model recovers the $\fone$ metric in the cases where it is the most convenient, whereas for \textsc{Tree $\diamond$ Grids}, it finds a more optimal solution.

\begin{table}[!t]
\small
\centering
\adjustbox{max width=0.75\linewidth}{
\begin{tabular}{crrrrrr}
\toprule
\multicolumn{1}{l}{} & \multicolumn{2}{c}{\textsc{4D Grid}} & \multicolumn{2}{c}{\textsc{Tree $\times$ Tree}} & \multicolumn{2}{c}{\textsc{Tree $\diamond$ Grids}} \\
\multicolumn{1}{l}{} & $D_{avg}$ & mAP & $D_{avg}$ & mAP & $D_{avg}$ & mAP \\
\cmidrule(lr){2-3}\cmidrule(lr){4-5}\cmidrule(lr){6-7}
$\mathcal S_{4}^{F_{\infty}}$ & 5.92 & 99.61 & 3.31 & 99.95 & 10.88 & 63.52 \\
$\mathcal S_{4}^{F_{1}}$ & 0.01 & \textbf{100.00} & \textbf{1.08} & \textbf{100.00} & 1.03 & 78.71 \\
$\mathcal S_{4}^{F_{W}}$ & \textbf{0.00} & \textbf{100.00} & 1.09 & \textbf{100.00} & \textbf{0.91} & \textbf{93.37} \\
\bottomrule
\end{tabular}
}
\caption{Comparison with learning weights for Finsler metrics.}
\label{tab:fweightedavg-results}
% \vspace{-4mm}
\end{table}

\section{Distance Algorithm Complexity}
\label{app:dist-alg-complexity}

\subsection{Theoretical Complexity}
In this section we discuss the computational theoretical complexity of the different operations involved in the development of this work. We employ Big O notation\footnote{\url{https://en.wikipedia.org/wiki/Big_O_notation}}. Since in all cases operations are not nested, but are applied sequentially, the costs can be added resulting in a polynomial expression. Thus, by applying the properties of the notation, we disregard lower-order terms of the polynomial.

\paragraph{Real Matrix Operations:}
For $n \times n$ matrices with real entries, the associated complexity of each operation is as follows:\footnote{\url{https://en.wikipedia.org/wiki/Computational_complexity_of_mathematical_operations}}
\begin{itemize}
    \item Addition and subtraction: $\bigO(n^2)$
    \item Multiplication: $\bigO(n^{2.4})$
    \item Inversion: $\bigO(n^{2.4})$
    \item Diagonalization: $\bigO(n^{3})$
\end{itemize}

\paragraph{Complex Matrix Operations:}
A complex symmetric matrix $Z\in \Sym(n,\C)$ can be written as $Z = X+ iY$, where $X=\Re(Z),Y=\Im(Z) \in \Sym(n,\R)$ are symmetric matrices with real entries. 
We implement the elemental operations for these matrices with the following associated costs:
\begin{itemize}
    \item Multiplication: $\bigO(n^{2.4})$. It involves $4$ real matrix multiplications, plus additions and subtractions. 
    \item Square root: $\bigO(n^{3})$. It involves a diagonalization and $2$ matrix multiplications.\footnote{\url{https://en.wikipedia.org/wiki/Square_root_of_a_matrix}}
    \item Inverse: $\bigO(n^{2.4})$. It involves real matrix inversions and multiplications \cite{falkenberg2007matrixinverse}.
\end{itemize}

\paragraph{Takagi Factorization:}
This factorization involves complex and real multiplications ($\bigO(n^{2.4})$), and diagonalizations ($\bigO(n^{3})$).
It also involves the diagonalization of a $2n x 2n$ matrix, which implies:
\begin{equation}
    \bigO((2n)^3) = \bigO(8n^3) \simeq \bigO(n^3)
\end{equation}
Therefore, the final boundary for its cost is $\bigO(n^3)$.

\paragraph{Cayley Transform:}
This operation along with its inverse are composed of matrix inversions and multiplications, thus the cost is $\bigO(n^{2.4})$.

\paragraph{Distance Algorithm:}
The full computation of the distance algorithm in the \textbf{upperhalf space} involves matrix square root, multiplications, inverses, and the application of the Cayley transform and the Takagi factorization. Since they are applied sequentially, without affecting the dimensionality of the matrices, we can take the highest value as the asymptotic cost of the algorithm, which is $\bigO(n^{3})$.

For the \textbf{bounded domain}, the matrices are mapped into the upperhalf space by an additional application of the inverse Cayley transform, and then the same distance algorithm is applied. Therefore, in this space the complexity also converges to $\bigO(n^{3})$.

\subsection{Empirical Complexity}
\begin{figure}[!t]
    \centering
    \includegraphics[width=0.9\linewidth,keepaspectratio]{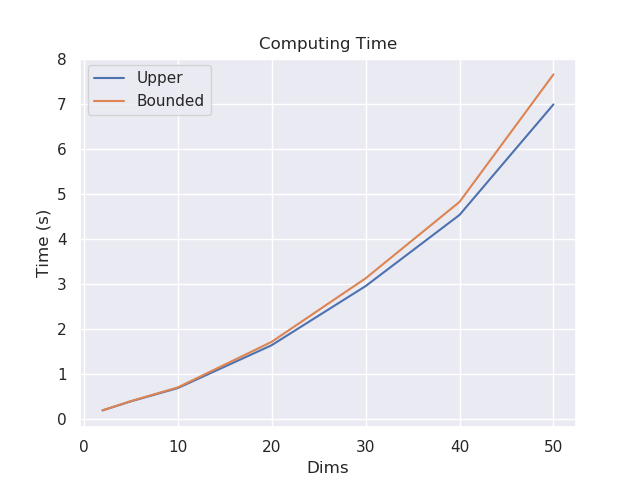}
    \caption{Time of distance calculation for a batch of $1024$ pairs of points for different matrix dimensions.}
    \label{fig:empirical-dist-cost}
\end{figure}

To empirically measure the time involved in the distance calculation we generate a batch of $1024$ pairs of points ($n \times n$ matrices). We perform the time evaluation for different values of $n$. Results can be observed in Figure~\ref{fig:empirical-dist-cost}. 

We observe that as we increase the dimensionality, the relation tends to be polynomial, in line with the theoretical cost stated in \S\ref{sec:complexity-of-distance}.

\section{Vector-valued Distance: a Tool for the Analysis of Embeddings}
\label{a.vvplot}
The vector-valued distance can also be used to develop other tools to analyze the embedding. More specifically we use it not only to create  a {\bf continuous edge coloring} as in Section \ref{sec:vectorialdistance}, but also a {\bf vector distance plot}, and a {\bf continuous node coloring with respect to a root}.

% We focus on HypSPD$_2$, the Siegel space of rank $k=2$, where the vector-valued distance is just a vector in a cone in $\R^2$: $\mathrm{vDist}(Z_1,Z_2)=(v_1,v_2)$ (see Algorithm~\ref{alg:Vdistances}, step $6$). We use the angle of this vector by assigning the ratio $\nicefrac{v_2}{v_1}$ to each edge in the graph. 

\begin{figure}[!t]
    \centering
    \subfloat{\includegraphics[width=0.33\linewidth, keepaspectratio,trim=13mm 7mm 18mm 12mm, clip]{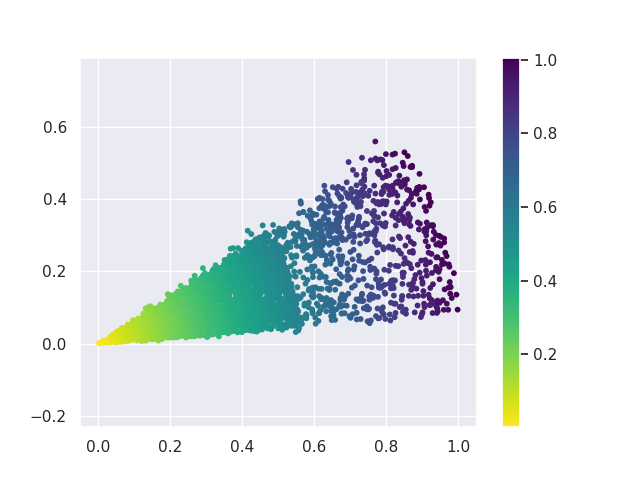}}
    \hfill
    \subfloat{\includegraphics[width=0.33\linewidth, keepaspectratio,trim=13mm 7mm 18mm 12mm, clip]{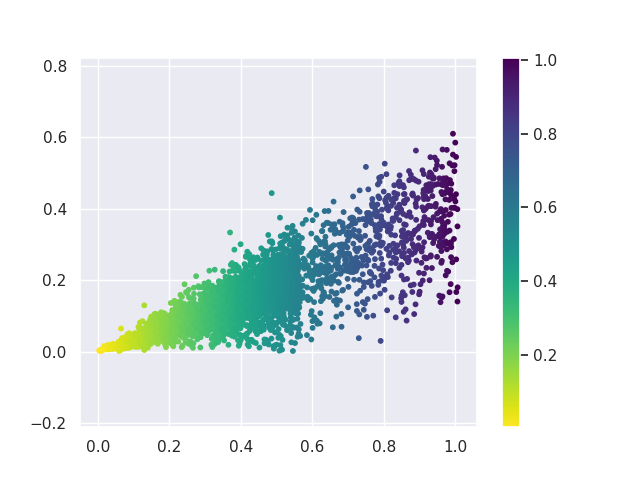}}
    \hfill
    \subfloat{\includegraphics[width=0.33\linewidth, keepaspectratio,trim=13mm 7mm 18mm 12mm, clip]{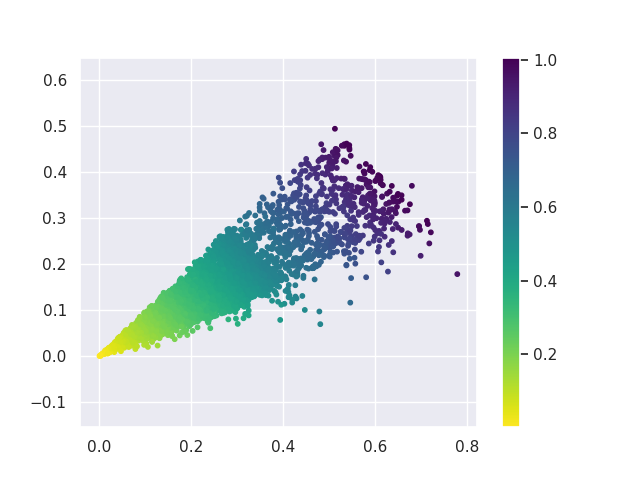}}
    \caption{Plot of $(v_1, v_2)$ of  $\mathcal S_{2}^{R}$ (left), $\mathcal S_{2}^{F1}$ (center), and $\mathcal S_{2}^{F_{\infty}}$ (right) for vertex pairs sampled from \textsc{USCA312}. Color indicates ground-truth distance.}
    \label{fig:d-values-usca}
    \vspace{-3mm}
\end{figure}

\begin{figure}[!t]
\vspace{-2mm}
    \centering
    \subfloat{\includegraphics[width=0.32\linewidth, keepaspectratio]{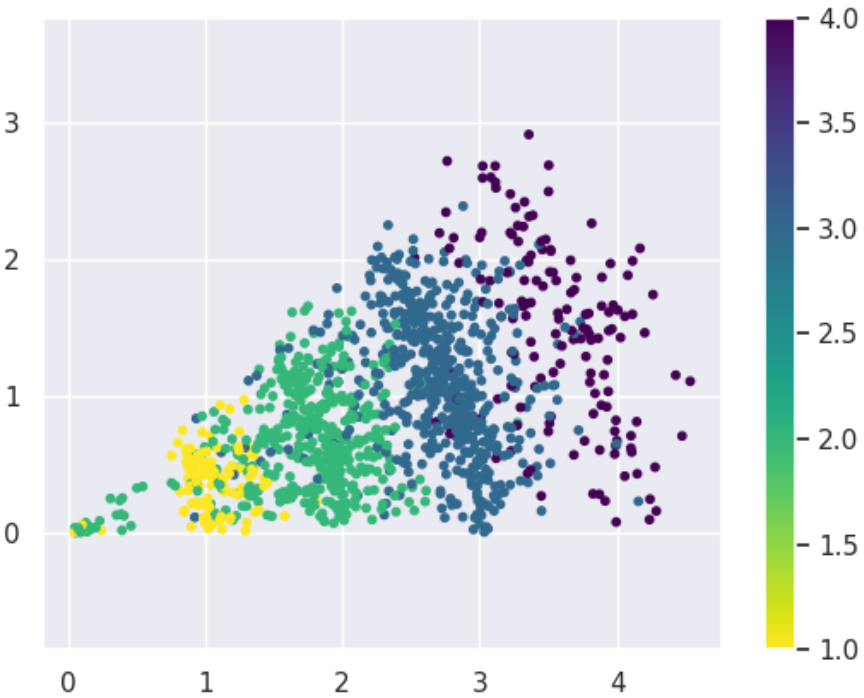}}
    \hfill
    \subfloat{\includegraphics[width=0.33\linewidth, keepaspectratio]{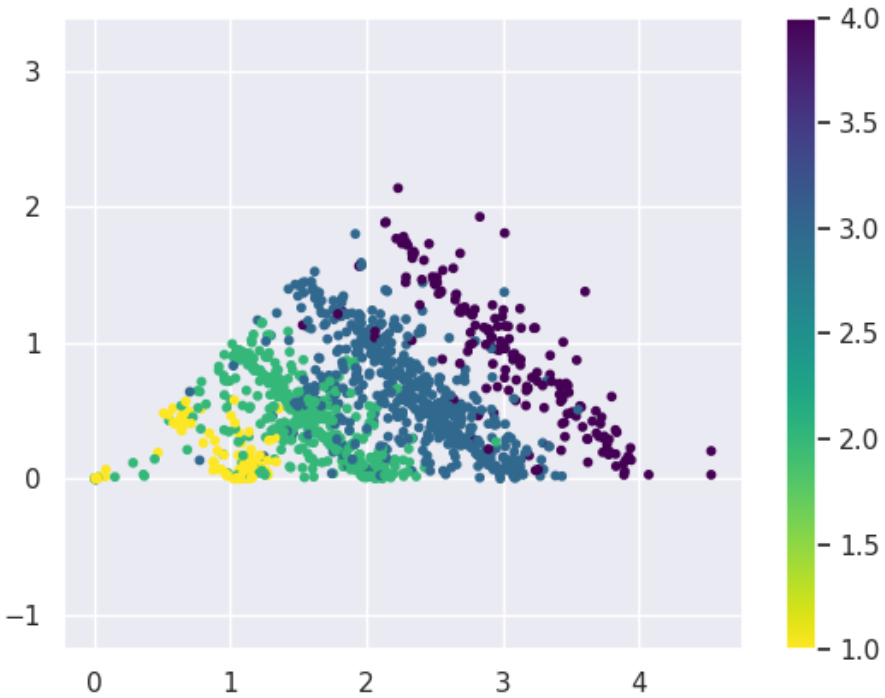}}
    \hfill
    \subfloat{\includegraphics[width=0.32\linewidth, keepaspectratio]{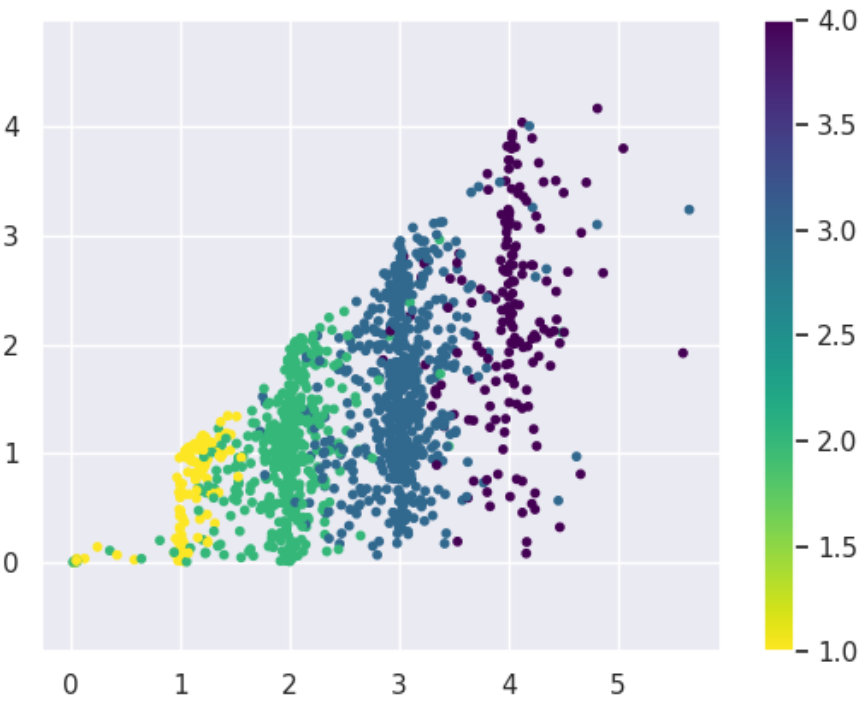}}
    \caption{Plot of $(v_1, v_2)$ of $\mathcal S_{2}^{R}$ (left), $\mathcal S_{2}^{F1}$ (center), and $\mathcal S_{2}^{F_{\infty}}$ (right) for vertex pairs sampled from \textsc{bio-diseasome}. Color indicates ground-truth distance.}
    \label{fig:d-values}
    \vspace{-3mm}
\end{figure}

\begin{figure}[!t]
\vspace{-2mm}
    \centering
    \subfloat{\includegraphics[width=0.33\linewidth, keepaspectratio,trim=13mm 7mm 18mm 12mm, clip]{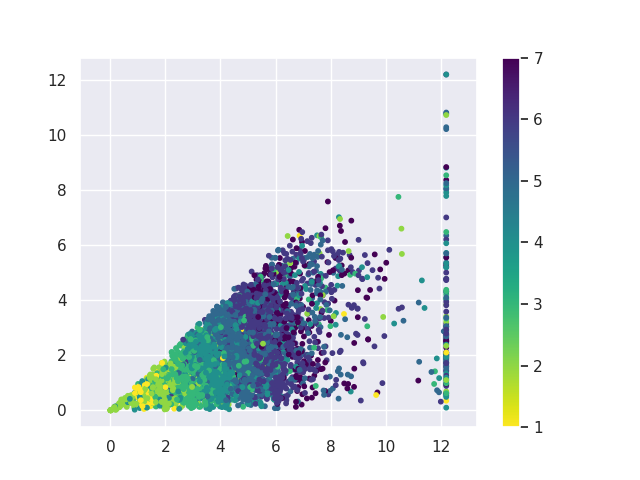}}
    \hfill
    \subfloat{\includegraphics[width=0.33\linewidth, keepaspectratio,trim=13mm 7mm 18mm 12mm, clip]{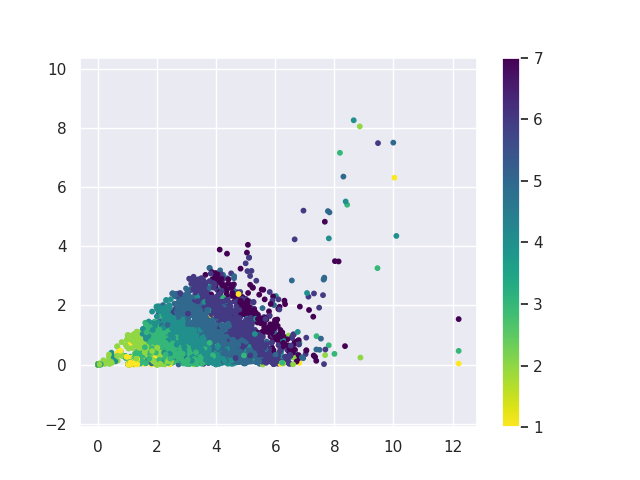}}
    \hfill
    \subfloat{\includegraphics[width=0.33\linewidth, keepaspectratio,trim=13mm 7mm 18mm 12mm, clip]{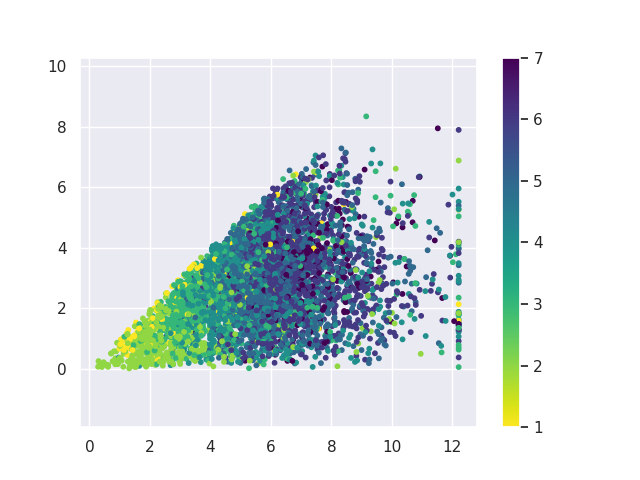}}
    \caption{Plot of $(v_1, v_2)$ of  $\mathcal S_{2}^{R}$ (left), $\mathcal S_{2}^{F1}$ (center), and $\mathcal S_{2}^{F_{\infty}}$ (right) for vertex pairs sampled from \textsc{csphd}. Color indicates ground-truth distance.}
    \label{fig:d-values-csphd}
    \vspace{-3mm}
\end{figure}

\begin{figure}[!t]
    \centering
    \subfloat{\includegraphics[width=0.33\linewidth, keepaspectratio,trim=13mm 7mm 18mm 12mm, clip]{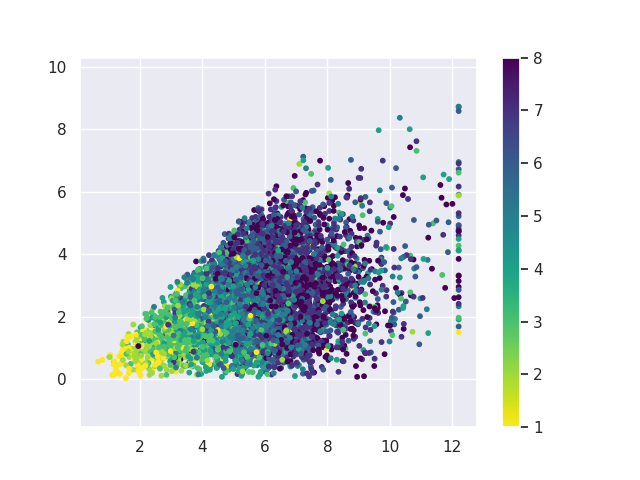}}
    \hfill
    \subfloat{\includegraphics[width=0.33\linewidth, keepaspectratio,trim=13mm 7mm 18mm 12mm, clip]{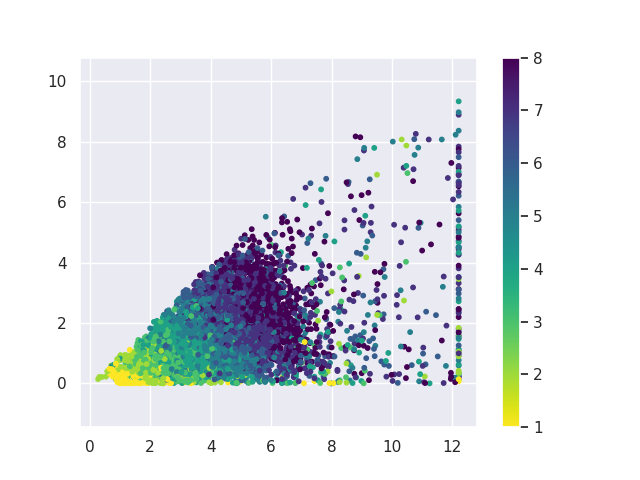}}
    \hfill
    \subfloat{\includegraphics[width=0.33\linewidth, keepaspectratio,trim=13mm 7mm 18mm 12mm, clip]{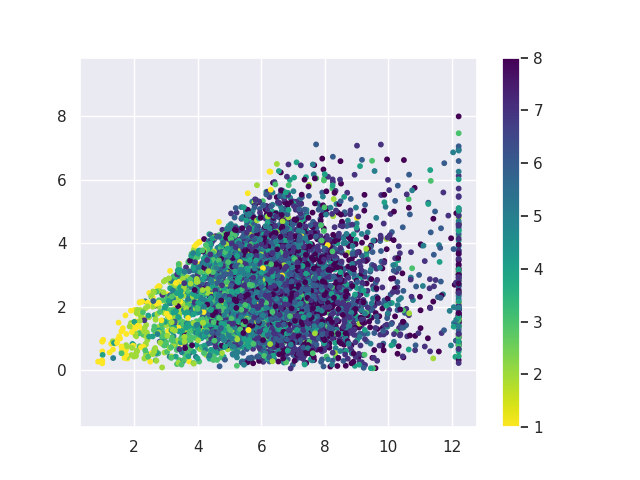}}
    \caption{Plot of $(v_1, v_2)$ of  $\mathcal S_{2}^{R}$ (left), $\mathcal S_{2}^{F1}$ (center), and $\mathcal S_{2}^{F_{\infty}}$ (right) for vertex pairs sampled from \textsc{EuroRoad}. Color indicates ground-truth distance.}
    \label{fig:d-values-euroroad}
    \vspace{-3mm}
\end{figure}

\begin{figure}[!t]
    \centering
    \subfloat{\includegraphics[width=0.33\linewidth, keepaspectratio,trim=13mm 7mm 18mm 12mm, clip]{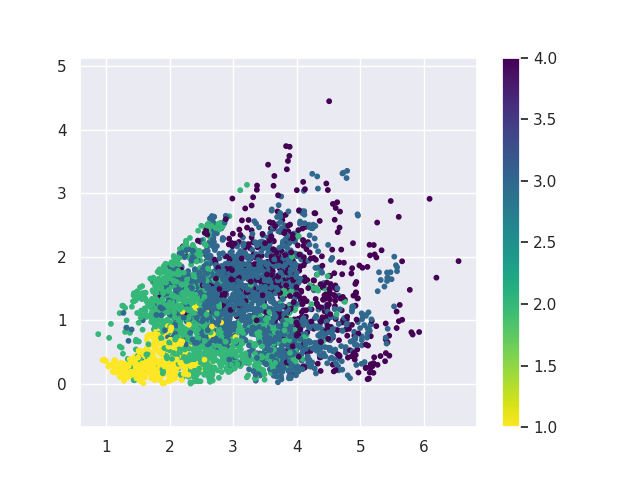}}
    \hfill
    \subfloat{\includegraphics[width=0.33\linewidth, keepaspectratio,trim=13mm 7mm 18mm 12mm, clip]{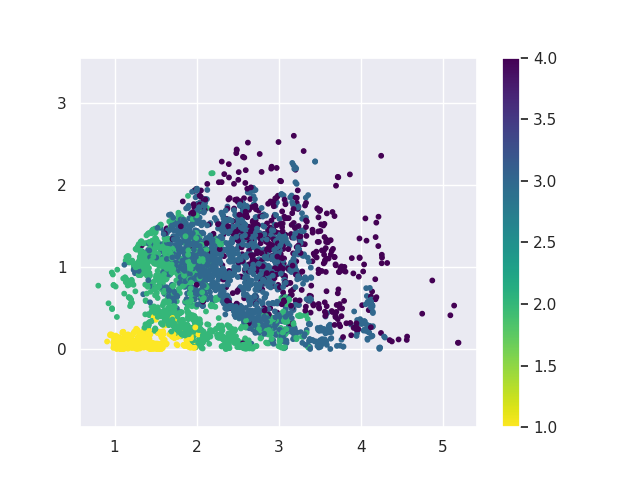}}
    \hfill
    \subfloat{\includegraphics[width=0.33\linewidth, keepaspectratio,trim=13mm 7mm 18mm 12mm, clip]{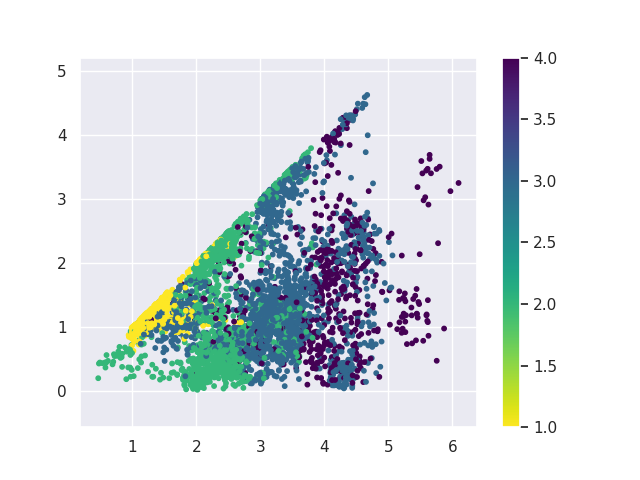}}
    \caption{Plot of $(v_1, v_2)$ of  $\mathcal S_{2}^{R}$ (left), $\mathcal S_{2}^{F1}$ (center), and $\mathcal S_{2}^{F_{\infty}}$ (right) for vertex pairs sampled from \textsc{Grid4D}. Color indicates ground-truth distance.}
    \label{fig:d-values-grid}
    \vspace{-3mm}
\end{figure}

\begin{figure}[!t]
    \centering
    \subfloat{\includegraphics[width=0.33\linewidth, keepaspectratio,trim=13mm 7mm 18mm 12mm, clip]{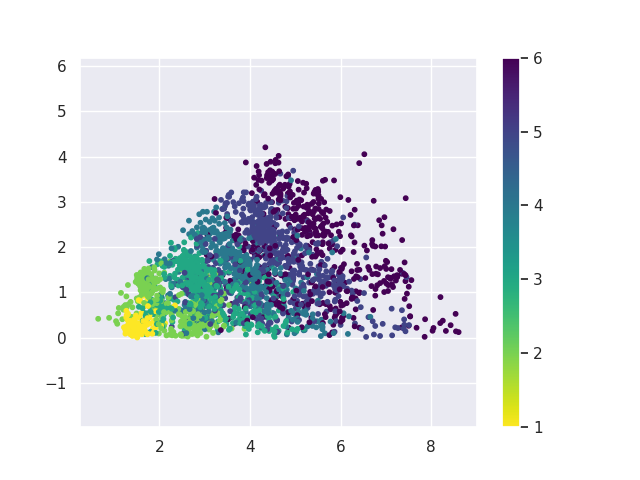}}
    \hfill
    \subfloat{\includegraphics[width=0.33\linewidth, keepaspectratio,trim=13mm 7mm 18mm 12mm, clip]{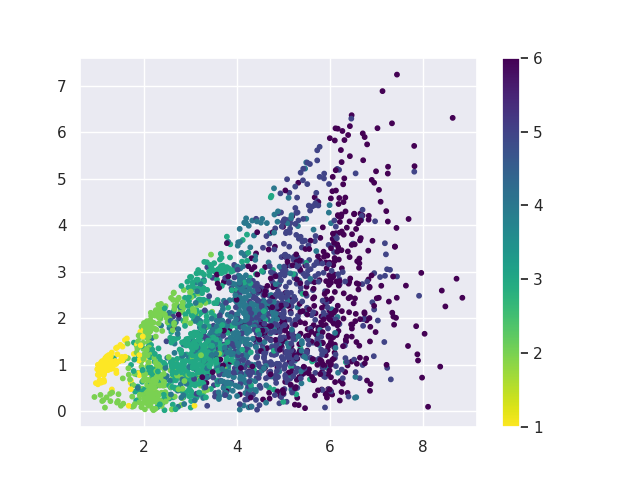}}
    \hfill
    \subfloat{\includegraphics[width=0.33\linewidth, keepaspectratio,trim=13mm 7mm 18mm 12mm, clip]{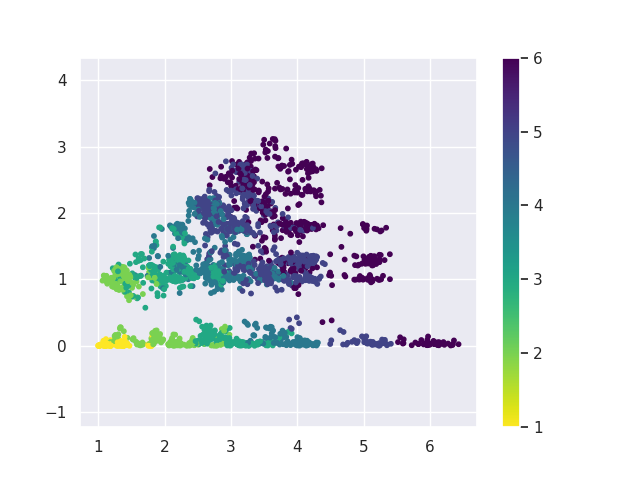}}
    \caption{Plot of $(v_1, v_2)$ of  $\mathcal S_{2}^{R}$ (left), $\mathcal S_{2}^{F1}$ (center), and $\mathcal S_{2}^{F_{\infty}}$ (right) for vertex pairs sampled from \textsc{Tree $\times$ Grid}. Color indicates ground-truth distance.}
    \label{fig:d-values-prod-cart-tree-grid}
    \vspace{-3mm}
\end{figure}

\begin{figure}[!t]
    \centering
    \subfloat{\includegraphics[width=0.33\linewidth, keepaspectratio,trim=13mm 7mm 18mm 12mm, clip]{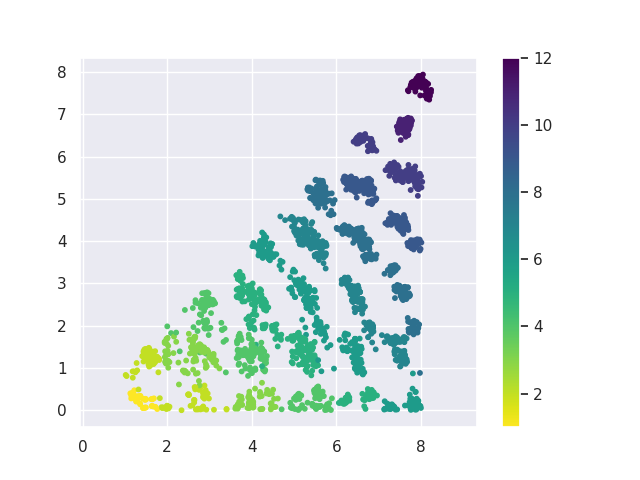}}
    \hfill
    \subfloat{\includegraphics[width=0.33\linewidth, keepaspectratio,trim=13mm 7mm 18mm 12mm, clip]{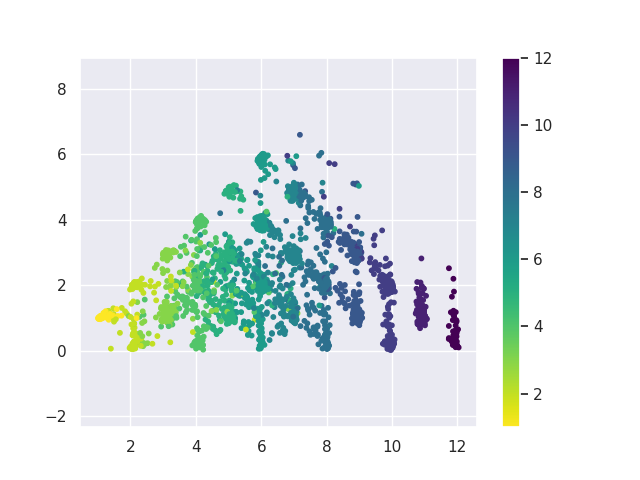}}
    \hfill
    \subfloat{\includegraphics[width=0.33\linewidth, keepaspectratio,trim=13mm 7mm 18mm 12mm, clip]{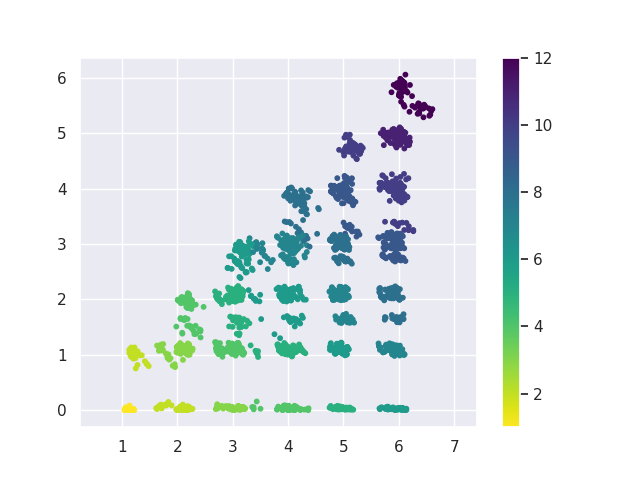}}
    \caption{Plot of $(v_1, v_2)$ of  $\mathcal S_{2}^{R}$ (left), $\mathcal S_{2}^{F1}$ (center), and $\mathcal S_{2}^{F_{\infty}}$ (right) for vertex pairs sampled from \textsc{Tree $\times$ Tree}. Color indicates ground-truth distance.}
    \label{fig:d-values-prod-cart-tree-tree}
    \vspace{-3mm}
\end{figure}

\begin{figure}[!t]
    \centering
    \subfloat{\includegraphics[width=0.33\linewidth, keepaspectratio,trim=13mm 7mm 18mm 12mm, clip]{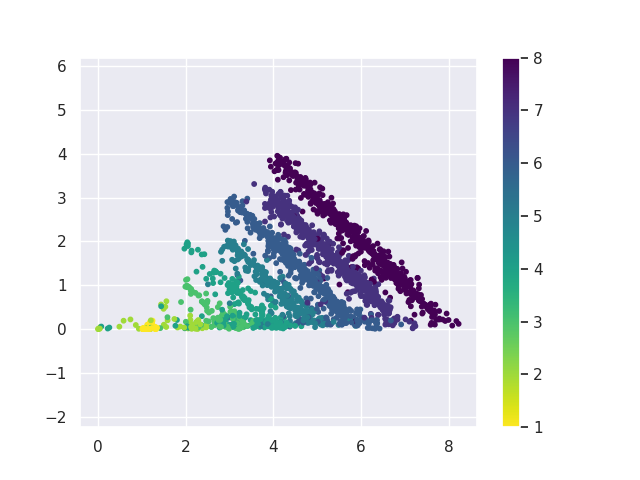}}
    \hfill
    \subfloat{\includegraphics[width=0.33\linewidth, keepaspectratio,trim=13mm 7mm 18mm 12mm, clip]{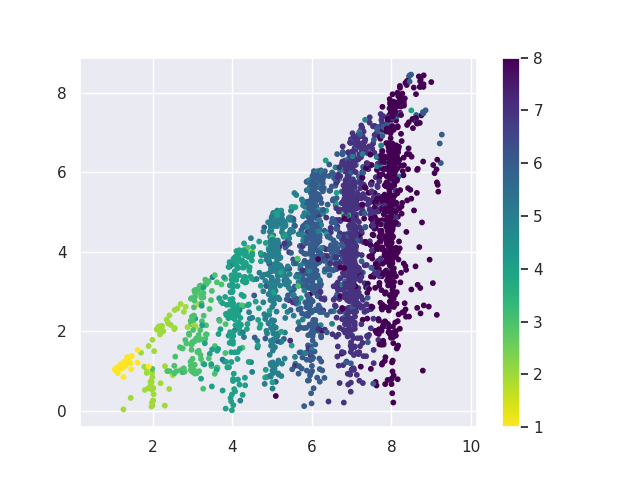}}
    \hfill
    \subfloat{\includegraphics[width=0.33\linewidth, keepaspectratio,trim=13mm 7mm 18mm 12mm, clip]{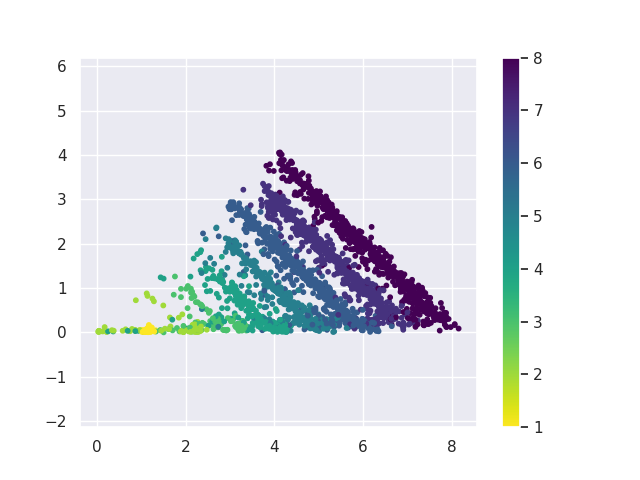}}
    \caption{Plot of $(v_1, v_2)$ of  $\mathcal S_{2}^{R}$ (left), $\mathcal S_{2}^{F1}$ (center), and $\mathcal S_{2}^{F_{\infty}}$ (right) for vertex pairs sampled from \textsc{Tree}. Color indicates ground-truth distance.}
    \label{fig:d-values-tree}
    \vspace{-3mm}
\end{figure}

For the {\bf vectorial distance plot} we sample pairs of connected vertices of the graph $\{z_i, z_j\}$ and plot the result of $\mathrm{vDist}(Z_i,Z_j)=(v_1,v_2)$ (see Algorithm~\ref{alg:Vdistances}, step $6$). In Figure~\ref{fig:d-values-usca}-\ref{fig:d-values-tree} we show the plots of $(v_1, v_2)$ for the embeddings of different dataset embedded into the Upper Half models with respect to Riemannian,   $\fone$ and $F_{\infty}$ metrics. 
In the $\fone$ case, the addition of both $d$-values sums up to the distance, whereas for the $F_{\infty}$, the largest $v$ ($v_1$) corresponds to the distance. The plots match the $\ell^1$ and $\ell^{\infty}$ metrics from Figure~\ref{fig:Finsler}, verifying the intuition about the distances.

The vectorial distance plots give a first qualitative visualization of the embedding. For dissimilar data sets, the edge plots look quite different. They can accumulate near the boundary of the cone (the diagonal or the horizontal), or be evenly distributed. The can also be refined by sampling only specific edges of the graph. 

\begin{figure}[h]
\vspace{-2mm}
    \centering
    \subfloat{\includegraphics[width=0.33\linewidth, keepaspectratio]{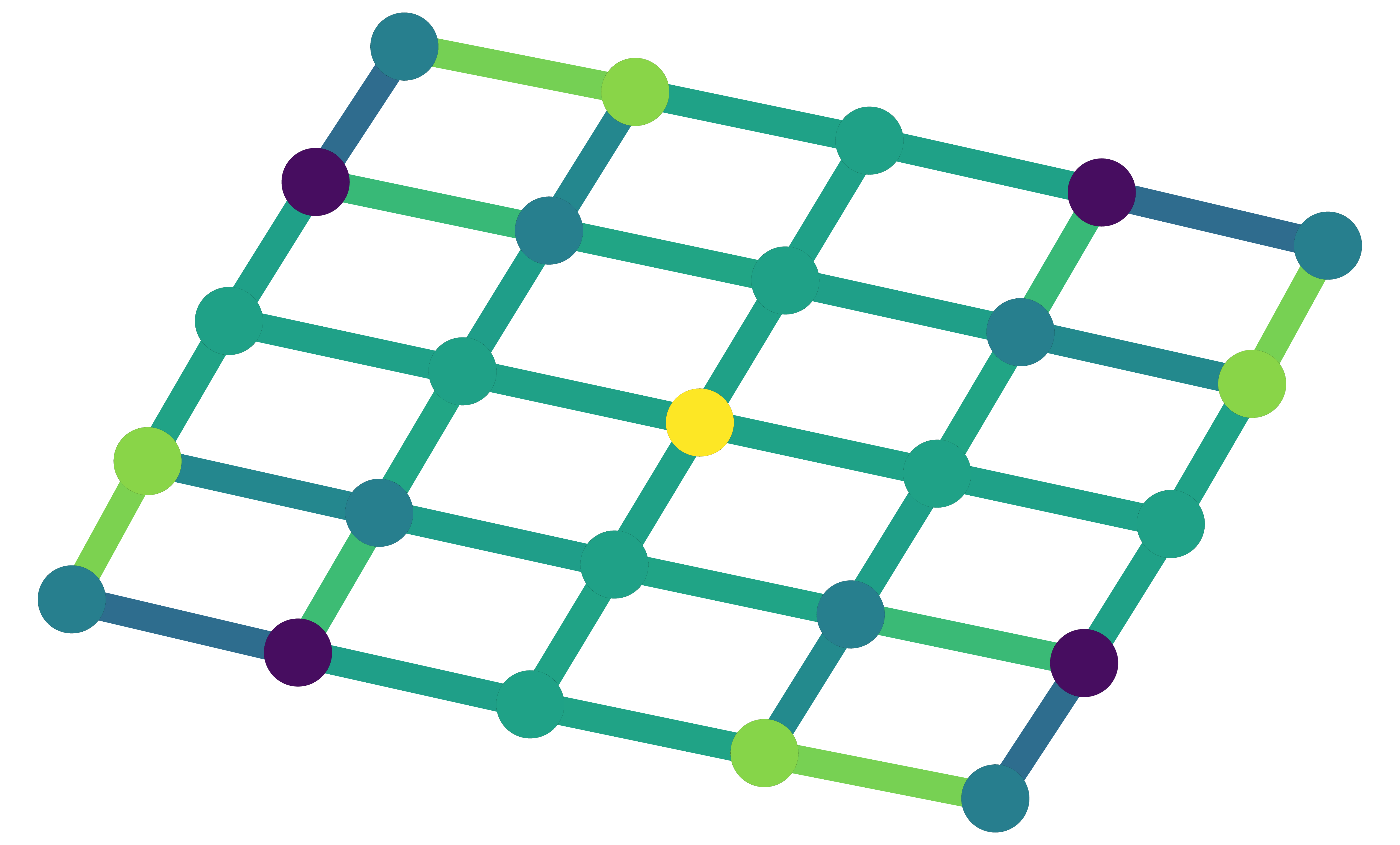}}
    \hfill
    \subfloat{\includegraphics[width=0.33\linewidth, keepaspectratio]{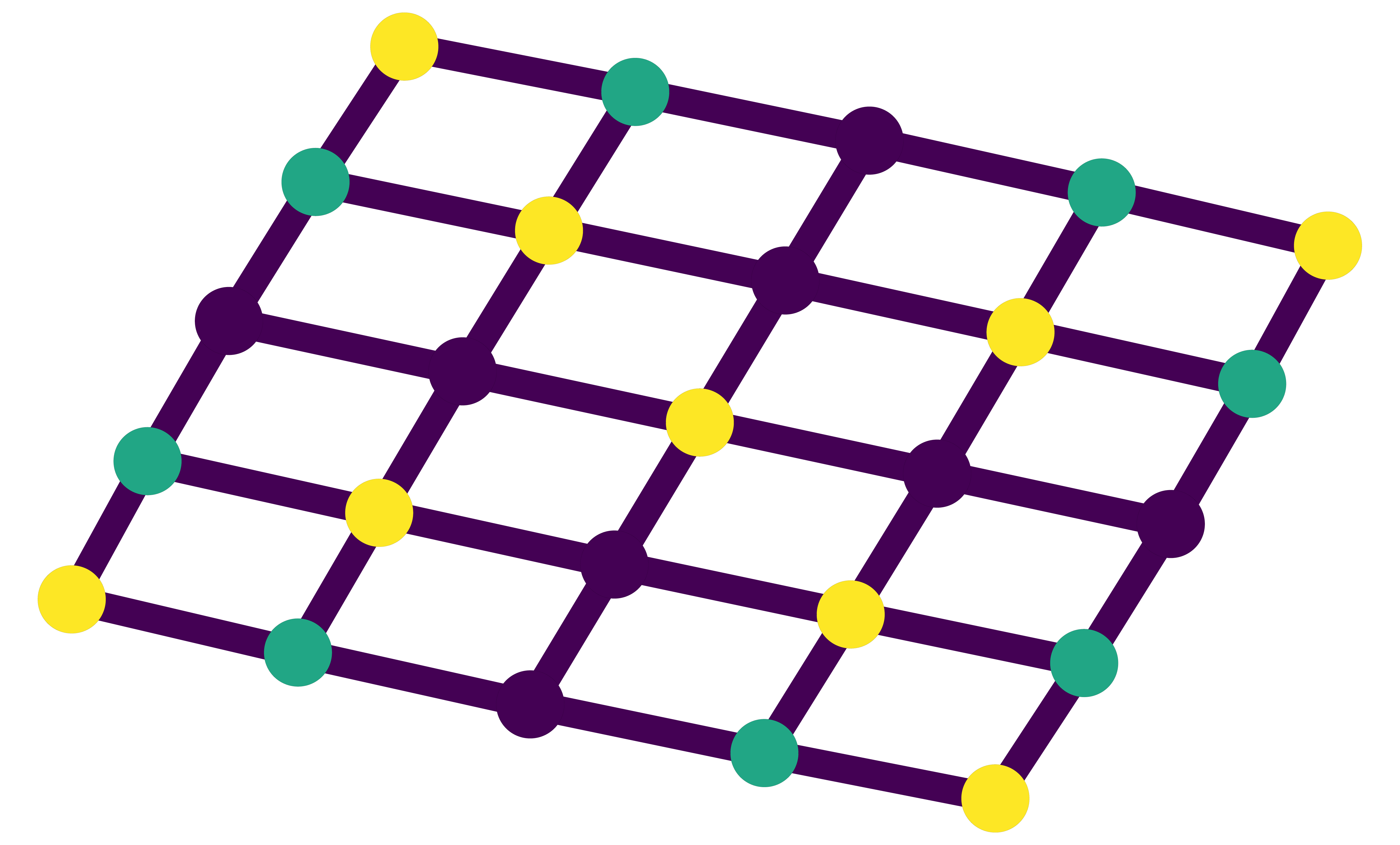}}
    \hfill
    \subfloat{\includegraphics[width=0.34\linewidth, keepaspectratio]{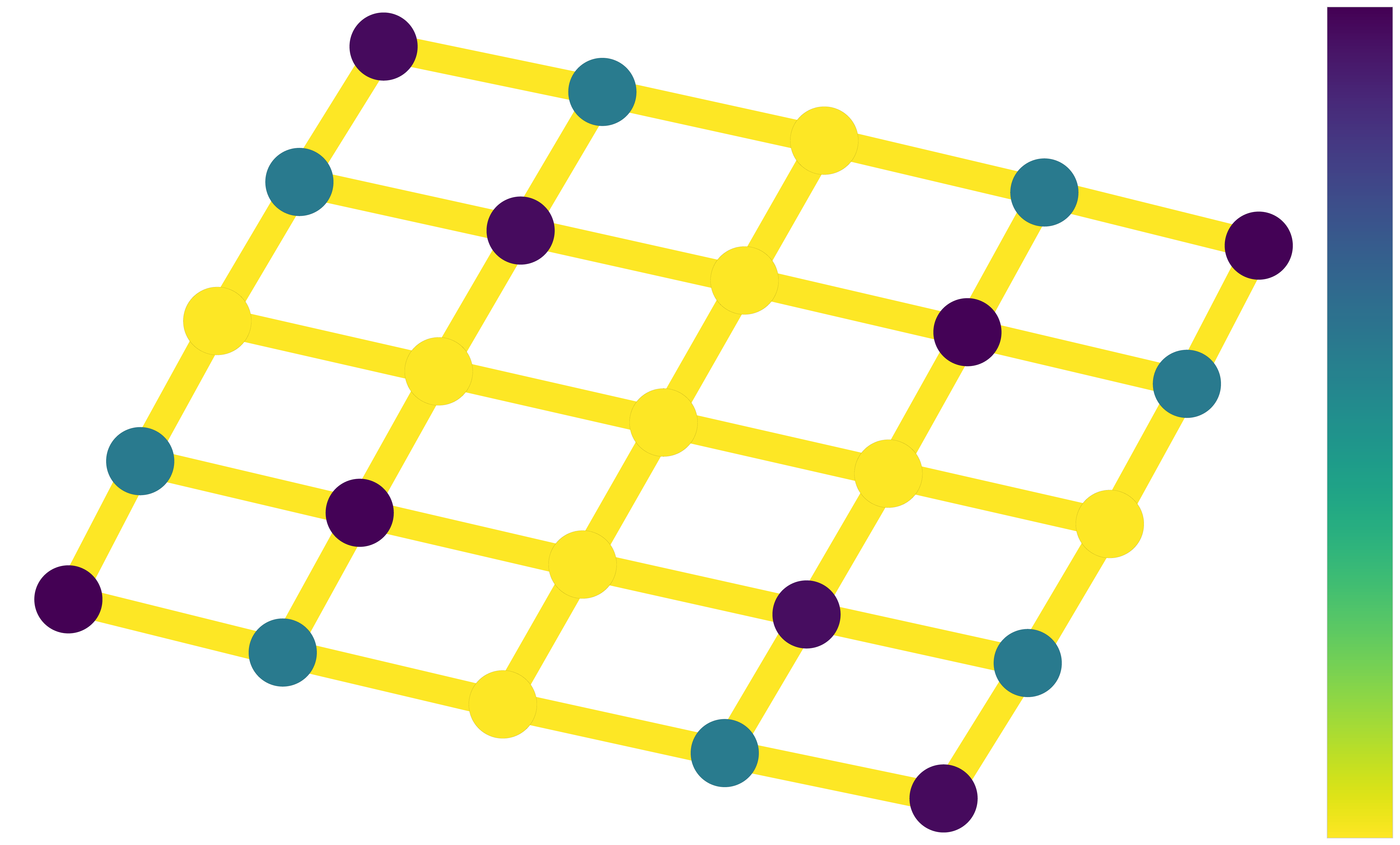}}
    \caption{Analysis of $\mathcal S_{2}^{R}$ (left), $\mathcal S_{2}^{F_{\infty}}$ (center), and $\mathcal S_{2}^{F1}$ (right) for a $5 \times 5$ grid. Node colors indicate the angle of the vector-valued distance by taking the path from the central node. Edge colors indicate the angle for each edge.}
    \label{fig:angle-analysis-grids}
    \vspace{-3mm}
\end{figure}

To construct a {\bf continuous node coloring} we choose one vertex of our graph as the root $r$. For every other node $z$ we take the vector-valued distance from $r$ to $z$, and assign the ratio $\nicefrac{v_2}{v_1}$ as a value for this node. We again represent the corresponding real number by a color shading, as in Figures~\ref{fig:angle-analysis-grids}. It can be thought as the accumulated angle over a path from the root $r$ to the node $z$. 

\textbf{Evaluating the Quality of Embeddings:}
In the case of synthetic graphs, where we have full knowledge of the internal structure, we can use the edge and the node angles to compare the embeddings with respect to the Riemannian distance and the Finsler metrics $\fone$ and $F_{\infty}$. 
Illustrating this with the two dimensional grid, we observe in Figure \ref{fig:angle-analysis-grids} that while in the Finsler metric all edges have the same angle, the embedding optimizing the Riemannian distance is more distorted and less geodesics. In the case of the Finsler distances one can also see more clearly that the symmetries of the graph are respected in the embedding. This shows that the Finsler embeddings are much better in representing structural features of the graphs than the Riemannian embeddings. 

\begin{figure}[!t]
    \centering
    \subfloat{\includegraphics[width=0.33\linewidth, keepaspectratio]{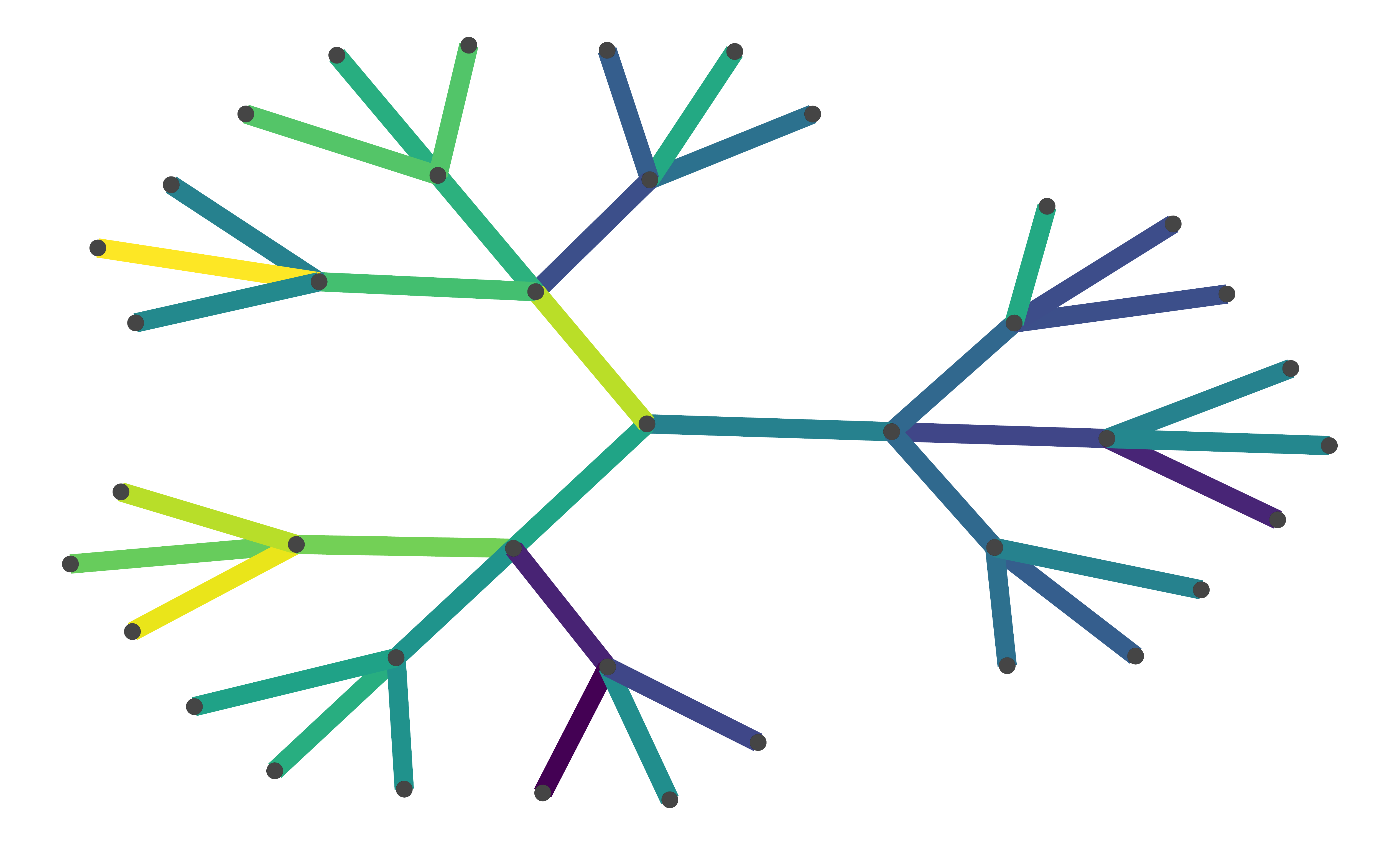}}
    \hfill
    \subfloat{\includegraphics[width=0.33\linewidth, keepaspectratio]{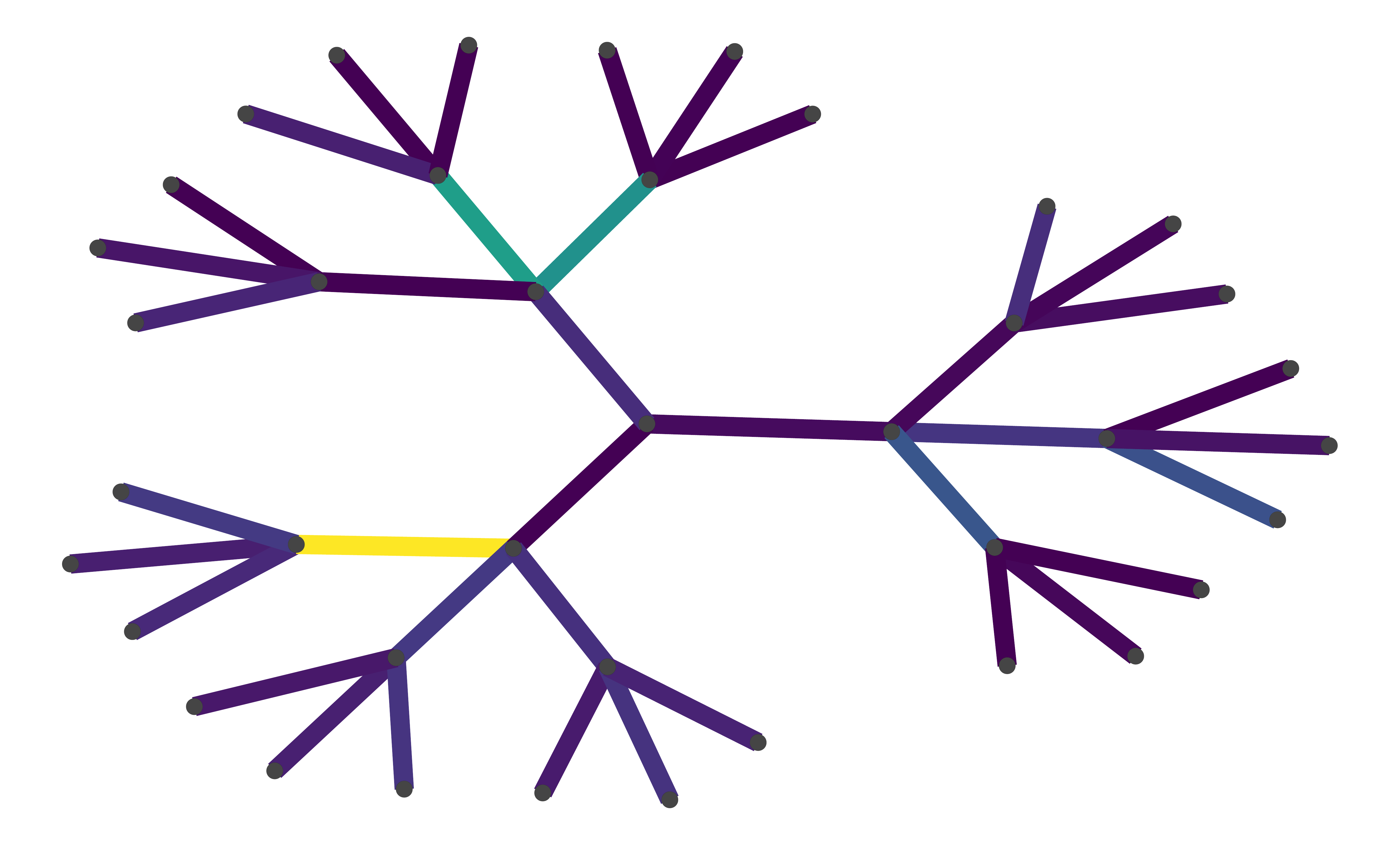}}
    \hfill
    \subfloat{\includegraphics[width=0.34\linewidth, keepaspectratio]{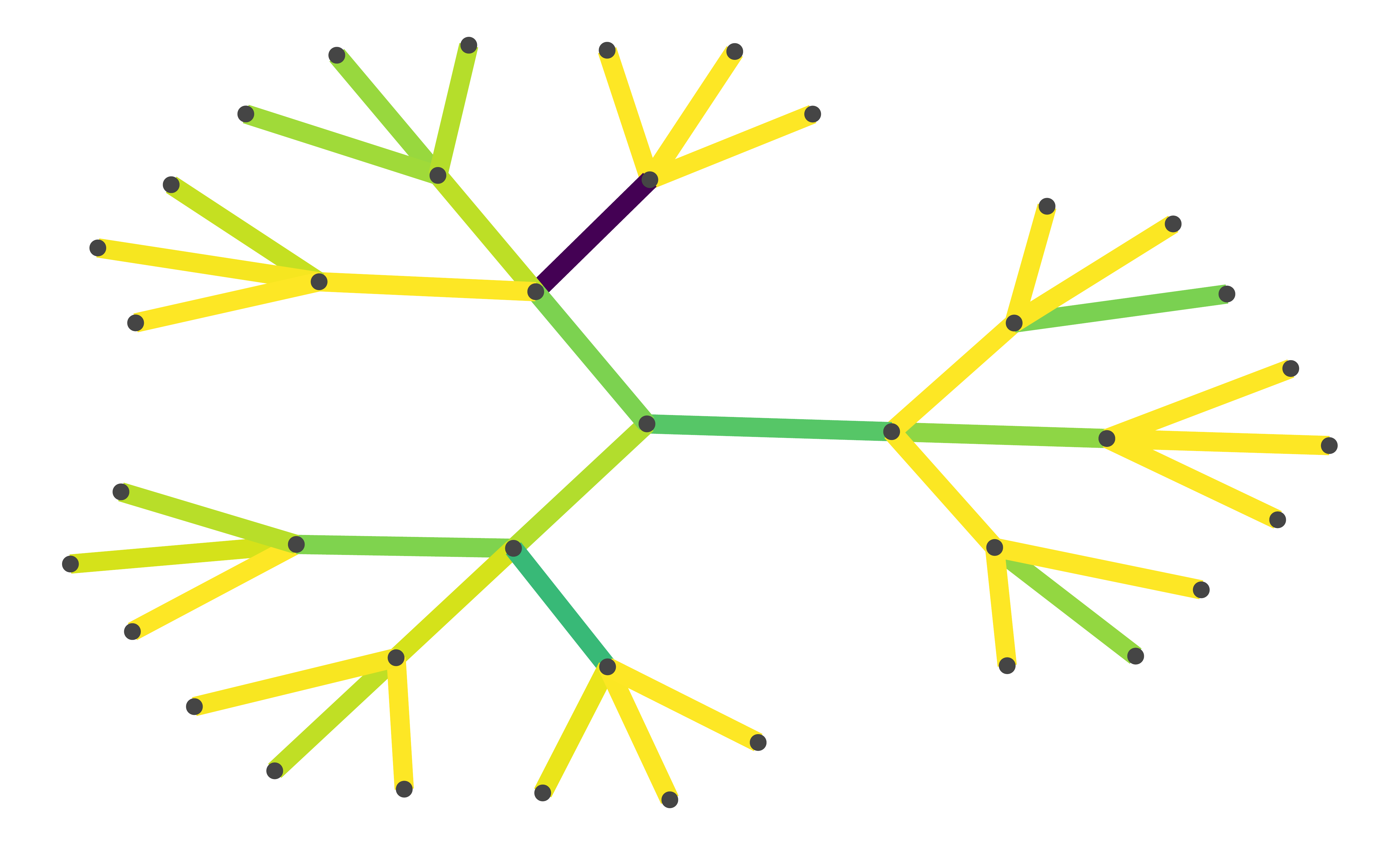}}
    \caption{Edge coloring of $\mathcal S_{2}^{R}$ (left), $\mathcal S_{2}^{F_{\infty}}$ (center), and $\mathcal S_{2}^{F1}$ (right) for a tree.}
    \label{fig:edges-pdf-tree}
\end{figure}

\begin{figure}[!t]
    \centering
    \subfloat{\includegraphics[width=0.33\linewidth, keepaspectratio]{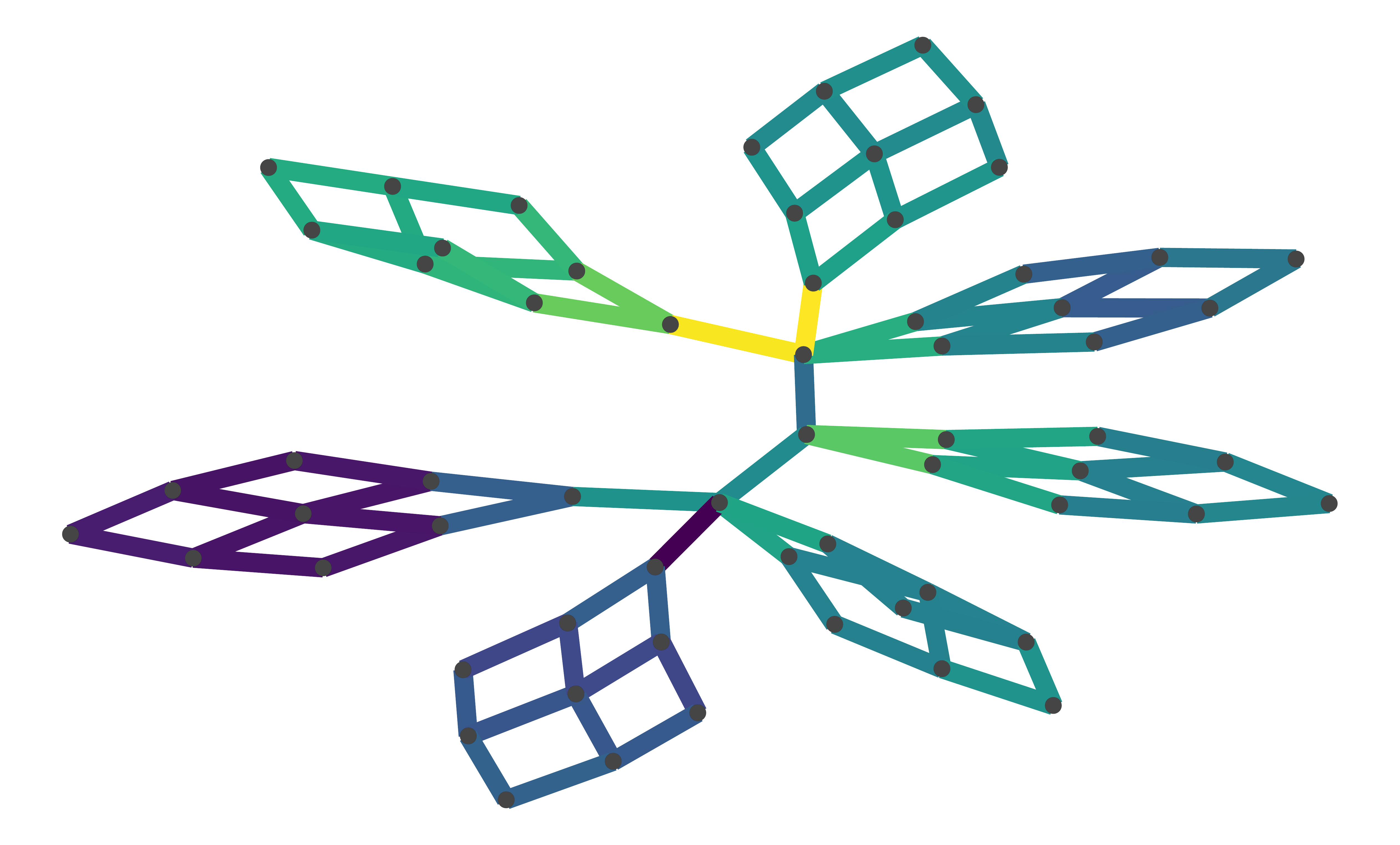}}
    \hfill
    \subfloat{\includegraphics[width=0.33\linewidth, keepaspectratio]{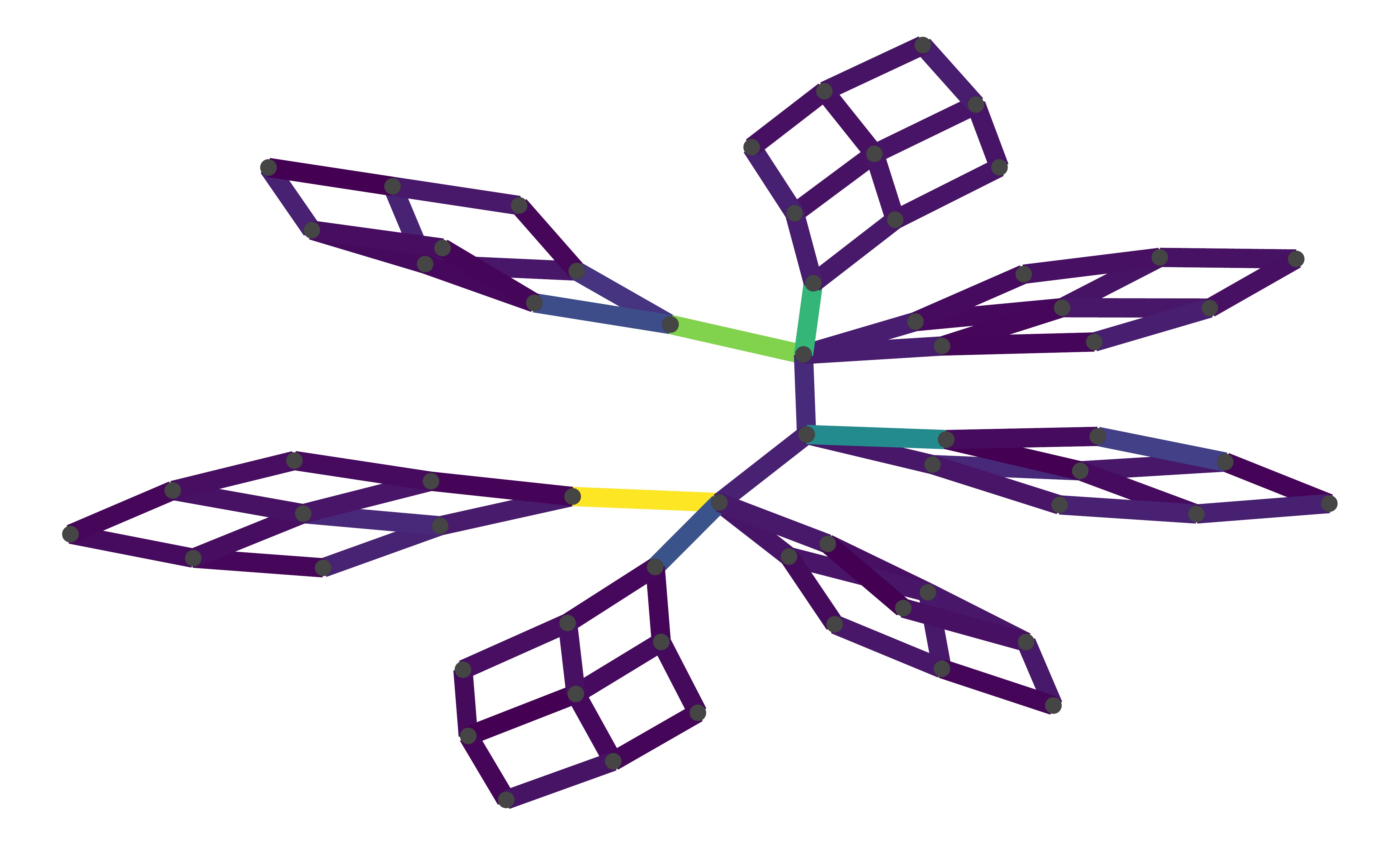}}
    \hfill
    \subfloat{\includegraphics[width=0.34\linewidth, keepaspectratio]{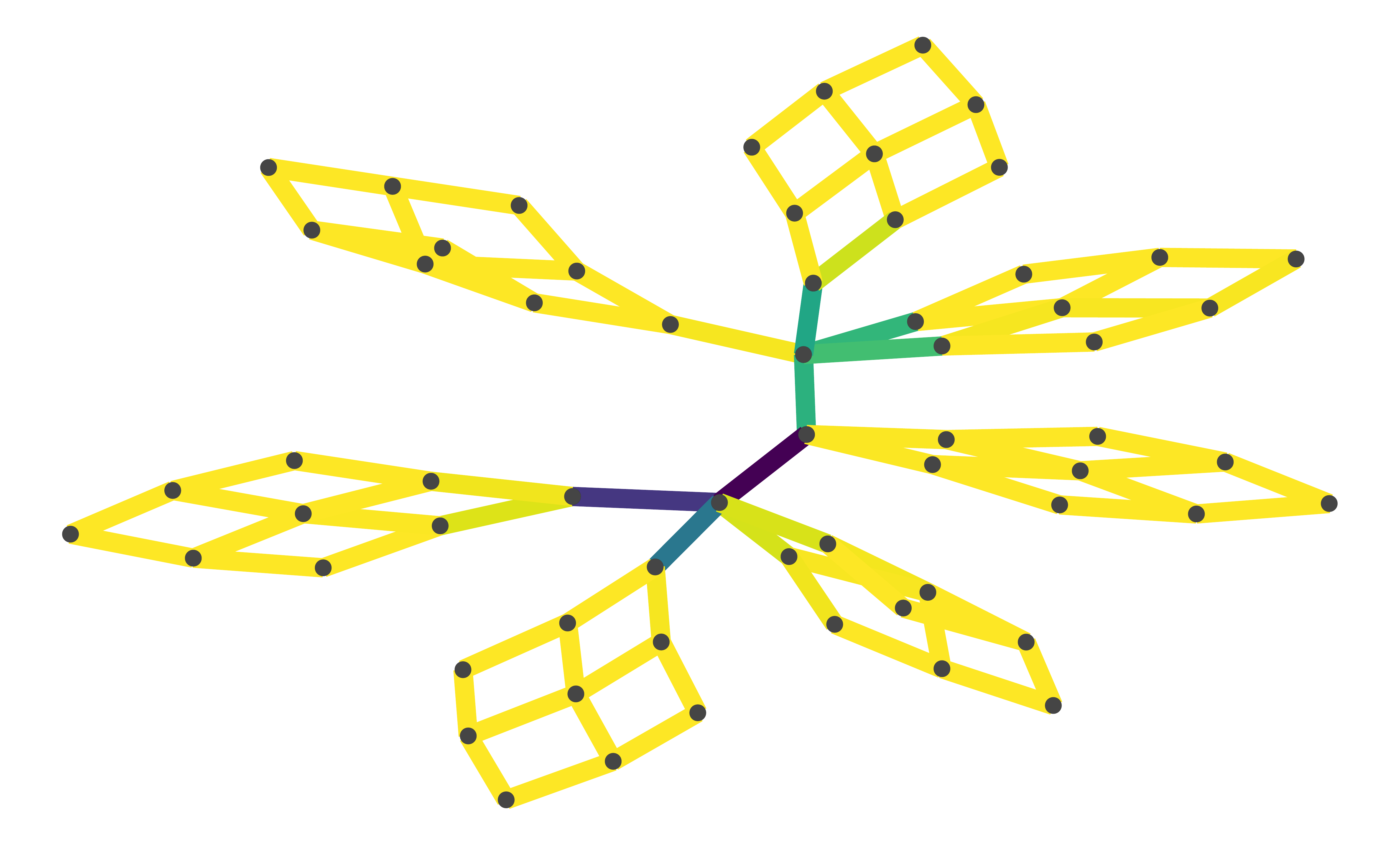}}
    \caption{Edge coloring of $\mathcal S_{2}^{R}$ (left), $\mathcal S_{2}^{F_{\infty}}$ (center), and $\mathcal S_{2}^{F1}$ (right) for a \textsc{tree} $\diamond$ \textsc{grids}.}
    \label{fig:edges-pdf-prod-root-tree-grids}
\end{figure}

\begin{figure}[!t]
    \centering
    \subfloat{\includegraphics[width=0.33\linewidth, keepaspectratio]{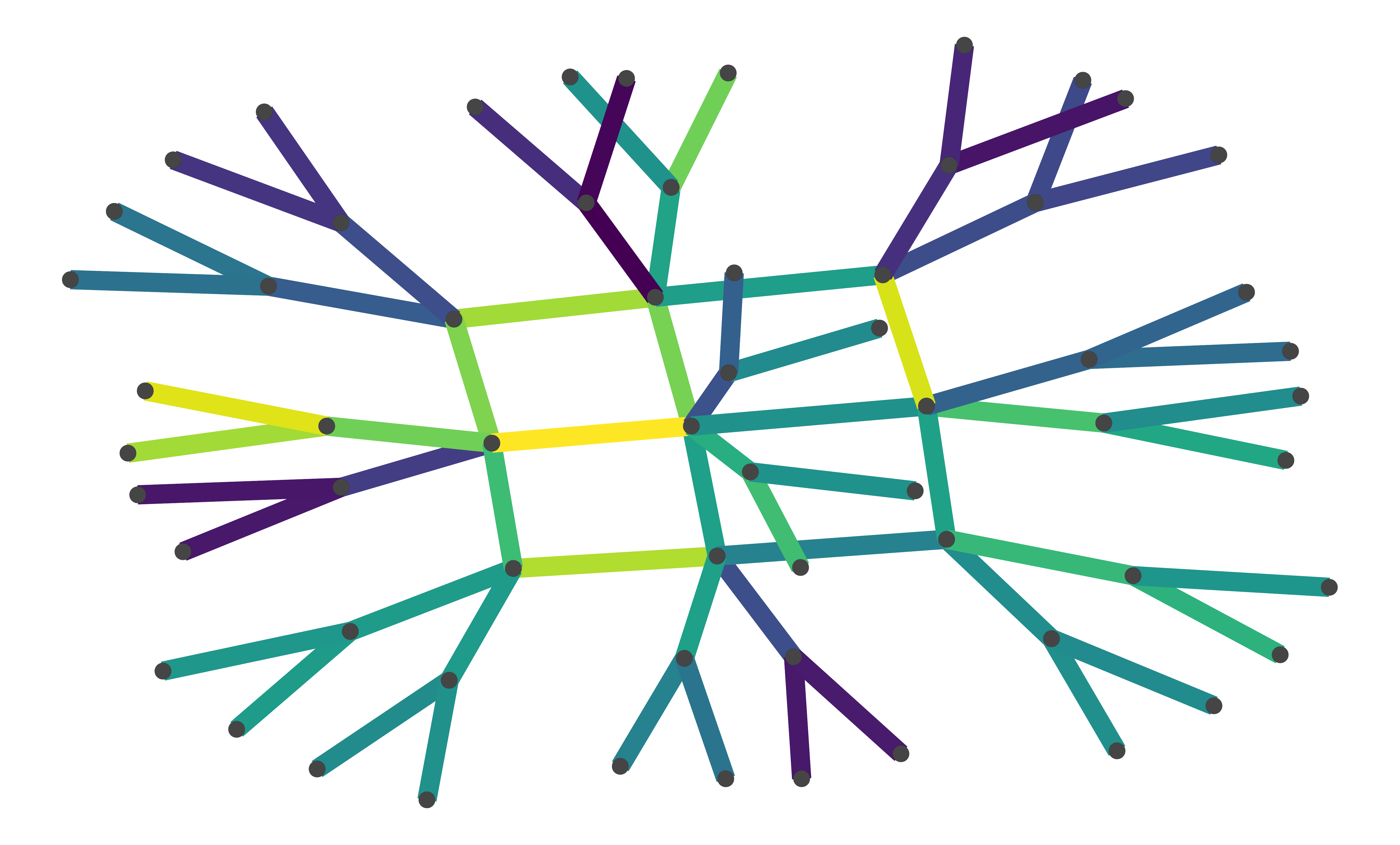}}
    \hfill
    \subfloat{\includegraphics[width=0.33\linewidth, keepaspectratio]{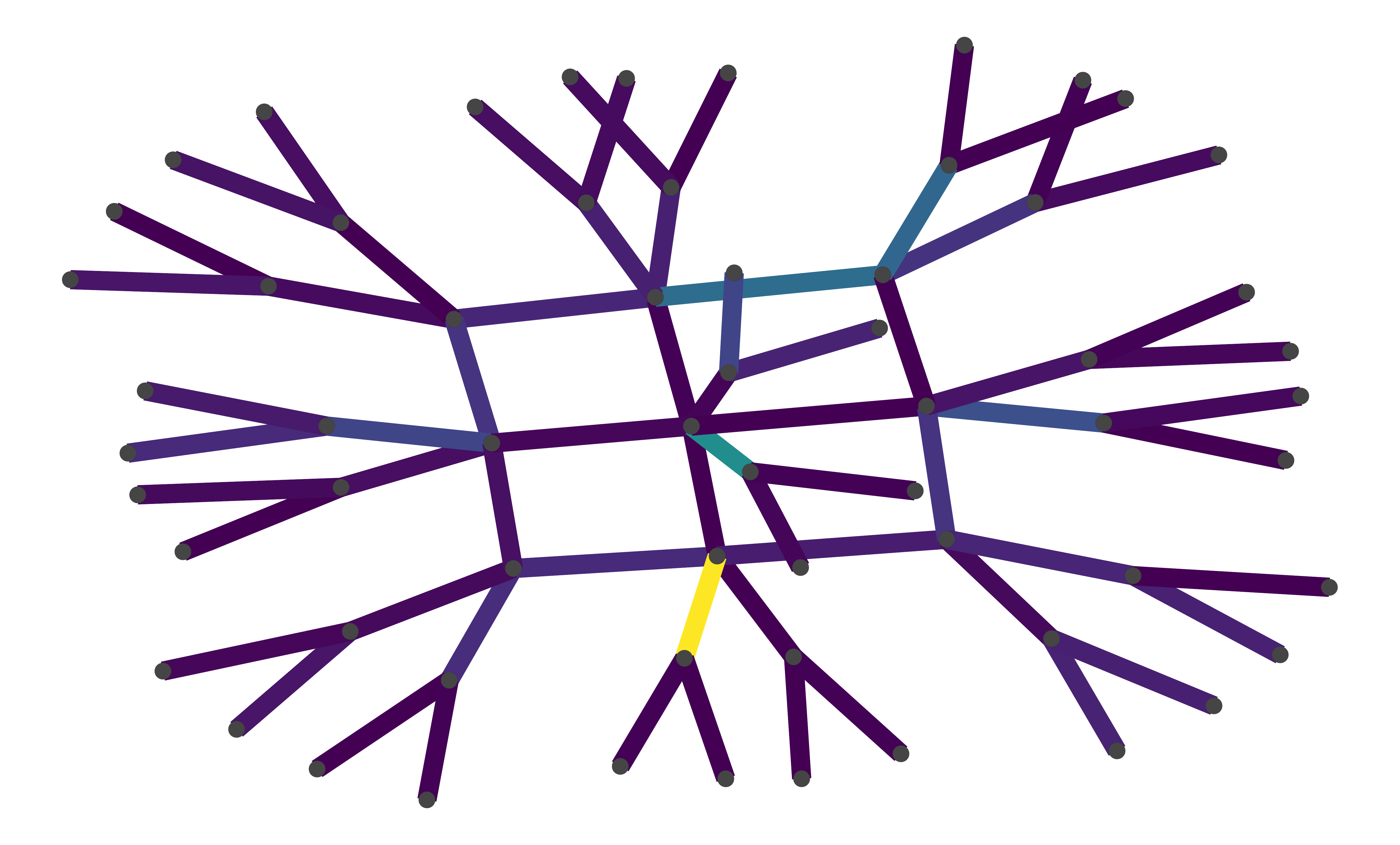}}
    \hfill
    \subfloat{\includegraphics[width=0.34\linewidth, keepaspectratio]{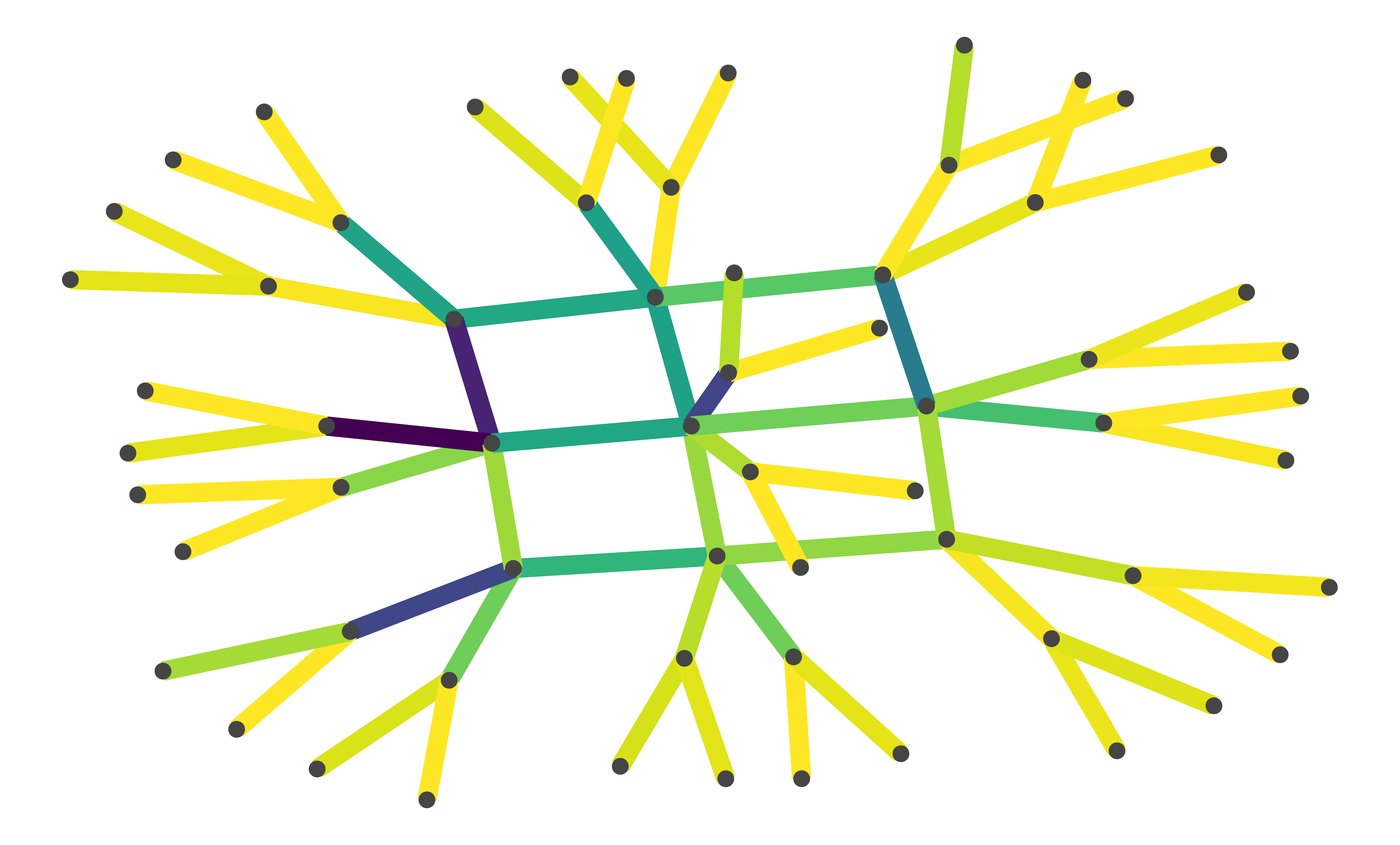}}
    \caption{Edge coloring of $\mathcal S_{2}^{R}$ (left), $\mathcal S_{2}^{F_{\infty}}$ (center), and $\mathcal S_{2}^{F1}$ (right) for a \textsc{grid} $\diamond$ \textsc{trees}.}
    \label{fig:edges-pdf-prod-root-grid-trees}
\end{figure}

\begin{figure}[!t]
    \centering
    \subfloat{\includegraphics[width=0.33\linewidth, keepaspectratio]{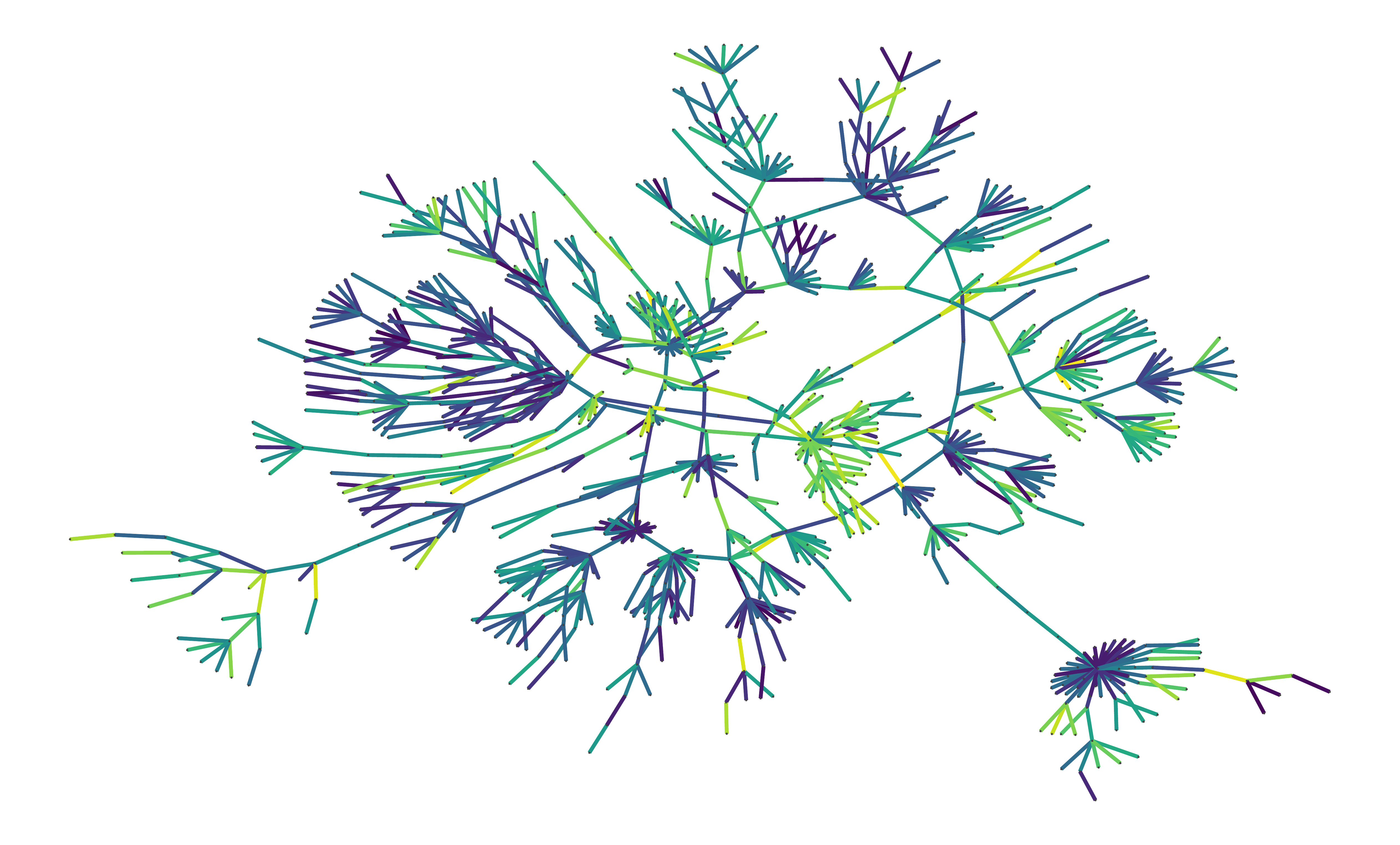}}
    \hfill
    \subfloat{\includegraphics[width=0.33\linewidth, keepaspectratio]{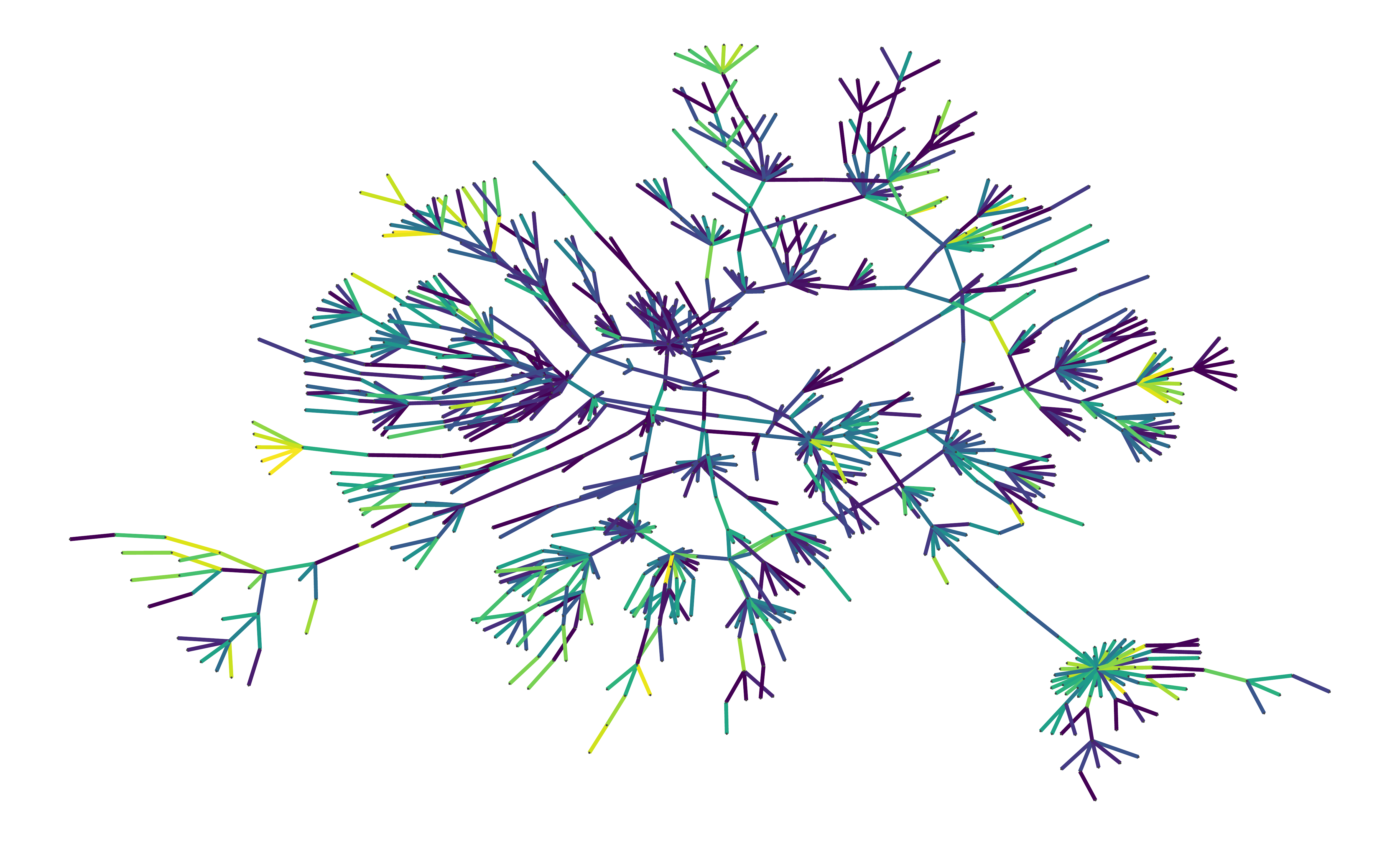}}
    \hfill
    \subfloat{\includegraphics[width=0.34\linewidth, keepaspectratio]{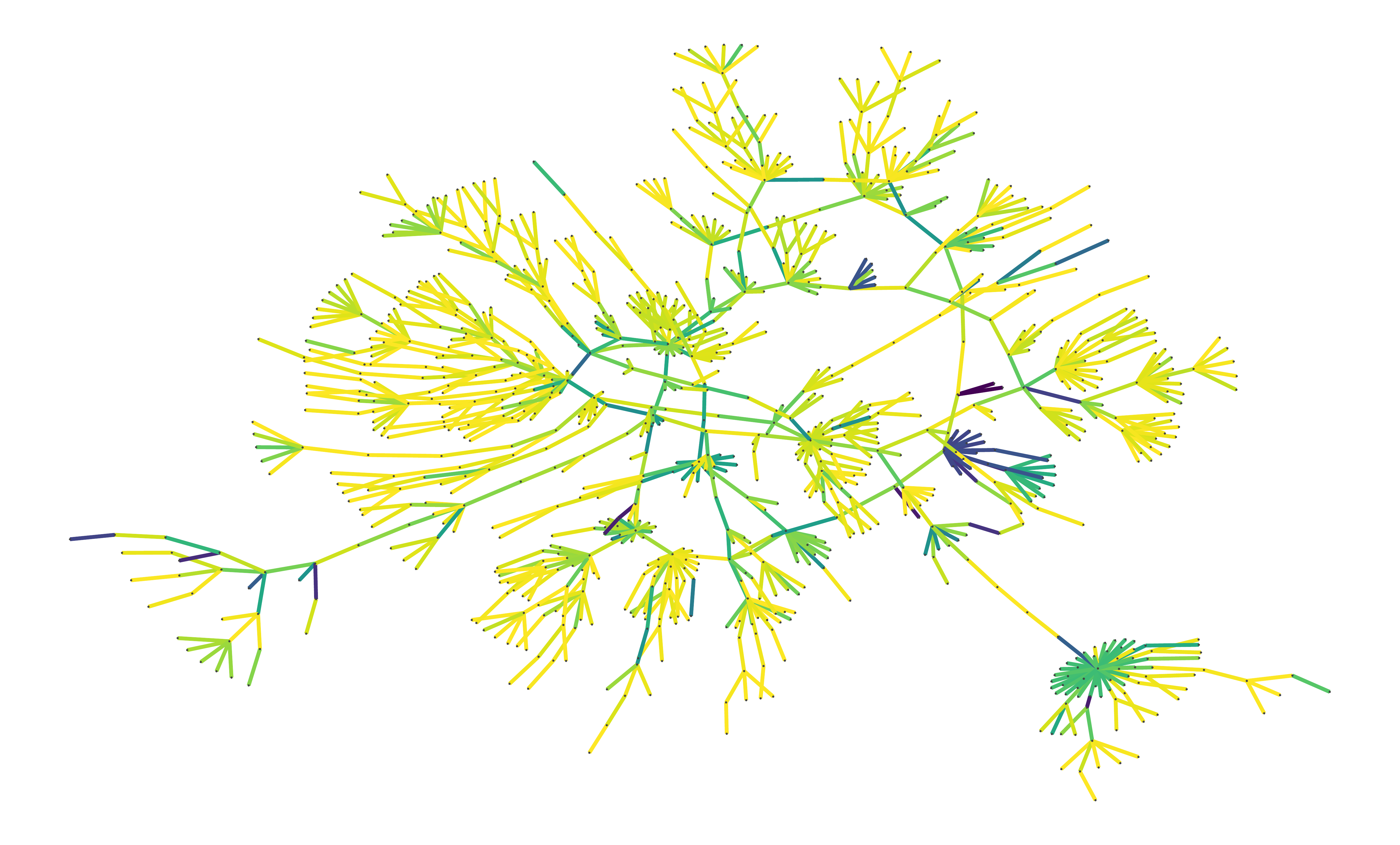}}
    \caption{Edge coloring of $\mathcal S_{2}^{R}$ (left), $\mathcal S_{2}^{F_{\infty}}$ (center), and $\mathcal S_{2}^{F1}$ (right) for \textsc{csphd}.}
    \label{fig:edges-pdf-csphd}
\end{figure}

\paragraph{More Edge Coloring:}
We plot the edge coloring for the three analyzed metric spaces, namely $\mathcal S_{2}^{R}, \mathcal S_{2}^{F_{\infty}}, \mathcal S_{2}^{F1}$ for the datasets analyzed in Figure~\ref{fig:angle-analysis-trees}, and for \textsc{csphd} in Figure~\ref{fig:edges-pdf-tree}-\ref{fig:edges-pdf-csphd}.
We can observe that in the Riemannian metric plots (left-hand side) there is no clear pattern that separates flat and hierarchical components in the graphs.
The $F_{\infty}$ and $F1$ metrics are the best at capturing the structural aspect of the datasets. They recognize very similar patterns, though they assign opposite angles to the vector-valued distance vectors, and this can be noticed from the fact that the colors assigned are in opposite sides of the spectrum (yellow means angles close to zero, blue means angles close to 45°).

To plot these visualizations and the ones of Figure~\ref{fig:angle-analysis-realworld}, we adapted code released\footnote{\url{https://github.com/dalab/matrix-manifolds/blob/master/analysis/plot_ricci_curv.py}} by \citet{cruceru20matrixGraph}. 

% Edge coloring of $\mathcal S_{2}^{F1}$ for a tree (left), and a rooted product of \textsc{tree} $\diamond$ \textsc{grids} (center), and \textsc{grid}  $\diamond$ \textsc{trees}.

\section{More Results}
\label{sec:appendix-results}

Results for graph reconstruction in lower dimensions are presented in Table~\ref{tab:low-dim-synthetic-results}

\begin{table*}[!t]
\small
\centering
\adjustbox{max width=\textwidth}{
\begin{tabular}{crrrrrrrrrrrr}
\toprule
\multicolumn{1}{l}{} & \multicolumn{2}{c}{\textsc{4D Grid}} & \multicolumn{2}{c}{\textsc{Tree}} & \multicolumn{2}{c}{\textsc{Tree $\times$ Grid}} & \multicolumn{2}{c}{\textsc{Tree $\times$ Tree}} & \multicolumn{2}{c}{\textsc{Tree $\diamond$ Grids}} & \multicolumn{2}{c}{\textsc{Grid $\diamond$ Trees}} \\
$(|V|, |E|)$ & \multicolumn{2}{c}{$(625, 2000)$} & \multicolumn{2}{c}{$(364, 363)$} & \multicolumn{2}{c}{$(496, 1224)$} & \multicolumn{2}{c}{$(225, 420)$} & \multicolumn{2}{c}{$(775, 1270)$} & \multicolumn{2}{c}{$(775, 790)$} \\
\multicolumn{1}{l}{} & $D_{avg}$ & mAP & $D_{avg}$ & mAP & $D_{avg}$ & mAP & $D_{avg}$ & mAP & $D_{avg}$ & mAP & $D_{avg}$ & mAP \\
\cmidrule(lr){2-3}\cmidrule(lr){4-5}\cmidrule(lr){6-7}\cmidrule(lr){8-9}\cmidrule(lr){10-11}\cmidrule(lr){12-13}
$\mathbb{E}^{12}$ & 11.24$\pm$0.00 & \textbf{100.00} & 5.71$\pm$0.01 & 32.72 & 9.80$\pm$0.00 & 83.25 & 9.79$\pm$0.00 & 95.97 & 5.11$\pm$0.05 & 22.24 & 5.48$\pm$0.03 & 21.84 \\
$\mathbb{H}^{12}$ & 25.23$\pm$0.06 & 63.86 & 2.09$\pm$0.28 & 97.32 & 17.12$\pm$0.00 & 83.29 & 20.55$\pm$0.12 & 75.98 & 14.12$\pm$0.45 & 44.06 & 14.76$\pm$0.23 & 31.96 \\
$\mathbb{E}^{6} \times \mathbb{H}^{6}$ & 11.24$\pm$0.00 & \textbf{100.00} & 1.61$\pm$0.07 & \textbf{100.00} & 9.20$\pm$0.03 & \textbf{100.00} & 9.34$\pm$0.05 & 98.14 & 2.53$\pm$0.07 & 58.86 & 2.38$\pm$0.04 & \textbf{97.56} \\
$\mathbb{H}^{6} \times \mathbb{H}^{6}$ & 18.76$\pm$0.02 & 79.05 & \textbf{0.92$\pm$0.04} & 99.95 & 12.92$\pm$0.86 & 89.71 & 9.71$\pm$2.47 & 96.82 & \textbf{1.32$\pm$0.08} & \textbf{72.62} & 3.10$\pm$0.62 & 86.40 \\
$\mathcal S_{3}^R$ & 13.26$\pm$0.01 & 99.54 & 1.69$\pm$0.03 & 71.64 & 9.26$\pm$0.01 & 99.57 & 8.80$\pm$0.21 & 97.47 & 1.82$\pm$0.07 & 64.52 & 2.27$\pm$0.18 & 79.10 \\
$\mathcal S_{3}^{F_{\infty}}$ & 11.82$\pm$0.03 & 98.71 & 1.35$\pm$0.39 & 99.35 & 7.98$\pm$0.66 & 99.47 & 3.97$\pm$0.34 & 99.64 & 13.01$\pm$0.64 & 55.89 & 11.26$\pm$0.59 & 68.30 \\
$\mathcal S_{3}^{F_{1}}$ & \textbf{6.41$\pm$0.00} & \textbf{100.00} & 1.07$\pm$0.04 & 74.98 & \textbf{2.02$\pm$0.02} & \textbf{100.00} & \textbf{1.84$\pm$0.02} & \textbf{100.00} & 1.43$\pm$0.01 & 65.90 & \textbf{1.45$\pm$0.05} & 81.25 \\
$\mathcal B_{3}^R$ & 13.29$\pm$0.17 & 99.54 & 1.63$\pm$0.06 & 70.70 & 10.04$\pm$0.03 & 92.20 & 9.10$\pm$0.10 & 96.78 & 4.71$\pm$0.15 & 65.04 & 5.68$\pm$0.37 & 89.19 \\
$\mathcal B_{3}^{F_{\infty}}$ & 12.45$\pm$0.18 & 97.70 & 2.59$\pm$0.34 & 98.94 & 10.33$\pm$0.47 & 93.58 & 4.74$\pm$0.00 & 96.66 & 11.33$\pm$0.10 & 65.07 & 10.39$\pm$0.15 & 79.43 \\
$\mathcal B_{3}^{F_{1}}$ & \textbf{6.41$\pm$0.00} & \textbf{100.00} & 1.13$\pm$0.03 & 79.21 & \textbf{2.02$\pm$0.00} & \textbf{100.00} & 1.92$\pm$0.07 & \textbf{100.00} & 1.51$\pm$0.06 & 71.07 & 1.51$\pm$0.00 & 83.64 \\
\bottomrule
\end{tabular}
}
\caption{Results for synthetic datasets. All models have same number of free parameters. Lower $D_{avg}$ is better. Higher mAP is better.}
\label{tab:low-dim-synthetic-results}
% \vspace{-4mm}
\end{table*}

\end{document}